# CASISR: Circular Arbitrary-Scale Image Super-Resolution

Honggui LI[1*], Zhengyang ZHANG[2], Dingtai LI[3], Sinan CHEN[4], Nahid MD LOKMAN HOSSAIN[5], Xinfeng XU[6], Yinlu QIN[7], Ruobing WANG[8], Hantao LU[9], Yuting FENG[10], Maria TROCAN[11], Dimitri GALAYKO[12], Amara AMARA[13], Mohamad SAWAN[14,15]

[1,2,4,5,6,7,8,9,10]School of Information and Artificial Intelligence, Yangzhou University, Yangzhou 225127, China

[3]Shanghai Qigong Research Institute, Shanghai University of Traditional Chinese Medicine, Shanghai 201203, China

[11]LISITE Research Laboratory, Institut Supérieur d'Électronique de Paris, Paris 75006, France

[12]Laboratoire d'Informatique de Paris 6, Sorbonne University, Paris 75020, France

[13]International Campus (Hangzhou), Beijing University of Aeronautics and Astronautics, Hangzhou 310000, China

[14]School of Engineering, Westlake University, Hangzhou 310024, China

[15]Polystim Neurotech Laboratory, Polytechnique Montreal, Montreal H3T1J4, Canada

[1]hgli@yzu.edu.cn, [2]mz120250945@stu.yzu.edu.cn, [3]22025067@shutcm.edu.cn, [4]mz120240994@stu.yzu.edu.cn, [5]mh23083@stu.yzu.edu.cn, [6]241301123@stu.yzu.edu.cn, [7]241301115@stu.yzu.edu.cn, [8]241303321@stu.yzu.edu.cn, [9]241301211@stu.yzu.edu.cn, [10]241301206@stu.yzu.edu.cn, [11]maria.trocan@isep.fr, [12]dimitri.galayko@sorbonne-universite.fr, [13]amara@buaa.edu.cn, [14]sawan@westlake.edu.cn, [15]mohamad.sawan@polymtl.ca

[*]Corresponding Author

**Abstract**: The generalization performance (GP) of deep learning-based arbitrary-scale image super-resolution (ASISR) methods is subject to limited training datasets and unlimited testing datasets. It is vitally significant to enhance the GP of the pretrained ASISR models by making full use of the testing samples. The ASISR models usually employ an open-loop architecture from low-resolution (LR) images to super-resolution (SR) images. The degradation model from SR samples to LR samples is known bicubic down-sampling for the classical ASISR, is supposed down-sampling with additive random noise for the blind ASISR, and is learnable for the real-world ASISR. Combining the ASISR and degradation models, it is potentially possible to adopt a closed-loop architecture based on the automatic control theory for strengthening the GP of the ASISR methods. Therefore, this paper proposes a closed-loop architecture, circular ASISR (CASISR), to lift the capability of image reconstruction. A mathematical nonlinear loop equation is established to describe the CASISR, the reasonability of the CASISR is proven by conditional probability theory, and the stability of the CASISR is proven by Taylor series approximation. The first-order and second-order absolute difference images are defined to compare the image reconstruction performance of the ASISR and the CASISR methods. Comprehensive simulation experiments show that the proposed CASISR approach outperforms the eight state-of-the-art ASISR approaches in the quality of image reconstruction. Especially, the proposed CASISR is extraordinarily suitable for fractional SR scale factors and is extremely effective for text and stripe images with drastically changed edges.

# 1 Introduction

Arbitrary-scale image super-resolution (ASISR) generates enlarged high-quality super-resolution (SR) images at any given integral or fractional scale factors from degraded low-resolution (LR) images [1]. Compared with the image super-resolution (ISR) at integral scale factors, the ASISR is a more fundamental low-level vision technique and holds more practical applications in high-level vison tasks, such as image classification, image detection, image segmentation, image recognition, image understanding, and image generation [2-3].

Deep learning-based ASISR is the current mainstream method. However, it is data-driven and its generalization performance (GP) is constrained to the fixed training datasets and the unfixed testing datasets [4-5]. The GP can be efficiently enhanced to a great extent by complicated deep neural network architecture, sophisticated loss function, smart training strategy, and sufficient training datasets. The further enhancement of the GP is hindered by the interior difference between the training samples and the testing samples. This is due to the fact that the total number of the training samples is limited and they cannot cover all the situations of the diverse testing samples. Especially for the out-of-sample testing images, their GP dramatically decreases compared with that of the in-sample testing images. The GP of the ASISR depends on both training and testing datasets. Therefore, it is necessary to strengthen the GP of the pretrained ASISR models via making fully use of the testing datasets.

Deep learning-based ASISR commonly utilizes an open-loop framework from the LR images to the SR images [6]. Actually, the degradation (DG) model from the SR images to the LR images for the ASISR is given, supposed, or learnable. The DG model of the classical ASISR is known bicubic down-sampling, that of the blind ASISR is supposed down-sampling followed by additive random noise, and that of the real-word ASISR is learnable. A closed-loop framework can be established by combining the forward SR model and the backward DG model. According to the automatic control theory, the closed-loop structure holds superiority over the open-loop counterpart in dynamic and static performance [7]. Hence, it is also necessary to build the closed-loop infrastructure based on the SR and DG models to lift the GP of the pretrained ASISR models.

This paper proposes a closed-loop architecture, circular ASISR (CASISR), to improve the capability of image reconstruction for the advanced pretrained open-loop ASISR models. The proposed CASISR is described by a mathematical nonlinear loop equation. The reasonability and stability of the proposed CASISR is respectively proven by conditional probability theory and Taylor series approximation. The key innovations of this paper are listed as follows:

- A closed-loop architecture CASISR including the pretrained ASISR and DG models;
- A negative feedback mechanism based on the difference between the LR images;
- A mathematical nonlinear loop equation to describe the closed-loop architecture CASISR;
- A mathematical proof of the reasonability of the CASISR by conditional probability;
- A mathematical proof of the stability of the CASISR by Taylor series expansion;

- A sophisticated comparison strategy incorporating the absolute difference images;
- A preeminent capability of the CASISR for the images with drastically varied edges.

It should be mentioned that this paper is an extension of our previous work [8-11] on circular generalized image restoration, including circular medical image SR at integral scale factors via negative feedback based on the difference between the SR images [8], circular medical signal SR at integral scale factors [9], circular medical image compressive sensing [10], and circular image compression [11]. Compared with our previous work, this paper focuses more on fractional scale factors, forms negative feedback based on the difference between the LR images, proves the reasonability for the first time, and updates the mathematical proof of the stability.

The rest of this paper is arranged as follows. Section 2 summarizes the related work on the ASISR, section 3 constructs the theoretical foundation of the CASISR, section 4 evaluates the performance of the CASISR, and section 5 draws some valuable conclusions and prospects future research directions.

## 2 Related-Work

The ASISR preliminarily attempted some core ideas of the proposed CASISR, such as degradation model, closed-loop structure, testing-phase operations, and Taylor series expansion [12-16].

The ASISR considered the image degradation model in the open-loop framework, but not in the closed-loop framework [12]. For example, Wan Wenbo et al adaptively resolved the problem of image degradation perception for different scale factors and local pixel positions through image refinement network and SR encoding guidance module [12]. The proposed CASISR will show that the closed-loop structure with the degradation model can efficiently reinforce the GP.

The ASISR also considered the closed-loop architecture [13-15]. For instance, Chen Honggang et al. proposed a self-supervised cycle-consistent learning method with two network modules, a SR network and an inverse DG network, and two training cycles, LR-SR-LR and SR-LR-SR [13]. They did not need the LR-SR training pairs, but only needed the unpaired LR training samples. However, they employed an open-loop architecture in the testing phase and cannot completely take advantage of the testing datasets. Mamba model utilized an interior closed-loop architecture. Yan Jin et al integrated Mamba model with fast Fourier convolution blocks for the ASISR [14]. Reinforcement learning leveraged a closed-loop architecture. Fang Jing et al presented a deep reinforcement learning based adaptive scale-aware feature division mechanism for the ASISR [15]. However, it is a pity that the pretrained Mamba and reinforcement learning models cannot update the close-loop parameters. The proposed CASISR will demonstrate that the closed-loop architecture in both training and testing phases with changeable close-loop parameters can competently intensify the GP.

The ASISR further considered the intervention in the testing stage [16]. For example, Akita Kazutoshi et al raised a cost-and-quality control at the testing time [16]. They controlled the computational cost and SR quality in a single model by adjusting the total number of regressions for the recurrent neural network at the testing time. The proposed ASISR will optimally upgrade the GP at the testing time.

The ASISR finally consider Taylor series expansion to achieve high-definition SR images [17]. For instance, Wang Shibo et al brought forward a Taylor series expansion-based texture and edge-preserving interpolation approach for the ASISR [17]. They utilized the first-order Taylor series expansion of neighbor pixels to obtain the high-quality SR images. The proposed CASISR will wield the first-order Taylor series approximation to prove the stability.

The ASISR adopted explicit neural representation (ENR) and implicit neural representation (INR) for the ASISR [18-28]. The ENR is an explicitly defined, continuous and analytical function derived from a parameterized neural network. The INR is an implicitly defined, continuous, differentiable function parameterized by a neural network. Pang Kaifeng et al proposed an analytical function-based ENR for the fast medical ASISR [18]. Luo Laigan et al presented bidirectional scale-aware module and ENR spatial pyramid up-sampling module for the video frame ASISR [19]. Lyu Jun et al raised a diffusion model prior-based INR for the cardiac cine magnetic resonance imaging (MRI) ASISR [20]. Wei Jinbao et al put forward a dynamic INR for the multi-contrast MRI ASISR [21]. Tu Xinyue et al synergized a frequency domain and texture-aware INR for the MRI ASISR [22]. Wu Qing et al brought forward an INR with implicit neural voxel function for the 3D MRI ASISR [23]. Tsai Yi-Ting proposed an INR with implicit image function modeling for the face image ASISR [24]. Wen Caizhe et al presented an INR with enhanced implicit function-based neural network for the ASISR [25]. Wang Shuangxing et al raised a diffusion MRI ASISR including a diffusion direction-specific image feature extraction module and a physics-driven INR and reconstruction module [26]. Chen Chuan et al put forward an INR with local Gaussian ensemble for the ASISR [27]. Chen Guochao et al brought forward a hyperspectral image ASISR which naturally incorporates a spatial prior in both feature extraction and local implicit image function [28].

The ASISR also adopted various deep learning models, such as densely-connected neural network, autoencoder neural network, residual neural network, convolutional neural network, generative adversarial network, transformer model, diffusion model, Mamba model, binary neural network, lightweight neural network, stacked neural network, meta learning, transfer learning, reinforcement learning, and so forth [29-48]. For example, Maciel Corbin et al proposed a frequency-domain informed autoencoder neural network for the 3D whole-heart MRI ASISR [29]. Rong Yi et al presented a residual dense neural network with feature aggregation for the logging image ASISR [30]. Gurrola-Ramos Javier et al raised a residual channel-spatial attention neural network for the ASISR [31]. Fu Ying et al put forward a residual scale attention network for the ASISR [32]. Shen Jialiang et al brought forward a deep convolutional neural network for the ASISR [33]. Li Guangping et al come up with a convolution based multi-scale cross-fusion network for the ASISR [34]. Zhao Yaoqian et al proposed convolution-based scale-aware local feature adaptation module and local feature adaptation up-sampling module for the ASISR [35]. He Zhi et al presented a residual block-based feature learning module and convolution-based arbitrary upscale module for the satellite video frame ASISR [36]. Zhu Jin et al raised a medical image ASISR based on generative adversarial network, meta learning, and transfer learning [37]. Zhu Jinchen et al brought forward a multi-scale implicit transformer model with re-parameterization for the ASISR [38]. Jiang Shuguo et al come up with a spectral-query transformer for the hyperspectral image ASISR [39]. Kim Min Hyuk et al proposed a transformer model with memory-efficient discrete cosine transformation domain weight modulation for the ASISR [40]. Zhao Ming et al presented a permuted cross-attention transformer for the multi-contrast MRI ASISR [41]. Liu Lanqing et al raised an implicit multi-contrast diffusion model for the MRI ASISR [42]. Han Zhitao et al put forward a diffusion model for the brain MRI ASISR [43]. Hwang Inje et al combined diffusion model with implicit image representation for the ASISR [44]. Cui Jizhou et al brought forward a coarse-to-fine meta-diffusion model for the hyperspectral image ASISR [45]. Wang Yufeng et al come up with binary lightweight neural networks for the remote sensing image ASISR [46]. Jia Haoran et al proposed a lightweight ASISR via texture-aware deformation up-sampling module and scale-aware image refinement module [47]. He Gang presented stacking multiple priors-guided processing neural blocks for the video frame ASISR [48].

The ASISR further adopted some high-level architectures of deep neural networks, such as dual-branch and bilateralism [49-54]. For instance, Duan Minghong et al proposed an efficient implicit self-texture enhancement-based dual-branch framework including feature aggregation and texture learning branches for the histopathology image ASISR [49]. Li Guangping et al presented an enhanced dual branches network including pixel and scale feature branches for the ASISR [50]. Zhang Menglei et al raised a bilateral up-sampling network including space and range up-sampling branches for the ASISR [51]. Fang Jing et al designed a two-branch edge reinforcement module for the satellite image ASISR [52]. Truong Anh Minhet et al put forward depth completion and generation branch and weight prediction branch for the indoor image ASISR [53]. An Tai et al combined discrete representation-based dense prediction branch and continuous representation-based resolution-specific refinement branch for the satellite image ASISR [54].

The ASISR can be divided into three categories: integrated single-scale model, unitary multi-scale model, and unitary arbitrary-scale model. The ASISR exploits different up-scaling methods, such as interpolation, learnable adaptive upscaling, neural representation-based upscaling, etc. The ASISR was suitable for diverse image categories, such as natural images, facial images, video frames, indoor images, medical images, magnetic resonance images, histopathology images, remote sensing images, hyperspectral images, satellite images, depth images, etc.

All in all, this paper proposes an implicit closed-loop infrastructure, CASISR, to boost the GP of the cutting-edge pretrained ASISR models in the testing phase for the natural images.

# 3 Theory

## 3.1 Mathematical Notations

The mathematical notations and abbreviations utilized in this paper are enumerated in Tab. 1 below.

**Tab. 1. Mathematical notations and abbreviations.**

| Notations/Abbreviations | Meanings |
|---|---|
| $\mathbf{a}_l$ / $\mathbf{a}_d$ | the original / generated LR image |
| $\mathbf{a}_s$ / $\mathbf{a}_g$ | the generated / ground-truth SR image |
| $\mathbf{a}_e$ / $\mathbf{a}_p$ | the LR error / the pre-position image |
| $\mathbf{a}_m$ / $\mathbf{a}_p(0)$ | the output of multiplier / the initial value of $\mathbf{a}_p$ |
| $a_{s;j}$ / $a_{p;j}(t)$ | the j-th element of $\mathbf{a}_s$ / $\mathbf{a}_p(t)$ |
| $\boldsymbol{\theta}_{SR}$ | the parameter vector of SR neural network |
| $\mathbf{o}(\bullet)$ / $\mathbf{e}(t)$ | the 2-order and higher-order terms / the steady-state error |
| $\boldsymbol{\varepsilon}$ / $\boldsymbol{\delta}$ | the small constant vectors |
| $\boldsymbol{\Gamma}_{PD}$ / $\boldsymbol{\Gamma}_{DP}$ | the matrices of Taylor series coefficients for 1-order term |
| $\Gamma_{PD;ij}$ / $\Gamma_{DP;ij}$ | the i-th and j-th element of $\boldsymbol{\Gamma}_{PD}$ / $\boldsymbol{\Gamma}_{DP}$ |
| $\mathbf{I}$ /$\boldsymbol{\Lambda}$ | the identity matrix / the matrix of multiplication coefficients |
| SR(•) / PP(•) | the SR / pre-position module |

| PT(•) / DG(•) | the pretrained / degradation module |
|---|---|
| L(•) | the loss function |
| PD(•) / DP(•) | the nonlinear functions |
| $PD_i$(•) / $DP_i$(•) | the i-th element of PD(•) / DP(•) |
| $P(\mathbf{a}_p \mid \mathbf{a}_e)$ | the conditional probability of successful prediction for PP module |
| $P(\mathbf{a}_s \mid \mathbf{a}_p)$ | the conditional probability of successful prediction for PT module |
| $P(\mathbf{a}_d \mid \mathbf{a}_s)$ | the conditional probability of successful prediction for DG module |
| m / n / M / N | the height / width of LR / SR image |
| d / D | the dimension of LR / SR sample |
| H / T / Z | the total number of image channels / iterations / images |
| $f_h$ / $f_w$ / f | the height / width SR scale factor / the SR scale factor |
| w | the average conditional probability of successful prediction |
| t / Δt / F / λ | the time / the time interval / Frobenius-norm / constant coefficient |
| A / B / C | the coefficients of the inequality |
| LR / SR / ISR | low-resolution / super-resolution / image SR |
| ASISR / CASISR | Arbitrary-scale ISR / circular ASISR |
| GP / MRI | generalization performance / magnetic resonance imaging |
| ENR / INR | explicit / implicit neural representation |

### 3.2 Recall of Deep Learning-Based ASISR

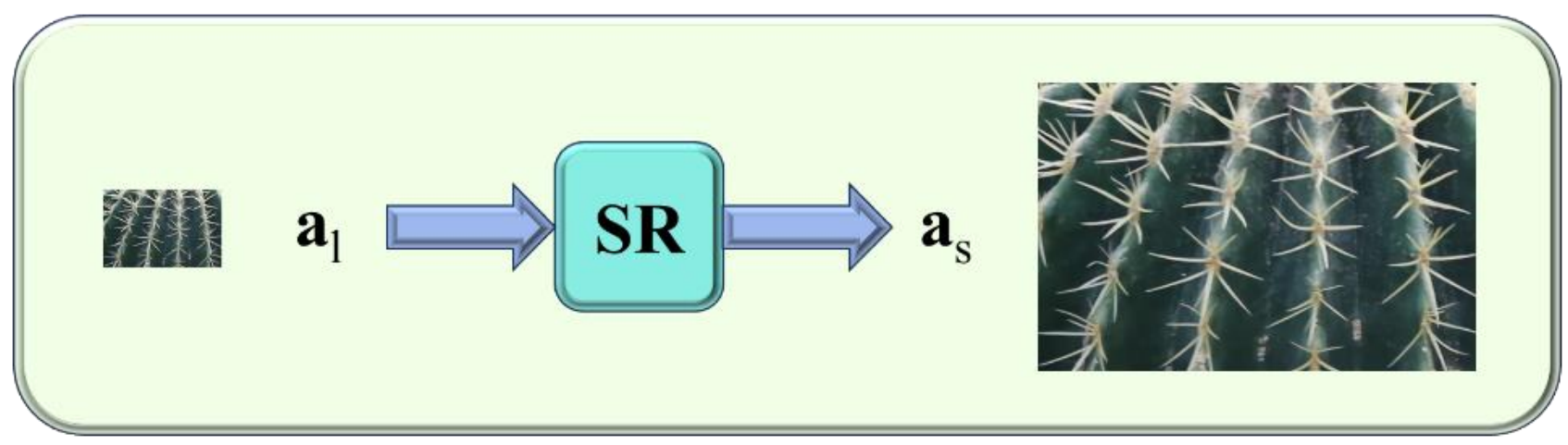

**Fig. 1. The framework of deep learning-based ASISR.**

The framework of deep learning-based ASISR is illustrated in Fig. 1. $\mathbf{a}_l$ is the original LR image and $\mathbf{a}_s$ is the SR image. The SR neural network reconstructs $\mathbf{a}_s$ from $\mathbf{a}_l$. The deep learning-based ASISR can be formulated as the following mathematical equations:

$$\mathbf{a}_s = SR(\mathbf{a}_l)$$
$$d = m \times n \times H, D = M \times N \times H, f_h = \frac{M}{m}, f_w = \frac{N}{n}, f = f_h = f_w, (1)$$
$$s.t.\ m,n,d,H,M,N,D \in Z^+; f_h, f_w, f \in R; \mathbf{a}_l \in R^d; \mathbf{a}_s \in R^D$$

where: function SR(•) is the SR neural network, m is the height of the LR image, n is the width of the LR image, d is the dimension of the LR sample, M is the height of the SR image, N is the width of the SR image, H is the total number of image channels, D is the dimension of the SR image, $f_h$ is the height SR scale factor, $f_w$ is the width SR scale factor, and f is the SR scale factor. It is reasonable to suppose that f, $f_h$, and $f_w$ share the same value which can be any integral or fractional value.

The deep learning-based ASISR can be solved via the following minimization problem:

$$\boldsymbol{\theta}_{SR}^{*} = \arg\min_{\boldsymbol{\theta}_{SR}} L\left(\mathbf{a}_s, \mathbf{a}_g\right), \text{s.t. } \mathbf{a}_s = SR\left(\mathbf{a}_l\right), (2)$$

where: $\boldsymbol{\theta}_{SR}$ denotes the parameter vector of the SR neural network; L denotes the loss function of $\mathbf{a}_s$ and $\mathbf{a}_g$, such as mean square error, cross entropy, total variation, sparse representation, perception loss, adversarial loss, and so on; $\mathbf{a}_g$ denotes the ground-truth (GT) SR image; $\mathbf{a}_l$ and $\mathbf{a}_g$ denote the training image pair.

### 3.3 Proposed Architecture of CASISR

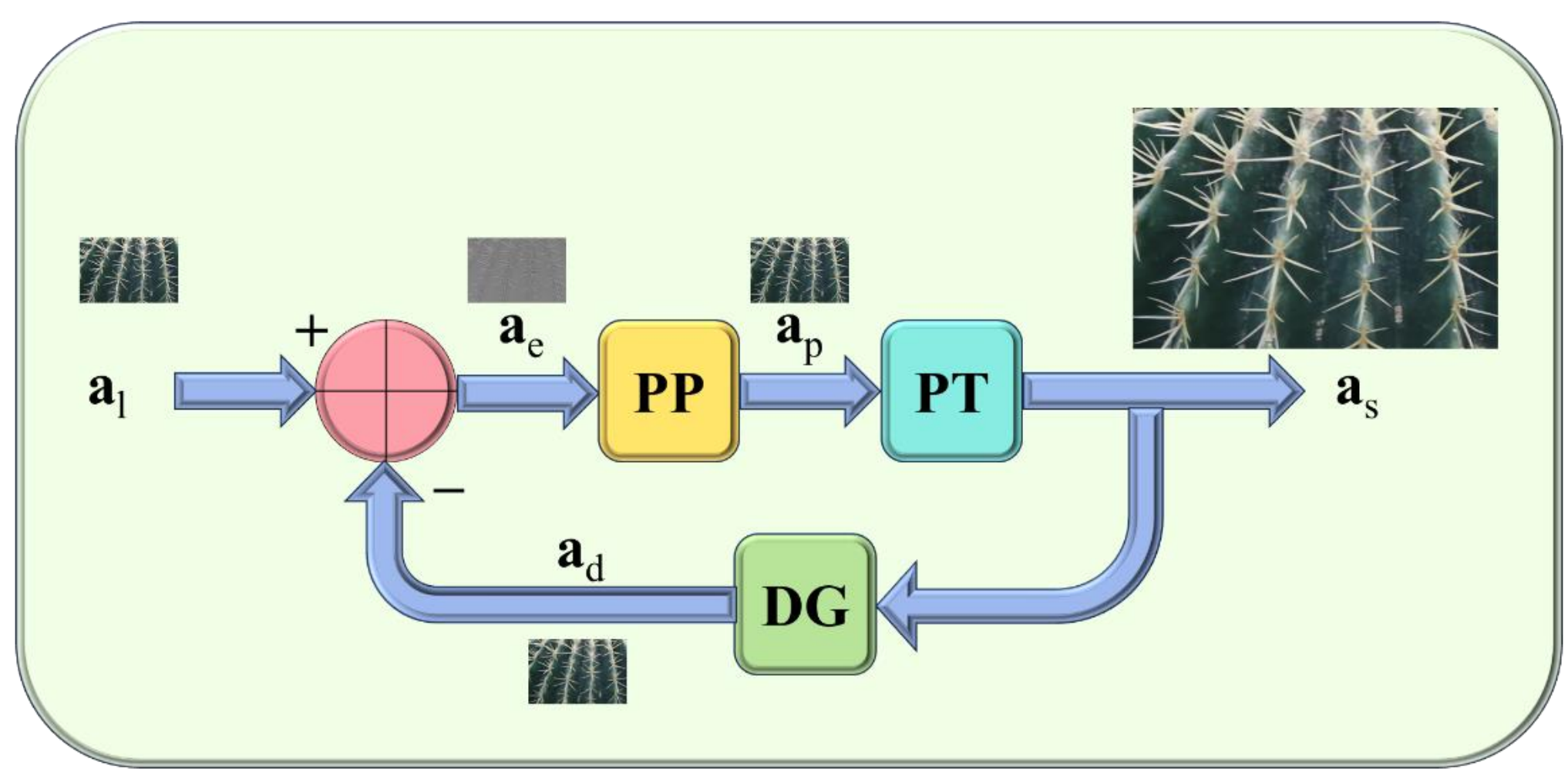

**Fig. 2. The proposed architecture of the CASISR.**

The proposed architecture of the CASISR is displayed in Fig. 2. $\mathbf{a}_d$ is the generated LR image, $\mathbf{a}_e$ is the LR error image, and $\mathbf{a}_p$ is the LR pre-position (PP) image. $\mathbf{a}_l$ is the architecture input and $\mathbf{a}_s$ is the architecture output. The adder module introduces negative feedback, compares $\mathbf{a}_l$ with $\mathbf{a}_d$, and produces $\mathbf{a}_e$. The negative feedback mechanism relies on the differences between the LR images. The PP module transforms $\mathbf{a}_e$ into $\mathbf{a}_p$. The PP module can be either a traditional proportional-integral (PI) control unit, a modern fuzzy control unit, or a modern adaptive control unit. The pretrained (PT) module reconstructs $\mathbf{a}_s$ from $\mathbf{a}_p$. The PT module can leverage any existing advanced pretrained open-loop SR modules. The DG module degrades $\mathbf{a}_s$ to $\mathbf{a}_d$. The DG module is known down-sampling in the classical ASISR, is assumed down-sampling with additive white noise in the blind ASISR, and is learnable in the real-world ASISR. Nearest-neighbor interpolation, bilinear interpolation, bicubic interpolation, and spline interpolation can be exploited to simulate the down-sampling process of the DG module in the classical and blind ASISR.

According to the automatic control theory, when the closed-loop system is steady, $\mathbf{a}_e$ is close to zero, $\mathbf{a}_d$ is close to $\mathbf{a}_l$, $\mathbf{a}_s$ is close to the original SR image or GT high resolution image $\mathbf{a}_g$, and perfect SR image is obtained.

### 3.4 Reasonability of CASISR

Reasonability means that the performance of the closed-loop architecture should surpass that of the open-loop architecture. Otherwise, the closed-loop architecture is unreasonable. The proof of the reasonability is based on the conditional probabilities which are shown in Fig. 3. P($\mathbf{a}_p$|$\mathbf{a}_e$) is the conditional probability of successful prediction for the PP module, P($\mathbf{a}_s$|$\mathbf{a}_p$) is the conditional probability of successful prediction for the PT module, and P($\mathbf{a}_d$|$\mathbf{a}_s$) is the conditional probability of successful prediction for the DG module. The aforementioned conditional probabilities can be described by the following mathematical equations:

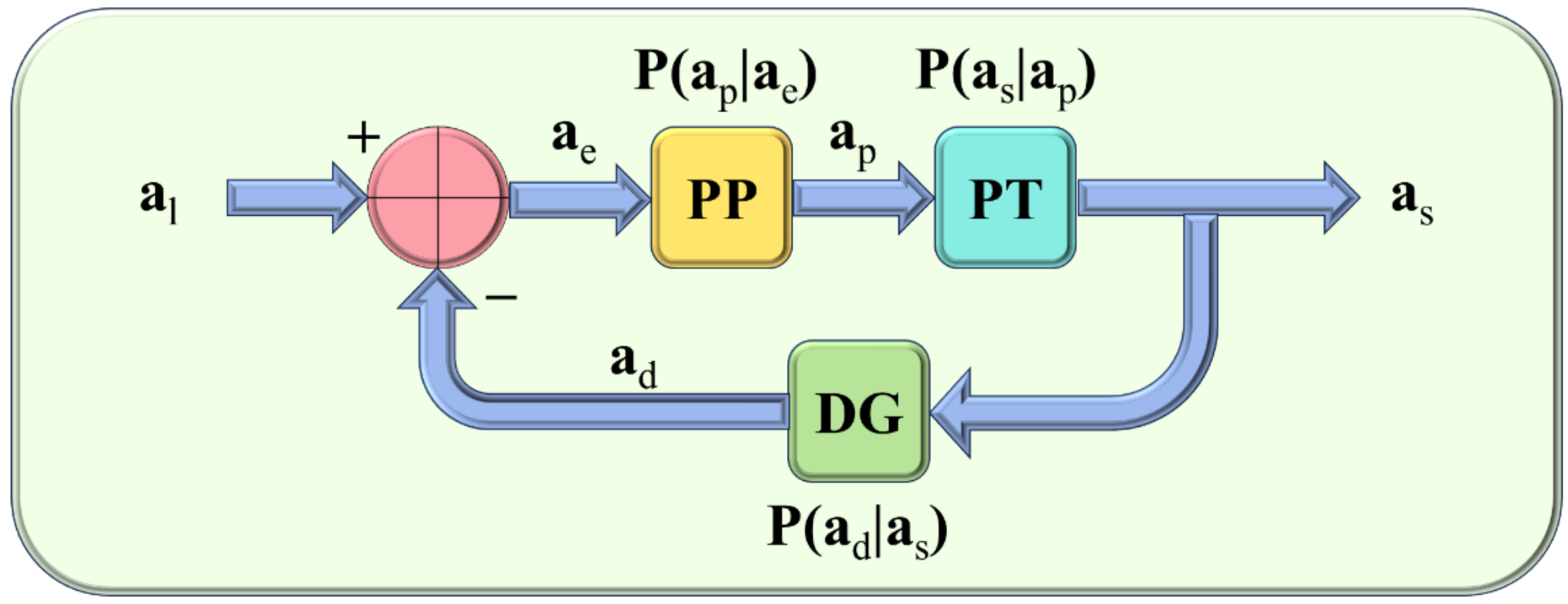

**Fig. 3. The reasonability of CASISR.**

$$\begin{cases} P(\mathbf{a}_p|\mathbf{a}_e) = d \cdot w^d \\ P(\mathbf{a}_s|\mathbf{a}_p) = d \cdot w^D \\ P(\mathbf{a}_d|\mathbf{a}_s) = D \cdot w^d \end{cases} ,(3)$$
$$\text{s.t. } \mathbf{a}_e, \mathbf{a}_p, \mathbf{a}_d \in R^d; \mathbf{a}_s \in R^D; w \in [0,1)$$

where w is the average conditional probability of successful prediction from one input pixel to one output pixel. Each conditional probability equals to the product of the input dimension and the output dimension-th power of w.

The following inequality can be achieved under the condition that d is less than D:

$$\begin{cases} P(\mathbf{a}_p|\mathbf{a}_e) > P(\mathbf{a}_s|\mathbf{a}_p) \\ P(\mathbf{a}_d|\mathbf{a}_s) > P(\mathbf{a}_s|\mathbf{a}_p) \end{cases} .(4)$$
$$\text{s.t. } d < D$$

P($\mathbf{a}_p$|$\mathbf{a}_e$) is greater than P($\mathbf{a}_s$|$\mathbf{a}_p$) and P($\mathbf{a}_d$|$\mathbf{a}_s$) is also greater that P($\mathbf{a}_s$|$\mathbf{a}_p$). Hence, it is reasonable to form the closed-loop framework. By this point, the reasonability of the CASISR is demonstrated.

### 3.5 Loop Equation of CASISR

The loop equation of the CASISR can be depicted by the following mathematical formulars:

$$\begin{cases} \mathbf{a}_e = \mathbf{a}_l - \mathbf{a}_d \\ \mathbf{a}_p = \mathrm{PP}(\mathbf{a}_e) \\ \mathbf{a}_s = \mathrm{PT}(\mathbf{a}_p) \\ \mathbf{a}_d = \mathrm{DG}(\mathbf{a}_s) \end{cases} ,(5)$$

$$\Rightarrow \mathbf{a}_s = \mathrm{PT}\left(\mathrm{PP}\left(\mathbf{a}_l - \mathrm{DG}(\mathbf{a}_s)\right)\right), \text{s.t. } \mathbf{a}_l, \mathbf{a}_d, \mathbf{a}_e, \mathbf{a}_p \in \mathrm{R}^d; \mathbf{a}_s \in \mathrm{R}^D$$

The loop equation represents a nonlinear implicit equation about $\mathbf{a}_s$.

### 3.6 Stability of CASISR

The stability of the CASISR refers to the capability of the closed-loop system to operate normally with a steady-state error approaching zero. Otherwise, it is unstable.

The aforementioned nonlinear loop equation can be rephrased as follows:

$$\mathbf{a}_s = \mathrm{PD}(\mathbf{a}_s) = \mathrm{PT}\left(\mathrm{PP}\left(\mathbf{a}_l - \mathrm{DG}(\mathbf{a}_s)\right)\right),(6)$$

where $\mathrm{PD}(\mathbf{s}_s)$ is a nonlinear function.

Assuming that $\mathrm{PD}(\mathbf{a}_s)$ has the partial derivatives at $\mathbf{a}_g$, its Taylor series expansion at $\mathbf{a}_g$ is:

$$\mathrm{PD}(\mathbf{a}_s) = \mathrm{PD}(\mathbf{a}_g) + \mathbf{\Gamma}_{PD}(\mathbf{a}_s - \mathbf{a}_g) + \mathbf{o}(\mathbf{a}_s)$$
$$\mathbf{\Gamma}_{PD} = \begin{bmatrix} \Gamma_{PD;11} & \Gamma_{PD;12} & \cdots & \Gamma_{PD;1D} \\ \Gamma_{PD;21} & \Gamma_{PD;22} & \cdots & \Gamma_{PD;2D} \\ \vdots & \vdots & \ddots & \vdots \\ \Gamma_{PD;D1} & \Gamma_{PD;D2} & \cdots & \Gamma_{PD;DD} \end{bmatrix} \in \mathrm{R}^{D\times D}, \Gamma_{PD;ij} = \left.\frac{\partial \mathrm{PD}_i(\mathbf{a}_s)}{\partial a_{s;j}}\right|_{\mathbf{a}_s=\mathbf{a}_g}; i,j = 0,1,\cdots,D ,(7)$$

where: $\mathbf{\Gamma}_{PD}$ denotes the matrix of Taylor series coefficients for the 1-order term, $\Gamma_{PD;ij}$ denotes the i-th and j-th element of $\mathbf{\Gamma}_{PD}$, $\mathbf{o}(\mathbf{a}_s)$ denotes the 2-order and higher-order terms, $\mathrm{PD}_i(\mathbf{a}_s)$ denotes the i-th element of $\mathrm{PD}(\mathbf{a}_s)$, and $a_{s;j}$ denotes the j-th element of $\mathbf{a}_s$.

Discarding the 2-order and higher-order terms of $\mathrm{PD}(\mathbf{a}_s)$, its locally linear approximation is:

$$\mathrm{PD}(\mathbf{a}_s) \approx \mathrm{PD}(\mathbf{a}_g) + \mathbf{\Gamma}_{PD}(\mathbf{a}_s - \mathbf{a}_g) = \mathbf{\Gamma}_{PD}(\mathbf{a}_s - \mathbf{a}_g)$$
$$\text{s.t. } \mathrm{DG}(\mathbf{a}_g) = \mathbf{a}_l; \qquad .(8)$$
$$\mathrm{PD}(\mathbf{a}_g) = \mathrm{PT}\left(\mathrm{PP}\left(\mathbf{a}_l - \mathrm{DG}(\mathbf{a}_g)\right)\right) = \mathrm{PT}\left(\mathrm{PP}(\mathbf{a}_l - \mathbf{a}_l)\right) = \mathrm{PT}\left(\mathrm{PP}(0)\right) = 0$$

The locally linear approximation of loop equation at $\mathbf{a}_g$ is:

$$\mathbf{a}_s = \mathrm{PD}(\mathbf{a}_s) \approx \mathbf{\Gamma}_{PD}(\mathbf{a}_s - \mathbf{a}_g).(9)$$

SR image $\mathbf{a}_s$ can be resolved as follows:

$$\mathbf{a}_s = \left(\mathbf{\Gamma}_{PD} - \mathbf{I}\right)^{-1} \mathbf{\Gamma}_{PD}\mathbf{a}_g ,(10)$$

where $\mathbf{I}$ represents the identity matrix.

The necessary and sufficient condition for errorless reconstruction is expressed as follows:

$$\left(\mathbf{\Gamma}_{PD} - \mathbf{I}\right)^{-1} \mathbf{\Gamma}_{PD} = \mathbf{I} \Leftrightarrow \mathbf{a}_s = \mathbf{a}_g .(11)$$

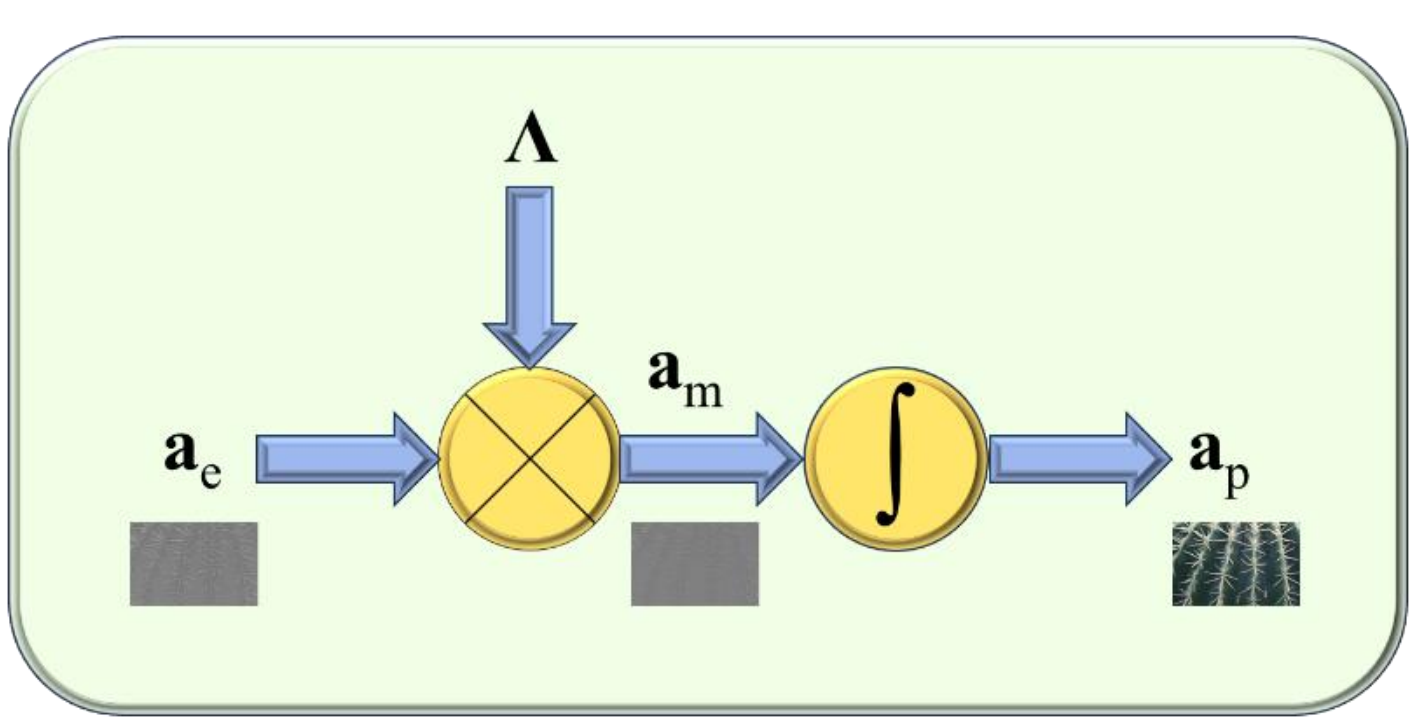

**Fig. 4. Traditional PI control unit.**

The aforementioned necessary and sufficient condition is ideal and cannot be acquired in practice. To derive a practical condition, it is essential to simplify the PP module. It is rational to assume that the PP module adopts a traditional PI control structure in Fig. 4, where $\mathbf{\Lambda}$ is the matrix of multiplication coefficients and $\mathbf{a}_m$ is the output of multiplier.

The traditional PI control unit can be formulated by the following mathematical equations:

$$\begin{aligned} \mathbf{a}_m(t) &= \mathbf{\Lambda}(t)\mathbf{a}_e(t), \mathbf{\Lambda}(t) \in R^{d\times d} \\ \mathbf{a}_p(t) &= \int_0^t \mathbf{a}_m(t)dt + \mathbf{a}_p(0) \end{aligned} ,(12)$$

where: t is the time and $\mathbf{a}_p(0)$ is the initial value of integral which can be chosen as $\mathbf{a}_l$, zero, or random value.

According to Figure 4, the nonlinear loop equation can be rewritten as follows:

$$\mathbf{a}_d(t) = DG\left(\mathbf{a}_s(t)\right) = DG\left(PT\left(\mathbf{a}_p(t)\right)\right) = DP\left(\mathbf{a}_p(t)\right),(13)$$

where $DP(\mathbf{a}_p(t))$ denotes a nonlinear function.

Supposing that $DP(\mathbf{a}_p(t))$ holds the partial derivatives at $\mathbf{a}_p(0)$, its Taylor series expansion at $\mathbf{a}_p(0)$ is:

$$\mathrm{DP}\left(\mathbf{a}_{\mathrm{p}}(\mathrm{t})\right)=\mathrm{DP}\left(\mathbf{a}_{\mathrm{p}}(0)\right)+\boldsymbol{\Gamma}_{\mathrm{DP}}\left(\mathbf{a}_{\mathrm{p}}(\mathrm{t})-\mathbf{a}_{\mathrm{p}}(0)\right)+\mathbf{o}\left(\mathbf{a}_{\mathrm{p}}(\mathrm{t})\right)$$

$$\boldsymbol{\Gamma}_{\mathrm{DP}}=\begin{bmatrix}\Gamma_{\mathrm{DP};11} & \Gamma_{\mathrm{DP};12} & \cdots & \Gamma_{\mathrm{DP};1\mathrm{d}}\\ \Gamma_{\mathrm{DP};21} & \Gamma_{\mathrm{DP};22} & \cdots & \Gamma_{\mathrm{DP};2\mathrm{d}}\\ \vdots & \vdots & \ddots & \vdots\\ \Gamma_{\mathrm{DP};\mathrm{d}1} & \Gamma_{\mathrm{DP};\mathrm{d}2} & \cdots & \Gamma_{\mathrm{DP};\mathrm{dd}}\end{bmatrix}\in\mathrm{R}^{\mathrm{d}\times\mathrm{d}} \qquad ,(14)$$

$$\Gamma_{\mathrm{DP};\mathrm{ij}}=\left.\frac{\partial\mathrm{DP}_{\mathrm{i}}\left(\mathbf{a}_{\mathrm{p}}(\mathrm{t})\right)}{\partial\mathrm{a}_{\mathrm{p};\mathrm{j}}(\mathrm{t})}\right|_{\mathbf{a}_{\mathrm{p}}(\mathrm{t})=\mathbf{a}_{\mathrm{p}}(0)} \quad ;\mathrm{i},\mathrm{j}=0,1,\cdots,\mathrm{d}$$

where: $\boldsymbol{\Gamma}_{DP}$ represents the matrix of Taylor series coefficients for the 1-order term, $\Gamma_{DP;ij}$ represents the i-th and j-th element of $\boldsymbol{\Gamma}_{DP}$, $\mathbf{o}(\mathbf{a}_p(t))$ represents the 2-order and higher-order terms, $DP_i(\mathbf{a}_p(t))$ represents the i-th element of $DP(\mathbf{a}_p(t))$, and $a_{p;j}(t)$ represents the j-th element of $\mathbf{a}_p(t)$.

Discarding the 2-order and higher-order terms of $DP(\mathbf{a}_p(t))$, its locally linear approximation is:

$$\mathrm{DP}\left(\mathbf{a}_{\mathrm{p}}(\mathrm{t})\right)\approx\mathrm{DP}\left(\mathbf{a}_{\mathrm{p}}(0)\right)+\boldsymbol{\Gamma}_{\mathrm{DP}}\left(\mathbf{a}_{\mathrm{p}}(\mathrm{t})-\mathbf{a}_{\mathrm{p}}(0)\right).(15)$$

The nonlinear loop equation can be linearly approximated as follows:

$$\mathbf{a}_{\mathrm{d}}(\mathrm{t})\approx\mathrm{DP}\left(\mathbf{a}_{\mathrm{p}}(0)\right)+\boldsymbol{\Gamma}_{\mathrm{DP}}\left(\mathbf{a}_{\mathrm{p}}(\mathrm{t})-\mathbf{a}_{\mathrm{p}}(0)\right).(16)$$

The nonlinear loop equation can further be rewritten as follows:

$$\mathbf{a}_{\mathrm{d}}(\mathrm{t})\approx\mathrm{DP}\left(\mathbf{a}_{\mathrm{p}}(0)\right)+\boldsymbol{\Gamma}_{\mathrm{DP}}\left(\int_0^{\mathrm{t}}\boldsymbol{\Lambda}(\mathrm{t})\mathbf{a}_{\mathrm{e}}(\mathrm{t})\mathrm{dt}\right).(17)$$

Supposing that time interval $\Delta t$ is greater than zero, $\mathbf{a}_d(t+\Delta t)$ can be gained as follows:

$$\mathbf{a}_{\mathrm{d}}(\mathrm{t}+\Delta\mathrm{t})=\mathbf{a}_{\mathrm{d}}(\mathrm{t})+\boldsymbol{\Gamma}_{\mathrm{DP}}\left(\int_{\mathrm{t}}^{\mathrm{t}+\Delta\mathrm{t}}\boldsymbol{\Lambda}(\mathrm{t})\mathbf{a}_{\mathrm{e}}(\mathrm{t})\mathrm{dt}\right).(18)$$

Subtracting both sides of aforementioned mathematical equation from $\mathbf{a}_l$, the following mathematical equation can be obtained:

$$\mathbf{a}_{\mathrm{l}}-\mathbf{a}_{\mathrm{d}}(\mathrm{t}+\Delta\mathrm{t})=\mathbf{a}_{\mathrm{l}}-\mathbf{a}_{\mathrm{d}}(\mathrm{t})-\boldsymbol{\Gamma}_{\mathrm{DP}}\left(\int_{\mathrm{t}}^{\mathrm{t}+\Delta\mathrm{t}}\boldsymbol{\Lambda}(\mathrm{t})\mathbf{a}_{\mathrm{e}}(\mathrm{t})\mathrm{dt}\right).(19)$$

According to the definition of error signal $\mathbf{a}_e$, the following mathematical formular can be attained:

$$\mathbf{a}_e(t+\Delta t) = \mathbf{a}_e(t) - \boldsymbol{\Gamma}_{DP}\left(\int_t^{t+\Delta t} \boldsymbol{\Lambda}(t)\mathbf{a}_e(t)dt\right).(20)$$

Giving that Δt is approximate to zero, the following mathematical expressions can be achieved:

$$\mathbf{a}_e(t+\Delta t) \approx \mathbf{a}_e(t) - \boldsymbol{\Gamma}_{DP}\boldsymbol{\Lambda}(t)\mathbf{a}_e(t)\Delta t = \left(\mathbf{I} - \boldsymbol{\Gamma}_{DP}\boldsymbol{\Lambda}(t)\Delta t\right)\mathbf{a}_e(t), \text{s.t. } \Delta t \to 0\ .(21)$$

By taking the norm of the both sides of the aforementioned mathematical expressions, the following inequality can be derived under certain conditions:

$$\left\|\mathbf{a}_e(t+\Delta t)\right\|_2 \le \left\|\mathbf{I} - \boldsymbol{\Gamma}_{DP}\boldsymbol{\Lambda}(t)\Delta t\right\|_F \left\|\mathbf{a}_e(t)\right\|_2, \text{s.t. } \Delta t \to 0\ ,(22)$$

where: subscript 2 denotes 2-norm and subscript F denotes Frobenius-norm.

The condition that satisfies the aforementioned inequality is the following one:

$$\left\|\mathbf{I} - \Gamma_{DP}\boldsymbol{\Lambda}(t)\Delta t\right\|_F < 1\ .(23)$$

Suitable $\boldsymbol{\Lambda}$ can always be chosen to meet the aforementioned condition. If $\boldsymbol{\Lambda}$ is proportional to an identity matrix, the following equations can be obtained:

$$\boldsymbol{\Lambda}(t) = \lambda\mathbf{I}\ ,(24)$$

where λ is a coefficient.

λ can be selected according to the following inequality:

$$\begin{aligned}
&\left\|\mathbf{I} - \boldsymbol{\Gamma}_{DP}\boldsymbol{\Lambda}(t)\Delta t\right\|_F = \left\|\mathbf{I} - \boldsymbol{\Gamma}_{DP}\cdot\lambda\mathbf{I}\cdot\Delta t\right\|_F = \left\|\mathbf{I} - \lambda\Delta t\cdot\boldsymbol{\Gamma}_{DP}\right\|_F < 1 \\
&\Rightarrow \sqrt{\sum_{i=1}^{d}\left(1 - 2\lambda\Delta t\cdot\Gamma_{DP;ii}\right) + \sum_{i=1}^{d}\sum_{j=1}^{d}\lambda^2\Delta^2 t\cdot\Gamma_{DP;ij}^2} < 1 \\
&\Rightarrow \sum_{i=1}^{d}\left(1 - 2\lambda\Delta t\cdot\Gamma_{DP;ii}\right) + \sum_{i=1}^{d}\sum_{j=1}^{d}\lambda^2\Delta^2 t\cdot\Gamma_{DP;ij}^2 < 1 \\
&\Rightarrow \left(\sum_{i=1}^{d}\sum_{j=1}^{d}\Delta^2 t\cdot\Gamma_{DP;ij}^2\right)\lambda^2 + \left(\sum_{i=1}^{d}(-2\Delta t)\cdot\Gamma_{DP;ii}\right)\lambda + \left(\sum_{i=1}^{d}1\right) < 0\ ,(25) \\
&\Rightarrow A\lambda^2 - B\lambda + C < 0 \\
&\Rightarrow \frac{-B - \sqrt{B^2 - 4AC}}{2A} < \lambda < \frac{-B + \sqrt{B^2 - 4AC}}{2A} \\
&\text{s.t. } A = \sum_{i=1}^{d}\sum_{j=1}^{d}\Delta^2 t\cdot\Gamma_{DP;ij}^2, B = \sum_{i=1}^{d}(-2\Delta t)\cdot\Gamma_{DP;ii}, C = \sum_{i=1}^{d}1 = d
\end{aligned}$$

where A, B, and C denote the coefficients of the inequality.

If the aforementioned condition is satisfied, the following inequality can be derived:

$$\left\|\mathbf{a}_e\left(t+\Delta t\right)\right\|_2 < \left\|\mathbf{a}_e\left(t\right)\right\|_2 .(26)$$

If t approaches infinite in the steady-state, the error image $\mathbf{a}_e$ approaches a small constant vector $\boldsymbol{\varepsilon}$:

$$\lim_{t\to\infty}\mathbf{a}_e\left(t\right)=\boldsymbol{\varepsilon},\ \boldsymbol{\varepsilon}\in R^d .(27)$$

Hence, $\mathbf{a}_d(t)$ is close to $\mathbf{a}_l$ in the steady-state:

$$\lim_{t\to\infty}\mathbf{a}_d\left(t\right)=\lim_{t\to\infty}\left(\mathbf{a}_l-\mathbf{a}_e\left(t\right)\right)=\mathbf{a}_l-\lim_{t\to\infty}\mathbf{a}_e\left(t\right)=\mathbf{a}_l-\boldsymbol{\varepsilon} .(28)$$

Furthermore, $\mathbf{a}(t)$ is approximate to $\mathbf{a}_g$ in the steady-state:

$$\begin{cases}\lim_{t\to\infty}\mathbf{a}_d\left(t\right)=\lim_{t\to\infty}DG\left(\mathbf{a}_s\left(t\right)\right)=DG\left(\lim_{t\to\infty}\mathbf{a}_s\left(t\right)\right)\\ \mathbf{a}_l=DG\left(\mathbf{a}_g\right)\end{cases}$$
$$\Rightarrow\lim_{t\to\infty}\mathbf{a}_s\left(t\right)=\mathbf{a}_g-\boldsymbol{\delta},\ \boldsymbol{\delta}\in R^D ,(29)$$

where $\boldsymbol{\delta}$ is a small constant vector.

Finally, the steady-state error $\mathbf{e}(t)$ is close to $\boldsymbol{\delta}$:

$$\begin{gathered}\mathbf{e}\left(t\right)=\mathbf{a}_g-\mathbf{a}_s\left(t\right)\\ \Rightarrow\lim_{t\to\infty}\mathbf{e}\left(t\right)=\lim_{t\to\infty}\left(\mathbf{a}_g-\mathbf{a}_s\left(t\right)\right)=\mathbf{a}_g-\lim_{t\to\infty}\mathbf{a}_s\left(t\right)=\boldsymbol{\delta}\end{gathered} .(30)$$

At this point, the stability of the CASISR is proven, and the steady-state error approaches a small constant vector.

### 3.7 Algorithm Description

The proposed CASISR algorithm is illustrated in Fig. 5, where T denotes the total number of iterations.

**Algorithm 1: CASISR.**

**Input:** $\mathbf{a_l}$
**Initialization:**
  $\mathbf{t = 1, T = constant, a_p(0) = a_l}$
**while t <= T**
  **compute** $\mathbf{a_s(t), a_d(t)}$, **and** $\mathbf{a_e(t)}$ **according to equation (5)**
  **compute** $\mathbf{\Gamma_{DP}}$ **according to equations (13) and (14)**
  **compute** $\boldsymbol{\lambda}$ **according to equation (25)**
  **compute** $\mathbf{a_p(t)}$ **according to equations (5), (12), and (24)**
  $\mathbf{t = t+1}$
**end while**
**Output:** $\mathbf{a_s(t)}$

**Fig. 5. Proposed algorithm of CASISR.**

# 4 Experiments

### 4.1 Experiment Design

The experimental section is designed to compare the proposed closed-loop CASISR algorithm and the classical pretrained open-loop ASISR algorithms in the performance of image reconstruction. For the proposed CASISR architecture in Fig. 2, the PP module is the classical PI unit, the PT module is the pretrained state-of-the-art ASISR models, and the DG module is the bicubic down-sampling interpolation. The experiments are accomplished on Google COLAB with Nvidia GPU.

### 4.2 Competing Algorithms

The eight competing open-loop ASISR algorithms with open-source codes released on GitHub website are listed in Tab. 2, including Real-World Scale Arbitrary Super-Resolution (RealArbiSR) [55], Functional Tensor Decomposition-Local Implicit Image Function (FTD-LIIF) [56] , Sobolev Fourier Neural Operator (SoFoNO) [57], Gaussian Splatting for Arbitrary-scale Super-Resolution (GSASR) [58], Thera [59], Continuous Optical Zooming (COZ) [60], Basic Arbitrary-Scale Video Super-Resolution BasicAVSR [61], and Scale-Arbitrary Super-Resolution (ArbSR) [62]. It is worth mentioning that each testing image frame of a video is independently evaluated for the BasicAVSR. According to the proposed CASISR and regarding the pretrained models of the aforementioned open-loop algorithms as the PT modules, the eight related close-loop CASISR algorithms are proposed: Circular RealArbiSR (CRealArbiSR), Circular FTD-LIIF (CFTD-LIIF), Circular SoFoNO (CSoFoNO), Circular GSASR (CGSASR), Circular Thera (CThera), Circular COZ (CCOZ), Circular BasicAVSR (CBasicAVSR), and Circular ArbSR (CArbSR).

**Tab. 2. Competing Algorithms**

| Names | Websites | References |
|---|---|---|
| RealArbiSR | https://github.com/nonozhizhiovo/RealArbiSR | [55] |
| FTD-LIIF | https://github.com/GuanchengZhou/FTD-LIIF | [56] |
| SoFoNO | https://github.com/CMLab-Korea/SoFoNO | [57] |

| GSASR | https://github.com/ChrisDud0257/GSASR | [58] |
|---|---|---|
| Thera | https://github.com/prs-eth/thera | [59] |
| COZ | https://github.com/pf0607/COZ.git | [60] |
| BasicAVSR | https://github.com/shangwei5/BasicAVSR | [61] |
| ArbSR | https://github.com/Lornatang/ArbSR-PyTorch | [62] |

**4.3 Experimental Datasets**

All competing algorithms share the relevant pretrained models and testing datasets on their websites. The testing datasets are enumerated in Tab. 3, including the size of GT images, the total number of images and the total number of scale factors. It should be noted that the testing data of the BasicAVSR contain 30 videos with 100 image frames per video. In addition, the FTD-LIIF, the SoFoNO, the Thera, and the ArbSR apply the same Urban100 testing dataset.

**Tab. 3. Testing Datasets**

| Competing Algorithms | Size of Ground-Truth Images | Total Number of Images | Total Number of Scale Factors |
|---|---|---|---|
| RealArbiSR | 1132×712~1241×816 | 50 or 83 | 11 |
| FTD-LIIF | 576×1024~1280×963 | 100 | 5 |
| SoFoNO | 576×1024~1280×963 | 100 | 9 |
| GSASR | 720×720 | 100 | 6 |
| Thera | 576×1024~1280×963 | 100 | 5 |
| COZ | 1500×1400~3200×2400 | 34 or 37 | 5 |
| BasicAVSR | 1280×720 | 30×100 | 5 |
| ArbSR | 576×1024~1280×963 | 100 | 5 |

**4.4 Experimental Results**

4.4.1 Raw Outcomes

The raw testing outcomes of the eight competing ASISR algorithms and the related CASISR algorithms are collected in Tables 4 to 11. The performance metrics comprise Peak Signal-to-Noise Ratio (PSNR) and Structural Similarity (SSIM). "Method" means the ASISR and the CASISR methods. "Open" means the open-loop ASISR algorithm, "Close" means the related closed-loop CASISR algorithm, and "Closed-Open" means the difference between the CASISR and the ASISR. "Scale" represents SR scale factors. "Mean" denotes the average in the column and "Max" denotes the maximum in the column with bold font. $PSNR_o$ is the average PSNR of the open-loop algorithm, $PSNR_c$ is the average PSNR of the closed-loop algorithm, $\Delta PSNR_a$ is the average increment of PSNR which equals the difference between $PSNR_c$ and $PSNR_o$, and $\Delta PSNR_m$ is the maximum increment of PSNR. $SSIM_o$ is the average SSIM of the open-loop algorithm, $SSIM_c$ is the average SSIM of the closed-loop algorithm, $\Delta SSIM_a$ is the average increment of SSIM which equals the difference between $SSIM_c$ and $SSIM_o$, and $\Delta SSIM_m$ is the maximum increment of SSIM. The aforementioned performance metrics are described by the following mathematical equations:

$$
\begin{gathered}
\mathrm{PSNR_c} = \frac{1}{Z}\sum_{i=1}^{Z}\mathrm{PSNR_{ci}}, \mathrm{PSNR_o} = \frac{1}{Z}\sum_{i=1}^{Z}\mathrm{PSNR_{oi}} \\
\Delta\mathrm{PSNR_a} = \frac{1}{Z}\sum_{i=1}^{Z}\Delta\mathrm{PSNR_i} = \frac{1}{Z}\sum_{i=1}^{Z}\left(\mathrm{PSNR_{ci}} - \mathrm{PSNR_{oi}}\right) = \mathrm{PSNR_c} - \mathrm{PSNR_o} \\
\Delta\mathrm{PSNR_m} = \max_{\Delta\mathrm{PSNR_i}}\left\{\Delta\mathrm{PSNR_i} \,|\, i = 1, 2, \cdots, Z\right\} \\
\mathrm{SSIM_c} = \frac{1}{Z}\sum_{i=1}^{Z}\mathrm{SSIM_{ci}}, \mathrm{SSIM_o} = \frac{1}{Z}\sum_{i=1}^{Z}\mathrm{SSIM_{oi}} \\
\Delta\mathrm{SSIM_a} = \frac{1}{Z}\sum_{i=1}^{Z}\Delta\mathrm{SSIM_i} = \frac{1}{Z}\sum_{i=1}^{Z}\left(\mathrm{SSIM_{ci}} - \mathrm{SSIM_{oi}}\right) = \mathrm{SSIM_c} - \mathrm{SSIM_o} \\
\Delta\mathrm{SSIM_m} = \max_{\Delta\mathrm{SSIM_i}}\left\{\Delta\mathrm{SSIM_i} \,|\, i = 1, 2, \cdots, Z\right\}
\end{gathered}
\tag{31}
$$

where: Z is total number of images in a testing dataset, $\mathrm{PSNR_{ci}}$ is the PSNR of the closed-loop algorithm for the i-th image, $\mathrm{PSNR_{oi}}$ is the PSNR of the open-loop algorithm for the i-th image, $\Delta\mathrm{PSNR_i}$ is the PSNR increment between $\mathrm{PSNR_{ci}}$ and $\mathrm{PSNR_{oi}}$, $\mathrm{SSIM_{ci}}$ is the SSIM of the closed-loop algorithm for the i-th image, $\mathrm{SSIM_{oi}}$ is the SSIM of the open-loop algorithm for the i-th image, and $\Delta\mathrm{SSIM_i}$ is the SSIM increment between $\mathrm{SSIM_{ci}}$ and $\mathrm{SSIM_{oi}}$.

Tab. 4 shows the original laboratory findings of the RealArbiSR and the CRealArbiSR, Tab. 5 displays the initial laboratory measurements of the FTD-LIIF and the CFTD-LIIF, Tab. 6 illustrates the primitive laboratory observations of the SoFoNO and the CSOFoNO, Tab. 7 demonstrates the original laboratory detections of the GSASR and the CGSASR, Tab. 8 exhibits the raw testing outcomes of the Thera and the CThera, Tab. 9 shows the original testing outputs of the COZ and the CCOZ, Tab. 10 displays the primitive experimental results of the BasicAVSR and the CBasicAVSR, and Tab. 11 illustrates the original experimental data of the ArbSR and the CArbSR.

**Tab. 4. Primitive Experimental Results of RealArbiSR and CRealArbiSR**

| Method | Open | Closed | Closed-Open | | Open | Closed | Closed-Open | |
|---|---|---|---|---|---|---|---|---|
| Scale | $\mathrm{PSNR_o}$ | $\mathrm{PSNR_c}$ | $\Delta\mathrm{PSNR_a}$ | $\Delta\mathrm{PSNR_m}$ | $\mathrm{SSIM_o}$ | $\mathrm{SSIM_c}$ | $\Delta\mathrm{SSIM_a}$ | $\Delta\mathrm{SSIM_m}$ |
| 1.5 | 35.2616 | 35.3015 | 0.0399 | 0.6632 | 0.9800 | 0.9802 | 0.0002 | 0.0075 |
| 1.7 | 32.7765 | 32.9290 | 0.1525 | 2.2226 | 0.9526 | 0.9534 | 0.0008 | 0.0270 |
| 2.0 | 32.7684 | 32.8187 | 0.0503 | 0.5924 | 0.9688 | 0.9692 | 0.0004 | 0.0060 |
| 2.3 | 30.6711 | 30.7882 | 0.1171 | 2.2333 | 0.9407 | 0.9412 | 0.0006 | 0.0124 |
| 2.5 | 30.9946 | 31.0309 | 0.0364 | 0.4478 | 0.9522 | 0.9525 | 0.0002 | 0.0045 |
| 2.7 | 29.7945 | 29.9112 | 0.1166 | 2.9872 | 0.9254 | 0.9400 | 0.0010 | 0.0208 |
| 3.0 | 29.7703 | 29.8034 | 0.0331 | 0.3523 | 0.9324 | 0.9327 | 0.0003 | 0.0063 |
| 3.3 | 28.6482 | 28.7604 | 0.1122 | 3.0186 | 0.8957 | 0.8981 | 0.0024 | 0.0909 |
| 3.5 | 28.8240 | 28.8511 | 0.0271 | 0.2325 | 0.9123 | 0.9127 | 0.0004 | 0.0080 |
| 3.7 | 28.0650 | 28.1761 | 0.1111 | 3.5520 | 0.8802 | 0.8815 | 0.0013 | 0.0297 |
| 4.0 | 28.0969 | 28.1248 | 0.0279 | 0.3173 | 0.8928 | 0.8931 | 0.0004 | 0.0111 |
| Mean | 30.5156 | 30.5905 | 0.0749 | 1.5108 | 0.9303 | 0.9322 | 0.0007 | 0.0204 |
| Max | 35.2616 | 35.3015 | **0.1525** | **3.5520** | 0.9800 | 0.9802 | **0.0024** | **0.0909** |

**Tab. 5. Primitive Experimental Results of FTD-LIIF and CFTD-LIIF**

| Method | Open | Closed | Closed-Open | | Open | Closed | Closed-Open | |
|---|---|---|---|---|---|---|---|---|
| Scale | $PSNR_o$ | $PSNR_c$ | $\Delta PSNR_a$ | $\Delta PSNR_m$ | $SSIM_o$ | $SSIM_c$ | $\Delta SSIM_a$ | $\Delta SSIM_m$ |
| 2.0 | 32.9207 | 32.9377 | 0.0170 | 0.3839 | 0.9987 | 0.9987 | 0.0000 | 0.0001 |
| 3.0 | 28.8444 | 28.8579 | 0.0134 | 0.2104 | 0.9966 | 0.9966 | 0.0000 | 0.0001 |
| 4.0 | 26.7340 | 26.7494 | 0.0154 | 0.1898 | 0.9845 | 0.9846 | 0.0001 | 0.0015 |
| 8.0 | 22.8856 | 22.9046 | 0.0190 | 0.3013 | 0.9263 | 0.9265 | 0.0002 | 0.0059 |
| 16.0 | 20.2629 | 20.2852 | 0.0223 | 0.5413 | 0.8004 | 0.8014 | 0.0010 | 0.0176 |
| Mean | 26.3295 | 26.3470 | 0.0174 | 0.3253 | 0.9413 | 0.9416 | 0.0003 | 0.0050 |
| Max | 32.9207 | 32.9377 | **0.0223** | **0.5413** | 0.9987 | 0.9987 | **0.0010** | **0.0176** |

**Tab. 6. Primitive Experimental Results of SoFoNO and CSOFoNO**

| Method | Open | Closed | Closed-Open | | Open | Closed | Closed-Open | |
|---|---|---|---|---|---|---|---|---|
| Scale | $PSNR_o$ | $PSNR_c$ | $\Delta PSNR_a$ | $\Delta PSNR_m$ | $SSIM_o$ | $SSIM_c$ | $\Delta SSIM_a$ | $\Delta SSIM_m$ |
| 2.0 | 30.9620 | 30.9908 | 0.0289 | 0.4562 | 0.9229 | 0.9232 | 0.0003 | 0.0041 |
| 3.0 | 27.0149 | 27.0316 | 0.0167 | 0.3092 | 0.8438 | 0.8442 | 0.0003 | 0.0031 |
| 4.0 | 24.9678 | 24.9848 | 0.0170 | 0.2730 | 0.7761 | 0.7765 | 0.0004 | 0.0080 |
| 5.0 | 23.5914 | 23.6153 | 0.0238 | 0.4706 | 0.7174 | 0.7180 | 0.0005 | 0.0143 |
| 6.0 | 22.5864 | 22.6131 | 0.0267 | 0.5968 | 0.6680 | 0.6688 | 0.0008 | 0.0169 |
| 7.0 | 21.8318 | 21.8607 | 0.0289 | 0.6437 | 0.6281 | 0.6287 | 0.0006 | 0.0126 |
| 8.0 | 21.2392 | 21.2614 | 0.0223 | 0.4900 | 0.5957 | 0.5962 | 0.0005 | 0.0094 |
| 9.0 | 20.7345 | 20.7510 | 0.0165 | 0.2856 | 0.5673 | 0.5677 | 0.0004 | 0.0094 |
| 10.0 | 20.3128 | 20.3302 | 0.0174 | 0.2527 | 0.5444 | 0.5449 | 0.0005 | 0.0154 |
| Mean | 23.6934 | 23.7154 | 0.0220 | 0.4198 | 0.6960 | 0.6965 | 0.0005 | 0.0104 |
| Max | 30.9620 | 30.9908 | **0.0289** | **0.6437** | 0.9229 | 0.9232 | **0.0008** | **0.0169** |

**Tab. 7. Primitive Experimental Results of GSASR and CGSASR**

| Method | Open | Closed | Closed-Open | | Open | Closed | Closed-Open | |
|---|---|---|---|---|---|---|---|---|
| Scale | $PSNR_o$ | $PSNR_c$ | $\Delta PSNR_a$ | $\Delta PSNR_m$ | $SSIM_o$ | $SSIM_c$ | $\Delta SSIM_a$ | $\Delta SSIM_m$ |
| 2.0 | 36.3731 | 36.4523 | 0.0792 | 2.5800 | 0.9520 | 0.9521 | 0.0001 | 0.0022 |
| 3.0 | 32.8726 | 32.9382 | 0.0655 | 1.7025 | 0.9039 | 0.9041 | 0.0002 | 0.0033 |
| 4.0 | 30.5670 | 30.6338 | 0.0668 | 2.0047 | 0.8573 | 0.8581 | 0.0008 | 0.0540 |
| 6.0 | 28.1626 | 28.2052 | 0.0426 | 0.7245 | 0.7815 | 0.7818 | 0.0004 | 0.0092 |
| 8.0 | 26.7361 | 26.7846 | 0.0485 | 0.8954 | 0.7283 | 0.7285 | 0.0003 | 0.0103 |
| 12.0 | 24.9049 | 24.9601 | 0.0552 | 0.3729 | 0.6615 | 0.6619 | 0.0004 | 0.0155 |
| Mean | 29.9361 | 29.9957 | 0.0596 | 1.3800 | 0.8141 | 0.8144 | 0.0004 | 0.0158 |
| Max | 36.3731 | 36.4523 | **0.0792** | **2.5800** | 0.9520 | 0.9521 | **0.0008** | **0.0540** |

**Tab. 8. Primitive Experimental Results of Thera and CThera**

| Method | Open | Closed | Closed-Open | | Open | Closed | Closed-Open | |
|---|---|---|---|---|---|---|---|---|
| Scale | $PSNR_o$ | $PSNR_c$ | $\Delta PSNR_a$ | $\Delta PSNR_m$ | $SSIM_o$ | $SSIM_c$ | $\Delta SSIM_a$ | $\Delta SSIM_m$ |
| 2.0 | 30.6558 | 30.6916 | 0.0359 | 0.3012 | 0.9264 | 0.9270 | 0.0006 | 0.0039 |
| 3.0 | 26.8078 | 26.8495 | 0.0418 | 0.5087 | 0.8533 | 0.8546 | 0.0013 | 0.0066 |

| 4.0 | 24.7237 | 24.7516 | 0.0279 | 0.4522 | 0.7862 | 0.7874 | 0.0013 | 0.0091 |
|---|---|---|---|---|---|---|---|---|
| 8.0 | 21.1179 | 21.1249 | 0.0070 | 0.1607 | 0.5966 | 0.5971 | 0.0004 | 0.0053 |
| 16.0 | 18.6666 | 18.6735 | 0.0068 | 0.1104 | 0.4468 | 0.4472 | 0.0004 | 0.0058 |
| Mean | 24.3944 | 24.4182 | 0.0239 | 0.3066 | 0.7219 | 0.7227 | 0.0008 | 0.0061 |
| Max | 30.6558 | 30.6916 | **0.0418** | **0.5087** | 0.9264 | 0.9270 | **0.0013** | **0.0091** |

**Tab. 9. Primitive Experimental Results of COZ and CCOZ**

| Method | Open | Closed | Closed-Open | | Open | Closed | Closed-Open | |
|---|---|---|---|---|---|---|---|---|
| Scale | $PSNR_o$ | $PSNR_c$ | $\Delta PSNR_a$ | $\Delta PSNR_m$ | $SSIM_o$ | $SSIM_c$ | $\Delta SSIM_a$ | $\Delta SSIM_m$ |
| 2.0 | 28.8474 | 28.9297 | 0.0822 | 0.8422 | 0.9150 | 0.9154 | 0.0003 | 0.0039 |
| 3.0 | 26.6515 | 26.7318 | 0.0803 | 0.9129 | 0.8781 | 0.8785 | 0.0004 | 0.0059 |
| 4.0 | 25.2185 | 25.2865 | 0.0679 | 0.5423 | 0.8544 | 0.8555 | 0.0012 | 0.0177 |
| 5.0 | 24.3350 | 24.3972 | 0.0622 | 0.4317 | 0.8380 | 0.8389 | 0.0009 | 0.0111 |
| 6.0 | 23.2702 | 23.3163 | 0.0462 | 0.4741 | 0.8076 | 0.8082 | 0.0006 | 0.0098 |
| Mean | 25.6645 | 25.7323 | 0.0678 | 0.6406 | 0.8586 | 0.8593 | 0.0007 | 0.0097 |
| Max | 28.8474 | 28.9297 | **0.0822** | **0.9129** | 0.9150 | 0.9154 | **0.0012** | **0.0177** |

**Tab. 10. Primitive Experimental Results of BasicAVSR and CBasicAVSR**

| Method | Open | Closed | Closed-Open | | Open | Closed | Closed-Open | |
|---|---|---|---|---|---|---|---|---|
| Scale | $PSNR_o$ | $PSNR_c$ | $\Delta PSNR_a$ | $\Delta PSNR_m$ | $SSIM_o$ | $SSIM_c$ | $\Delta SSIM_a$ | $\Delta SSIM_m$ |
| 2.0 | 32.7605 | 32.8084 | 0.0480 | 0.1711 | 0.9225 | 0.9231 | 0.0006 | 0.0018 |
| 3.0 | 28.6397 | 28.6486 | 0.0089 | 0.0268 | 0.8345 | 0.8346 | 0.0001 | 0.0002 |
| 4.0 | 27.1059 | 27.1249 | 0.0190 | 0.0466 | 0.7787 | 0.7791 | 0.0004 | 0.0016 |
| 6.0 | 24.8031 | 24.8114 | 0.0083 | 0.0220 | 0.6975 | 0.6976 | 0.0001 | 0.0003 |
| 8.0 | 23.4240 | 23.4323 | 0.0083 | 0.0572 | 0.6456 | 0.6457 | 0.0001 | 0.0009 |
| Mean | 27.3466 | 27.3651 | 0.0185 | 0.0647 | 0.7758 | 0.7760 | 0.0003 | 0.0010 |
| Max | 32.7605 | 32.8084 | **0.0480** | **0.1711** | 0.9225 | 0.9231 | **0.0006** | **0.0018** |

**Tab. 11. Primitive Experimental Results of ArbSR and CArbSR**

| Method | Open | Closed | Closed-Open | | Open | Closed | Closed-Open | |
|---|---|---|---|---|---|---|---|---|
| Scale | $PSNR_o$ | $PSNR_c$ | $\Delta PSNR_a$ | $\Delta PSNR_m$ | $SSIM_o$ | $SSIM_c$ | $\Delta SSIM_a$ | $\Delta SSIM_m$ |
| 2.0 | 27.5141 | 28.4402 | 0.9262 | 2.3334 | 0.8627 | 0.8799 | 0.0171 | 0.0350 |
| 3.0 | 24.7491 | 25.1236 | 0.3744 | 1.7179 | 0.7593 | 0.7683 | 0.0091 | 0.0269 |
| 4.0 | 23.2328 | 23.4080 | 0.1752 | 0.6928 | 0.6779 | 0.6867 | 0.0088 | 0.0220 |
| 8.0 | 20.4036 | 20.5906 | 0.1870 | 1.2190 | 0.5047 | 0.5093 | 0.0046 | 0.0360 |
| 16.0 | 18.1727 | 18.3892 | 0.2165 | 1.1057 | 0.4275 | 0.4298 | 0.0023 | 0.0196 |
| Mean | 22.8145 | 23.1903 | 0.3759 | 1.4138 | 0.6464 | 0.6548 | 0.0084 | 0.0279 |
| Max | 27.5141 | 28.4402 | **0.9262** | **2.3334** | 0.8627 | 0.8799 | **0.0171** | **0.0360** |

4.4.2 Performance Comparison

Tables 4 to 11 indicate that $\Delta PSNR_a$ and $\Delta PSNR_m$ hold positive values while $\Delta SSIM_a$ and $\Delta SSIM_m$ simultaneously hold non-negative values. The maximum $\Delta PSNR_a$ is 0.9262 dB in Tab. 11 of the CArbSR, the maximum $\Delta PSNR_m$ is 3.5520 dB in Tab. 4 of the CRealArbiSR, the maximum $\Delta SSIM_a$ is 0.0171 in Tab. 11 of the CArbSR, and the maximum $\Delta SSIM_m$ is 0.0909 in Tab. 4 of the CRealArbiSR. $\Delta PSNR_a$ and $\Delta PSNR_m$ of the proposed eight closed-loop CASISR algorithms are shown in Fig.6. $\Delta SSIM_a$ and $\Delta SSIM_m$ of the proposed eight closed-loop CASISR algorithms are shown in Fig.7. Therefore, the proposed CASISR outperforms the classical open-loop ASISR in performance metrics of PSNR and SSIM.

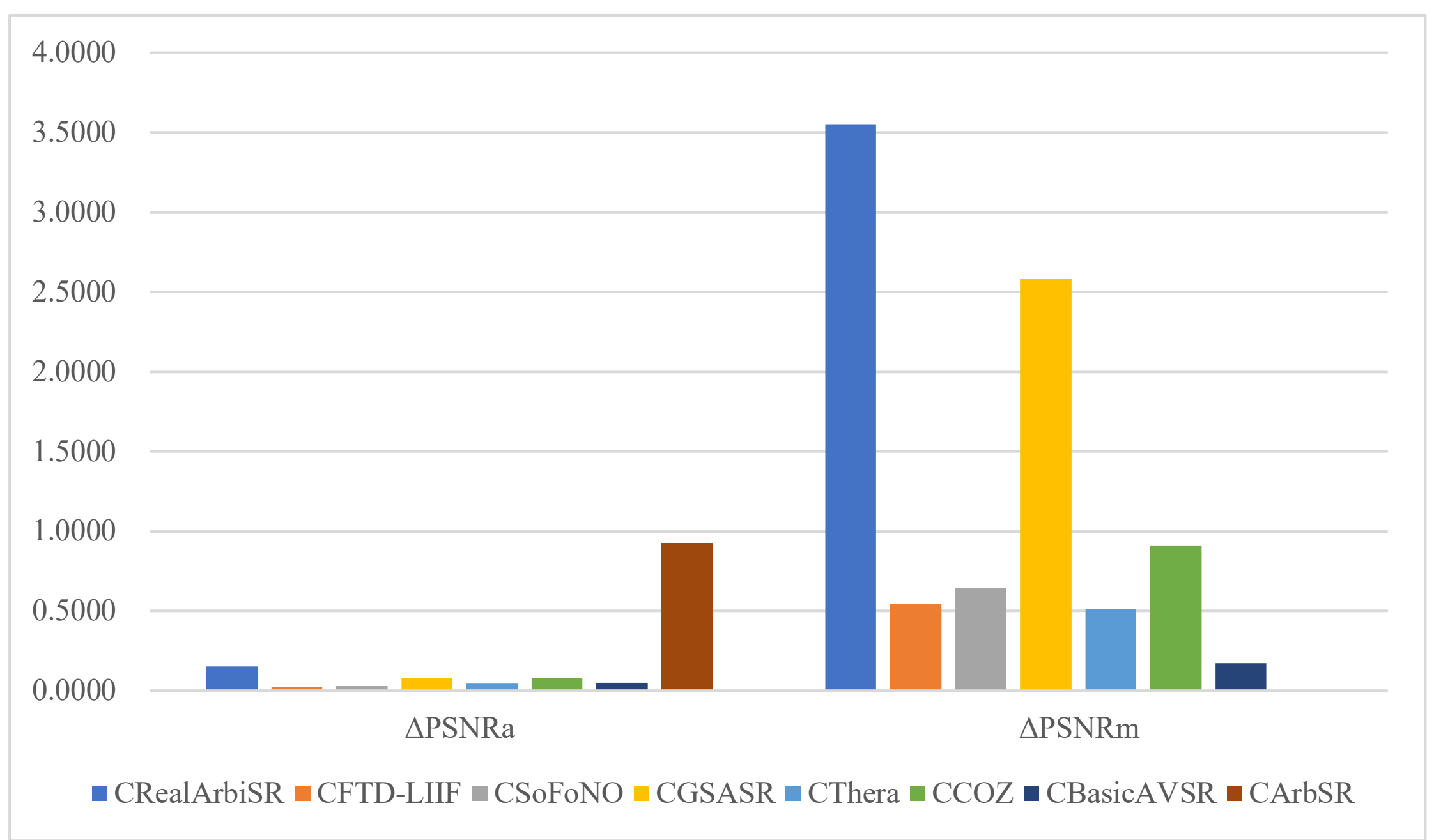

**Fig. 6. $\Delta PSNR_a$ and $\Delta PSNR_m$ of the proposed 8 closed-loop CASISR algorithms.**

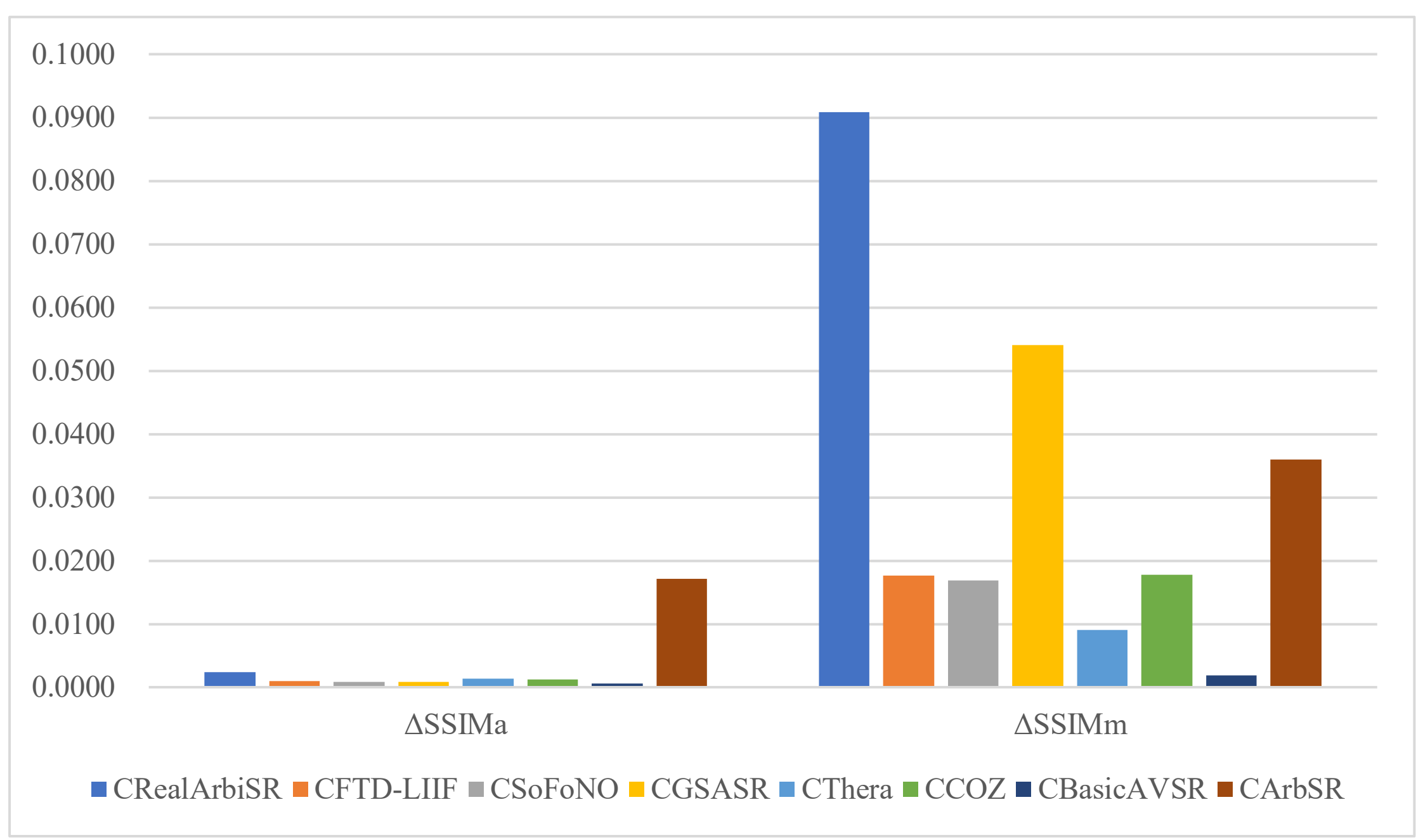

**Fig. 7. $\Delta SSIM_a$ and $\Delta SSIM_m$ of the proposed 8 closed-loop CASISR algorithms.**

4.4.3 Scale Factors

Tables 4 to 11 reveal that PSNR and SSIM decrease while SR scale factor increases. The left-half of Fig. 8 displays the relationship curves of $PSNR_o$ and scale factors, and the right-half of Fig. 8 displays the relationship curves of $PSNR_c$ and scale factors. This is due to the fact that the LR image at larger scale factor loses more details than that at smaller scale factor. Especially, Tab. 4 comprises 11 scale factors from 1.5 to 4.0, including integral and fractional values.

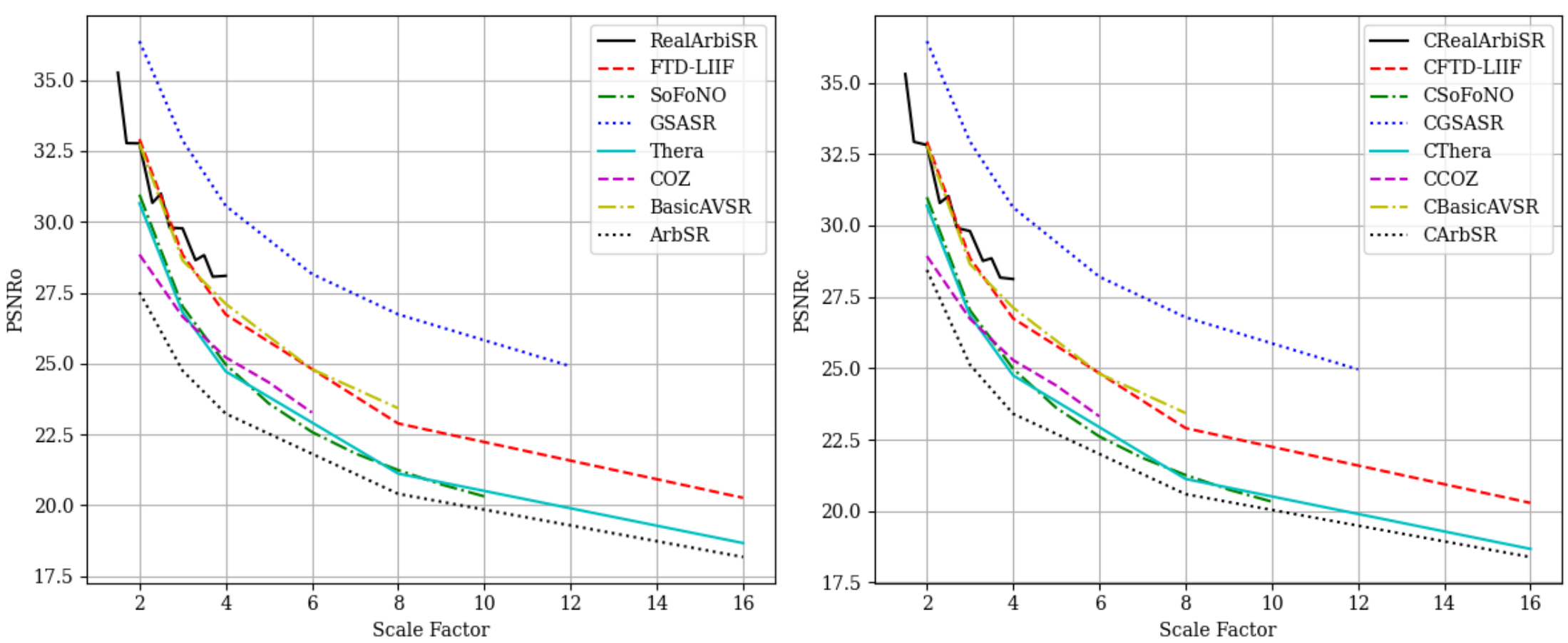

**Fig. 8. Relationship curves of $PSNR_c$/$PSNR_o$ and scale factors.**

According to Tab. 4, Fig. 9 illustrates the relationship curve of $\Delta PSNR_a$ and scale factors, and Fig. 10 illustrates the relationship curve of $\Delta PSNR_m$ and scale factors. $\Delta PSNR_a$ is a sawtooth curve with oscillation damping, and $\Delta PSNR_m$ is a sawtooth curve with oscillation rising. Fig. 9 and 10 manifest that $\Delta PSNR_a$ and $\Delta PSNR_m$ at scale factors 1.7, 2.3, 2.7, 3.3, and 3.7 are greater than those at scale factors 1.5, 2, 2.5, 3.5, and 4. Hence, the proposed CASISR is more suitable for fractional scale factors except for 1.5, 2.5, and 3.5. This is owing to the fact that down-sampling at fractional scale factors except for 1.5, 2.5, and 3.5 misses more image details than that of integral scale factors and the classical ASISR cannot adeptly recover the dropped nuances.

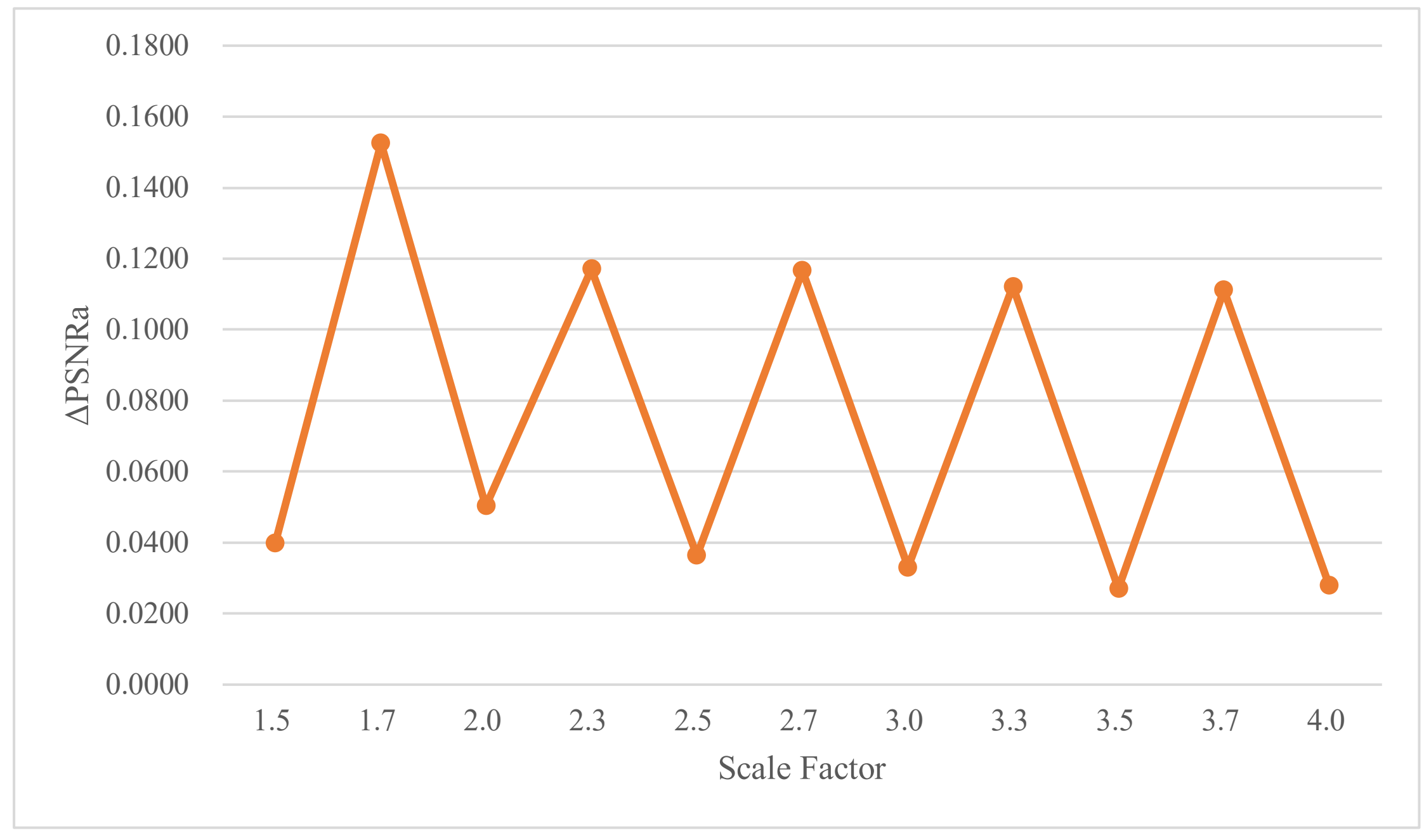

**Fig. 9. Relationship curve of $\Delta PSNR_a$ and scale factors in Tab. 4.**

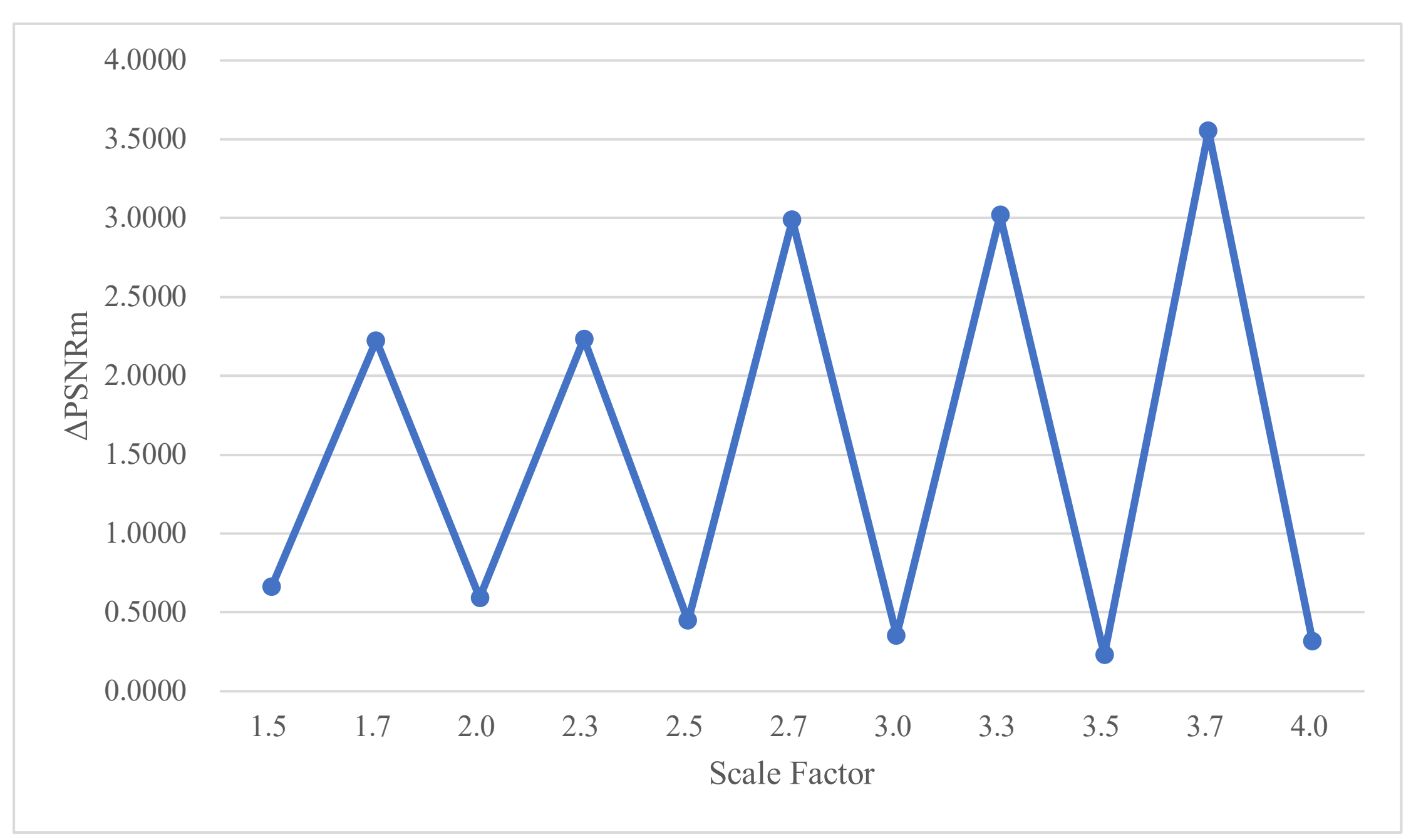

**Fig. 10. Relationship curve of $\Delta PSNR_m$ and scale factors in Tab. 4.**

Fig. 11 demonstrates the relationship curves of $\Delta PSNR_a$ and scale factors in Tables 5 to 11. Because the CArbSR possesses the maximum $\Delta PSNR_a$, it is separately demonstrated in the right part of Fig. 11. $\Delta PSNR_a$ descends with oscillation while scale factor ascends. The only exception is the CFD-LIIF algorithm. Fig. 12 demonstrates the relationship curves of $\Delta PSNR_m$ and scale factors in Tables 5 to 11. Although the CArbSR has the similar data range as the other algorithms, it is also separately demonstrated in the right part of Fig. 12 to keep consistent with Fig. 11. $\Delta PSNR_m$ descends with oscillation while scale factor ascends. The only exception is also the CFD-LIIF algorithm.

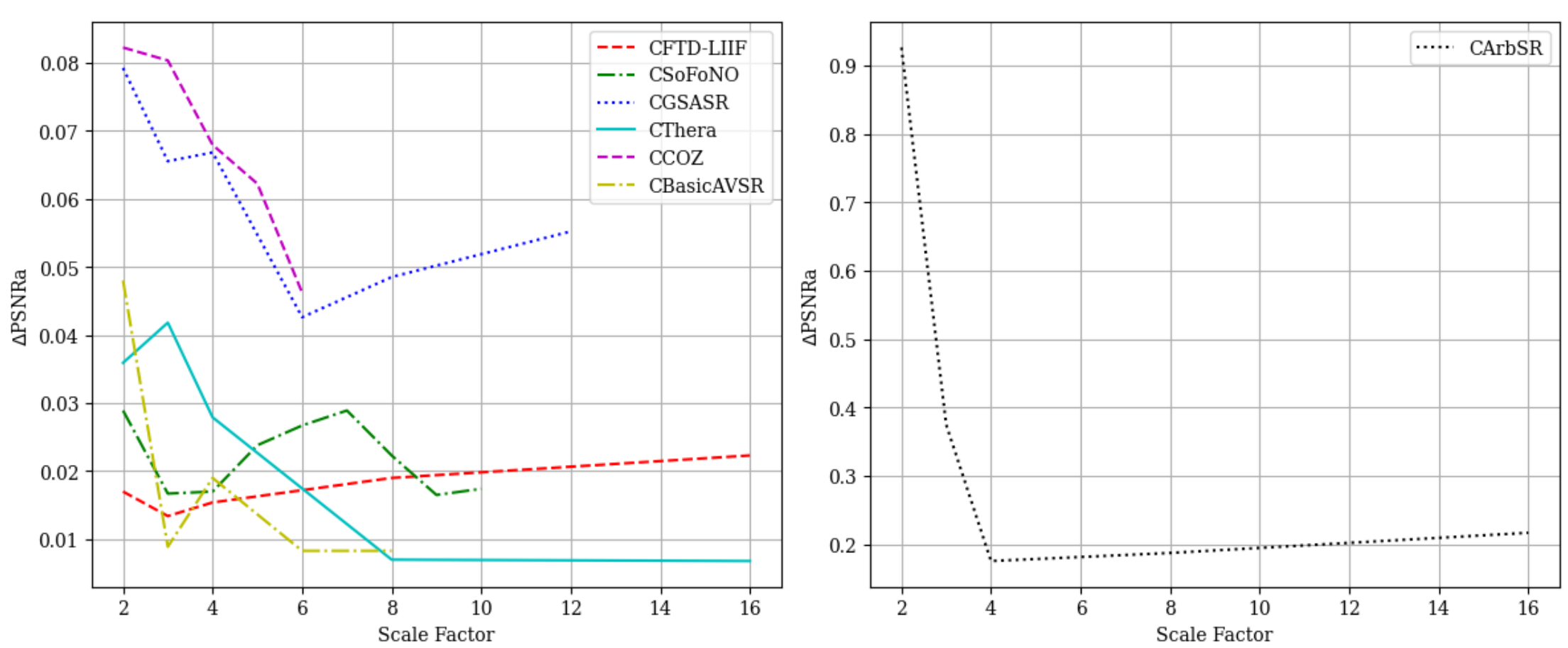

**Fig. 11. Relationship curves of $\Delta PSNR_a$ and scale factors in Tables 5 to 11.**

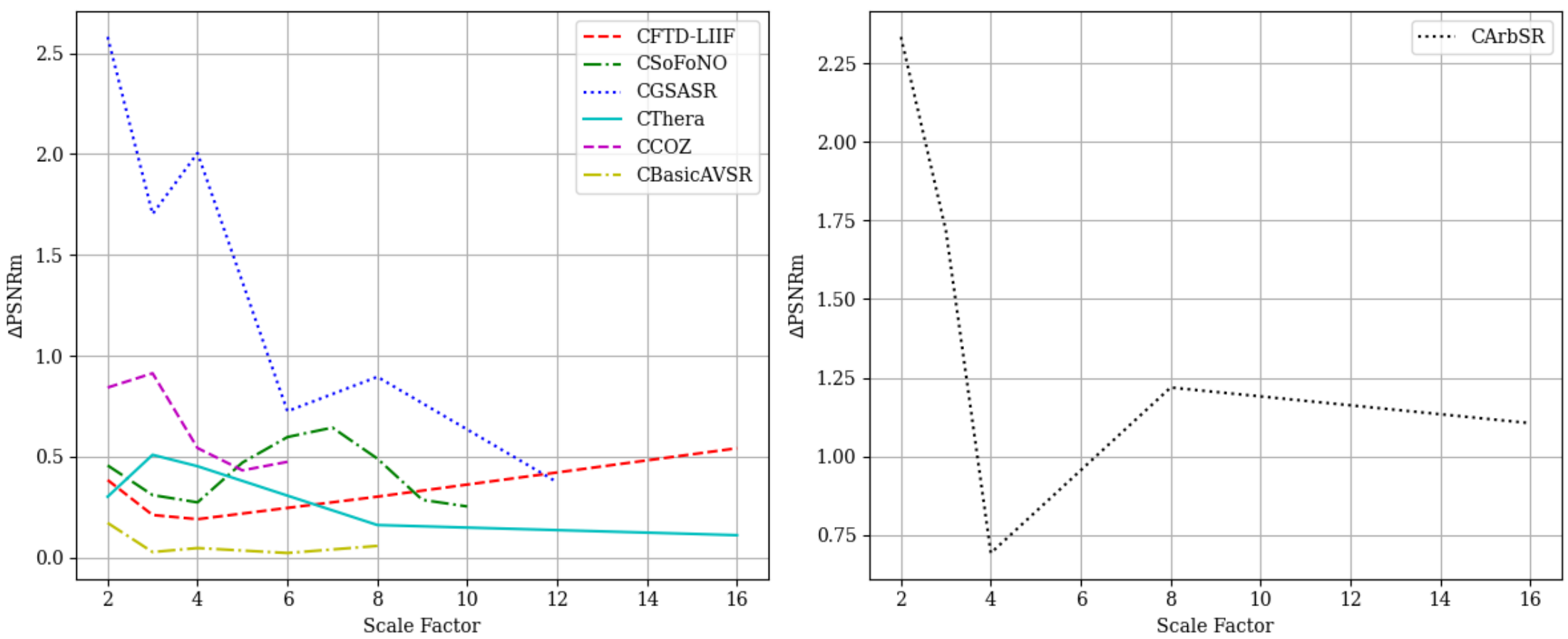

**Fig. 12. Relationship curves of $\Delta PSNR_m$ and scale factors in Tables 5 to 11.**

4.4.4 Visual Comparison

In order to clearly compare the performance of image reconstruction between the CASISR and the ASISR, Figures 13 to 20 exhibit the related images at the scale factor with maximum PSNR increment.

Taking Fig. 13 as an example, it exhibits the experimental results of the CRealArbiSR and the RealArbiSR at scale factor 3.7 with maximum PSNR increment 3.5520 dB. Subfigure (a) is the LR image, subfigure (b) is the GT image, subfigure (d) is the SR image of the RealArbiSR, subfigure (f) is the SR image of the CRealArbiSR. It is hard for human eyes to find the discrimination between subfigures (d) and (f). Hence, the absolute difference image between the SR image and the GT image is computed. The first-order absolute difference image $\mathbf{a}_a$ between image $\mathbf{a}_1$ and image $\mathbf{a}_2$ is defined by the following mathematical equations:

$$
\begin{gathered}
\mathbf{a}_a = |\mathbf{a}_1 - \mathbf{a}_2| \\
a_{a;i} = |a_{1;i} - a_{2;i}| \\
\mathbf{a}_a, \mathbf{a}_1, \mathbf{a}_2 \in R^D; i = 1, 2, \cdots, D
\end{gathered}
\quad .(32)
$$

where: $\mathbf{a}_{a;i}$ is the i-th element of $\mathbf{a}_a$, $\mathbf{a}_{1;i}$ is the i-th element of $\mathbf{a}_1$, and $\mathbf{a}_{2;i}$ is the i-th element of $\mathbf{a}_2$; $|\bullet|$ is the element-wise operator of absolute value.

Subfigure (e) is the absolute difference image of the SR image for the RealArbiSR. Subfigure (g) is the absolute difference image of the SR image for the CRealArbiSR. It is easy for human vision system to check the distinction between subfigures (e) and (g). Compared with subfigure (g), subfigure (e) holds obvious channel shift of red color. For the purpose of magnifying the discrepancy, the second-order logical absolute difference image, also the logical absolute difference image between the absolute difference images, is calculated. The second-order logical absolute difference image $\mathbf{a}_{aa}$ between the absolute difference images $\mathbf{a}_{a1}$ and $\mathbf{a}_{a2}$ is defined by the following mathematical formulas:

$$
\begin{aligned}
&\mathbf{a}_{aa} = \left|\mathbf{a}_{a1} - \mathbf{a}_{a2}\right|_{L} \\
&a_{aa;i} = \begin{cases} \left|a_{1;i} - a_{2;i}\right|, a_{1;i} > a_{2;i} \\ 0, a_{1;i} \le a_{2;i} \end{cases} \\
&\mathbf{a}_{aa}, \mathbf{a}_{a1}, \mathbf{a}_{a2} \in R^{D}; i = 1, 2, \cdots, D
\end{aligned} \quad .(31)
$$

where: $\mathbf{a}_{aa;i}$ is the i-th element of $\mathbf{a}_{aa}$, $\mathbf{a}_{a1;i}$ is the i-th element of $\mathbf{a}_{a1}$, and $\mathbf{a}_{a2;i}$ is the i-th element of $\mathbf{a}_{a2}$; $|\bullet|_{L}$ is the element-wise operator of logical absolute value.

The second-order logical absolute difference image between subfigures (g) and (e), is exhibited in subfigure (c). Subfigure (c) clearly indicates that the CRealArbiSR is superior to the RealArbiSR in the quality of SR image.

Figures 14 to 20 are similar to Fig. 13. Fig. 14 shows the experimental results of the CFTD-LIIF and the FTD-LIIF at scale factor 16 with maximum PSNR increment 0.5413 dB. Fig. 15 displays the laboratory results of the CSoFoNO and the SoFoNO at scale factor 7 with maximum PSNR increment 0.6437 dB. Fig. 16 illustrates the testing results of the CGSASR and the GSASR at scale factor 2 with maximum PSNR increment 1.0381 dB. Fig. 17 demonstrates the experimental results of the CThera and the Thera at scale factor 3 with maximum PSNR increment 0.4771 dB. Fig. 18 shows the laboratory results of the CCOZ and the COZ at scale factor 3 with maximum PSNR increment 0.9129 dB. Fig. 19 displays the testing results of the CBasicAVSR and the BasicAVSR at scale factor 3 with maximum PSNR increment 0.1771 dB. Fig. 20 illustrates the experimental results of the CArbSR and the ArbSR at scale factor 3 with maximum PSNR increment 2.1451 dB.

The divergence between subfigure (g) and (e) can be almost easily examined and subfigure (c) further distinctly demonstrates the divergence between the CASISR and the ASISR in Figures 14 to 20. The mere exception is Fig. 19 and only subfigure (c) plainly demonstrates the divergence.

Figures 13 and 18 reveal that the proposed CASISR is extremely effective for text images. Figures 14, 15, 16,17, 19, and 20 reveal that the proposed CASISR is also remarkably effective for stripe images. These stem from the fact that the text and stripe images possess many suddenly changed edges which are prone to be corrupted in the LR images.

In order to further explicitly compare the capability of image reconstruction between the CASISR and the ASISR, Figures 21 to 31 show the related images of the RealArbiSR and the CRealArbiSR at 11 scale factors. However, the difference between subfigure (g) and (e) cannot be effortlessly checked by human eyes and only subfigure (c) explicitly shows the difference between the CRealArbiSR and the RealArbiSR. Figure 23, 24, 25, 26, 27, and 30 make clear that the proposed CRealArbiSR is exceptionally beneficial for stripe images. Fig. 31 also makes clear that the proposed CRealArbiSR is also exceedingly beneficial for text image.

To sum up, the proposed closed-loop CASISR holds superiority over the classical ASISR in the capability of image reconstruction.

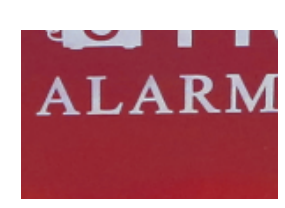

(a) LR image

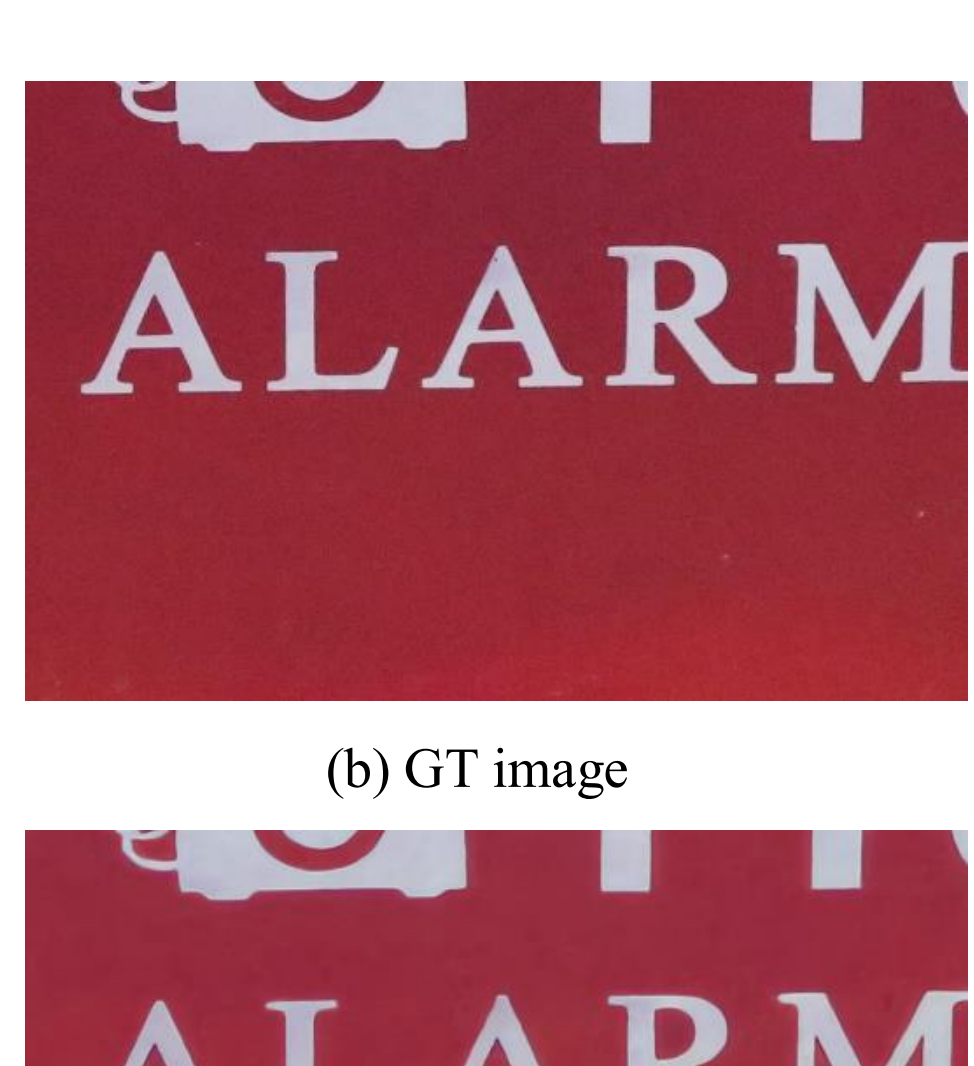

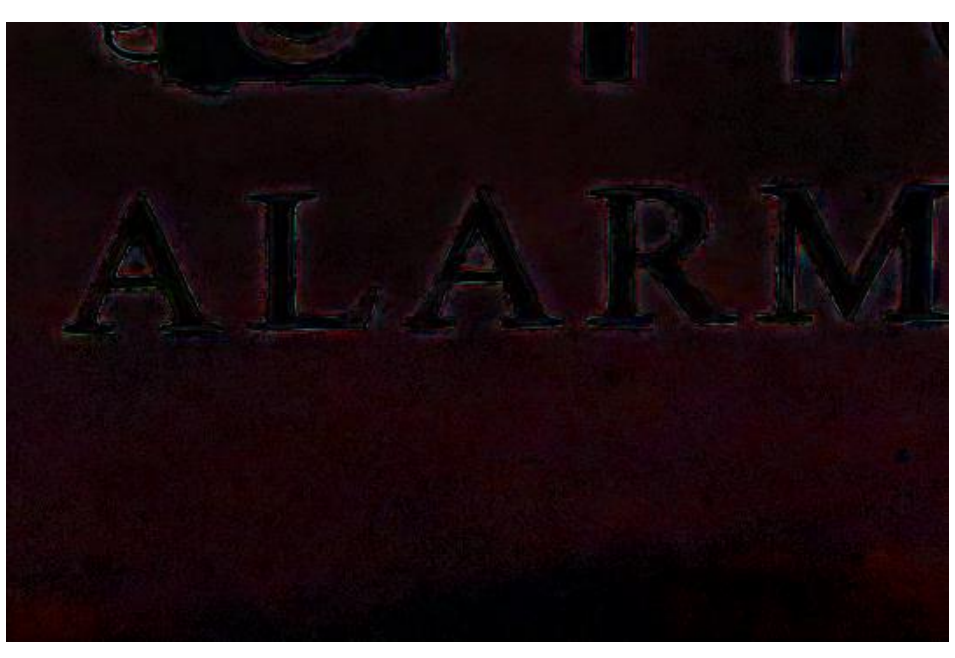

(b) GT image (c) Second-order difference image

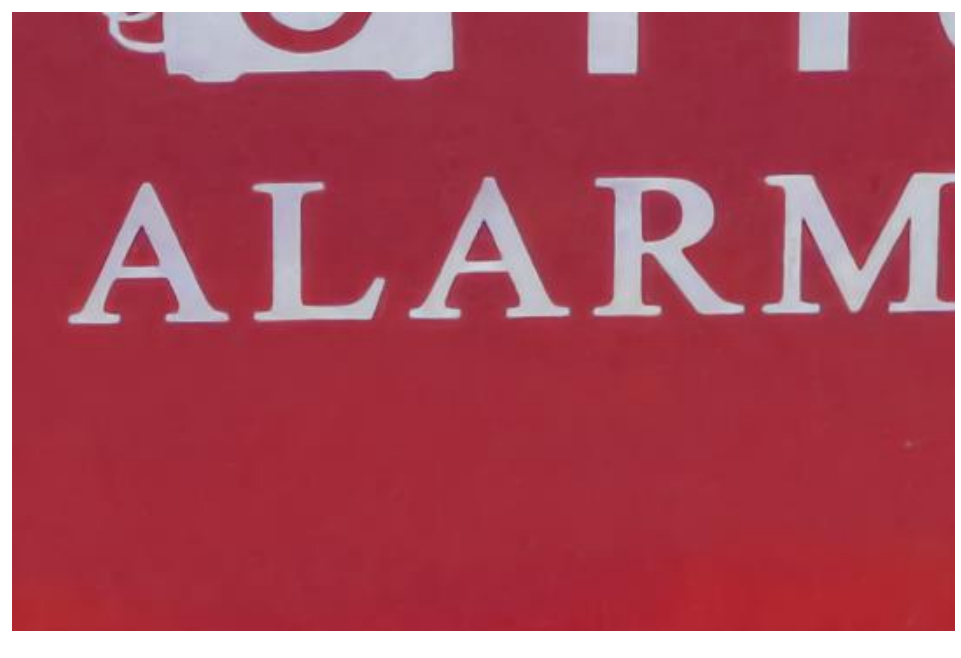

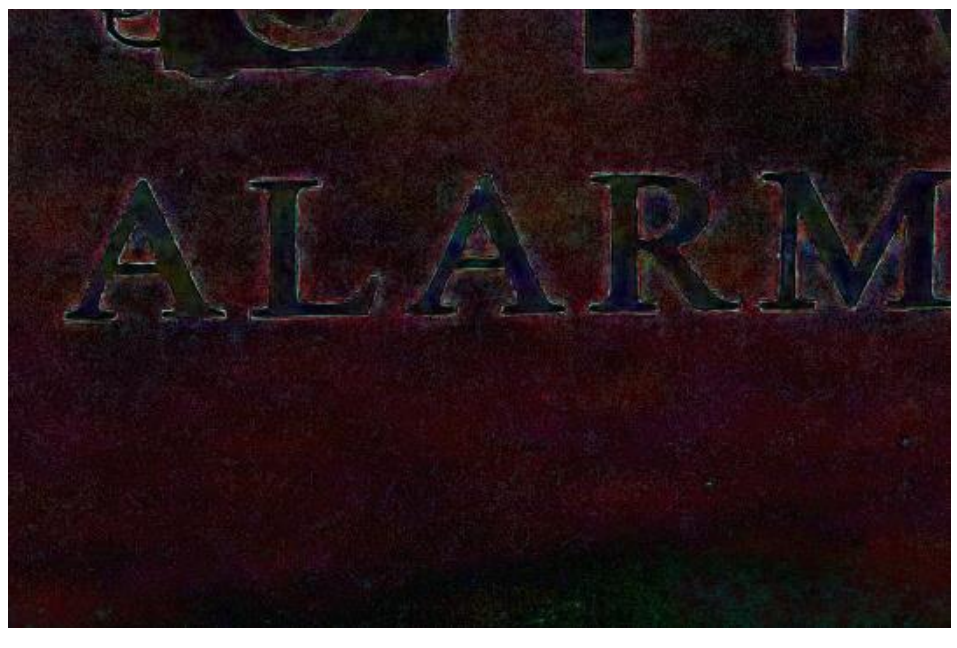

(d) SR image of RealArbiSR (PSNR=28.2003,SSIM=0.9170) (e) Difference image of RealArbiSR

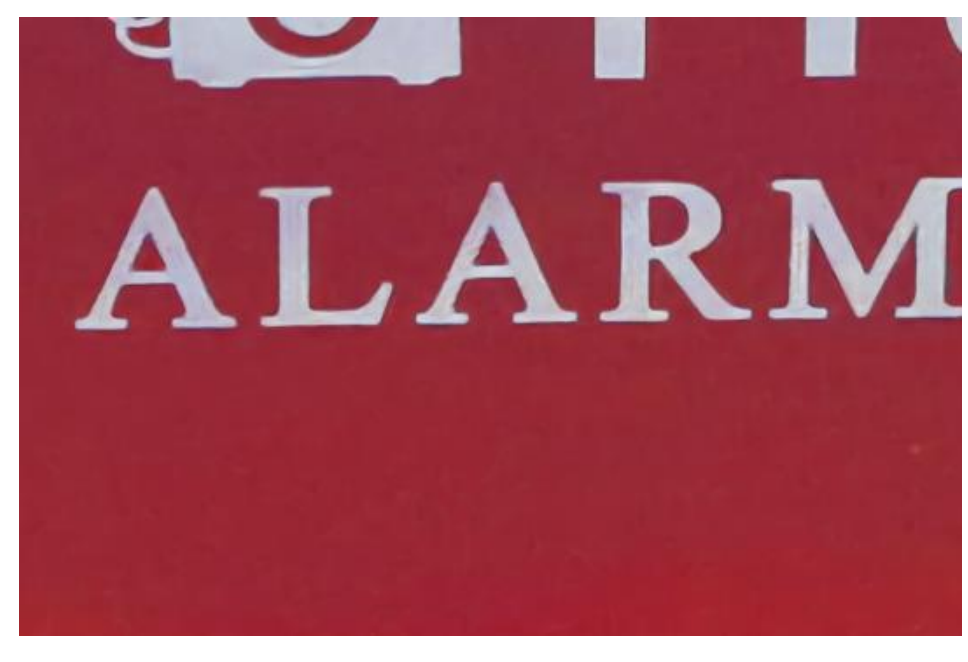

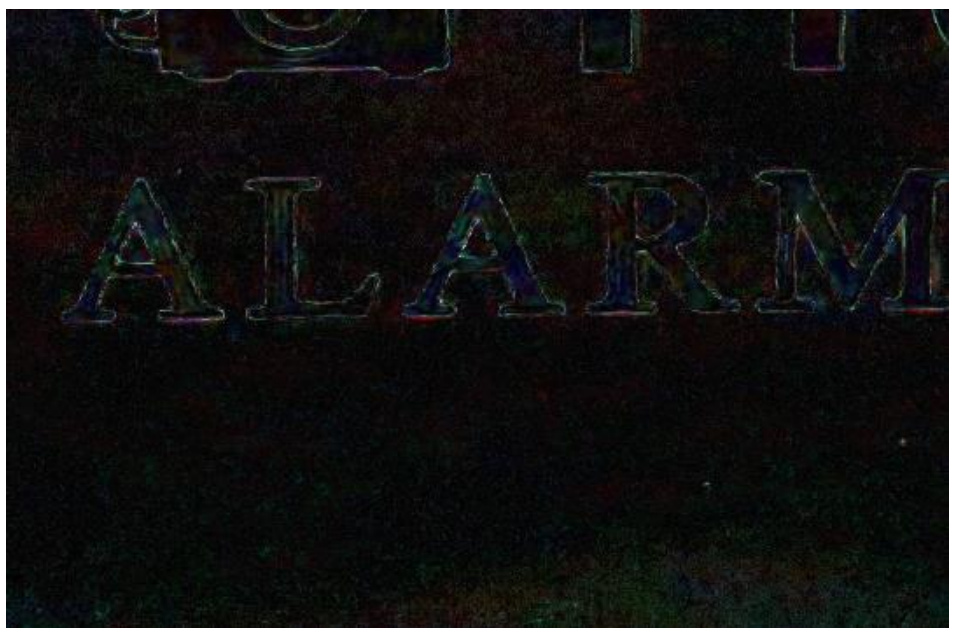

(f) SR image of CRealArbiSR (PSNR=31.7523,SSIM=0.9221) (g) Difference image of CRealArbiSR

**Fig. 13. The results at scale factor 3.7 with maximum PSNR increment 3.5520 dB.**

(a) LR image

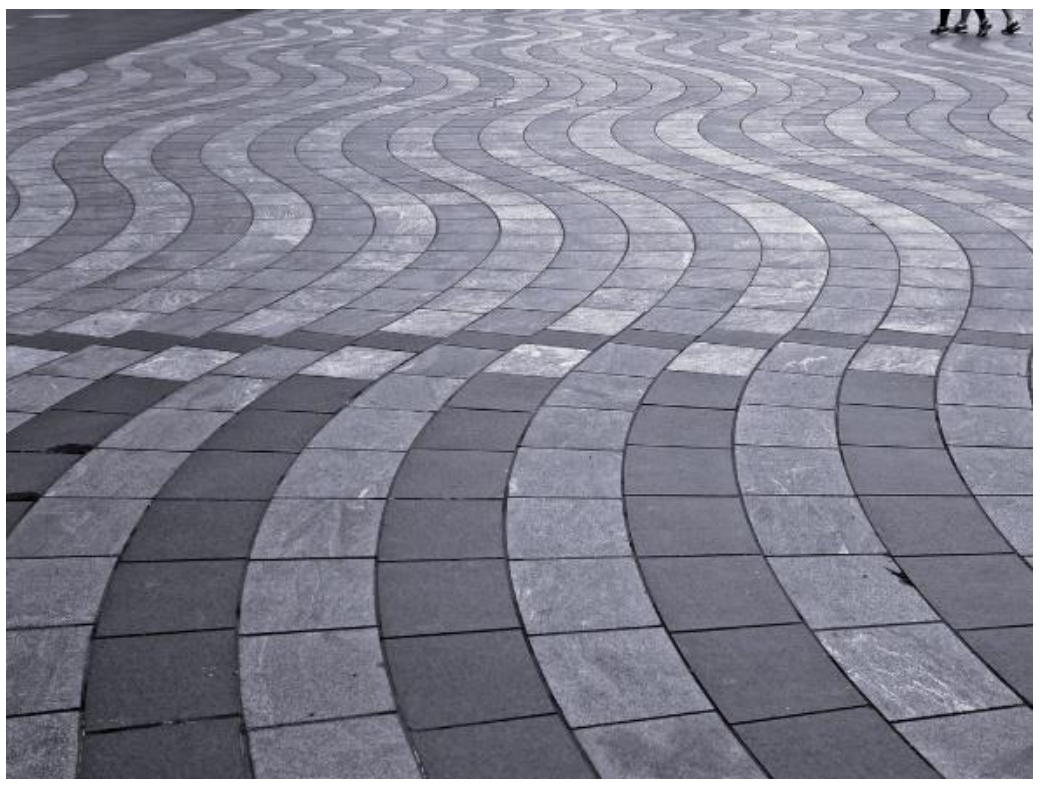

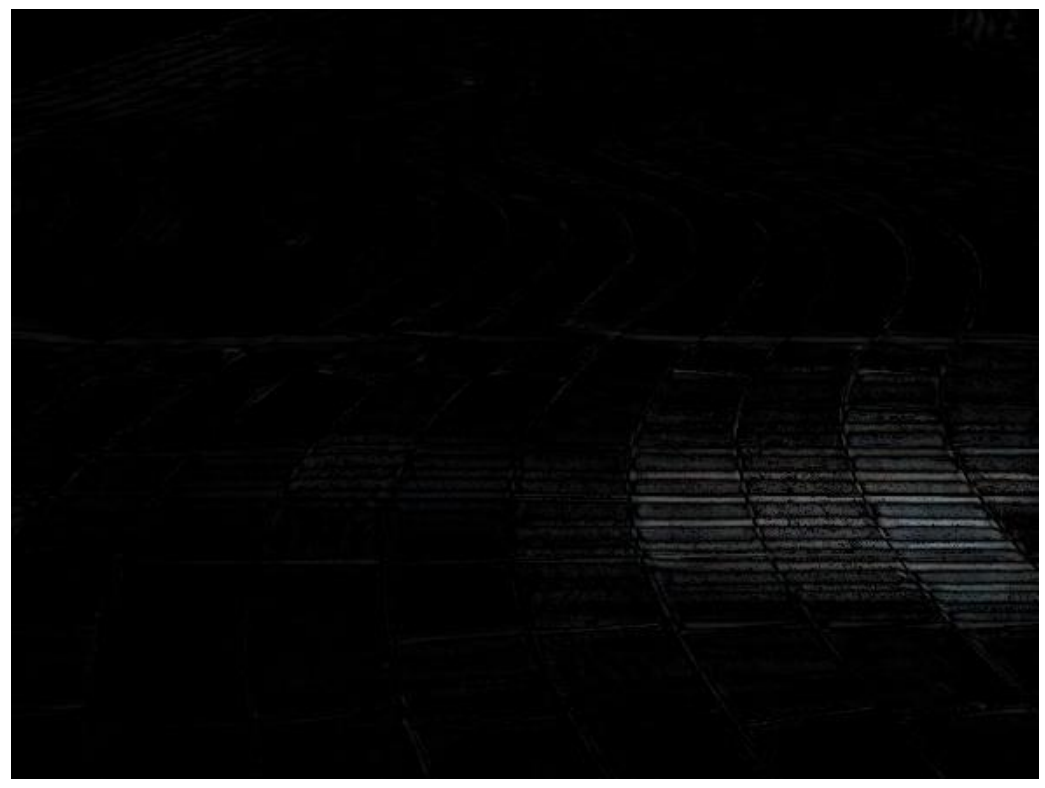

(b) GT image (c) Second-order difference image

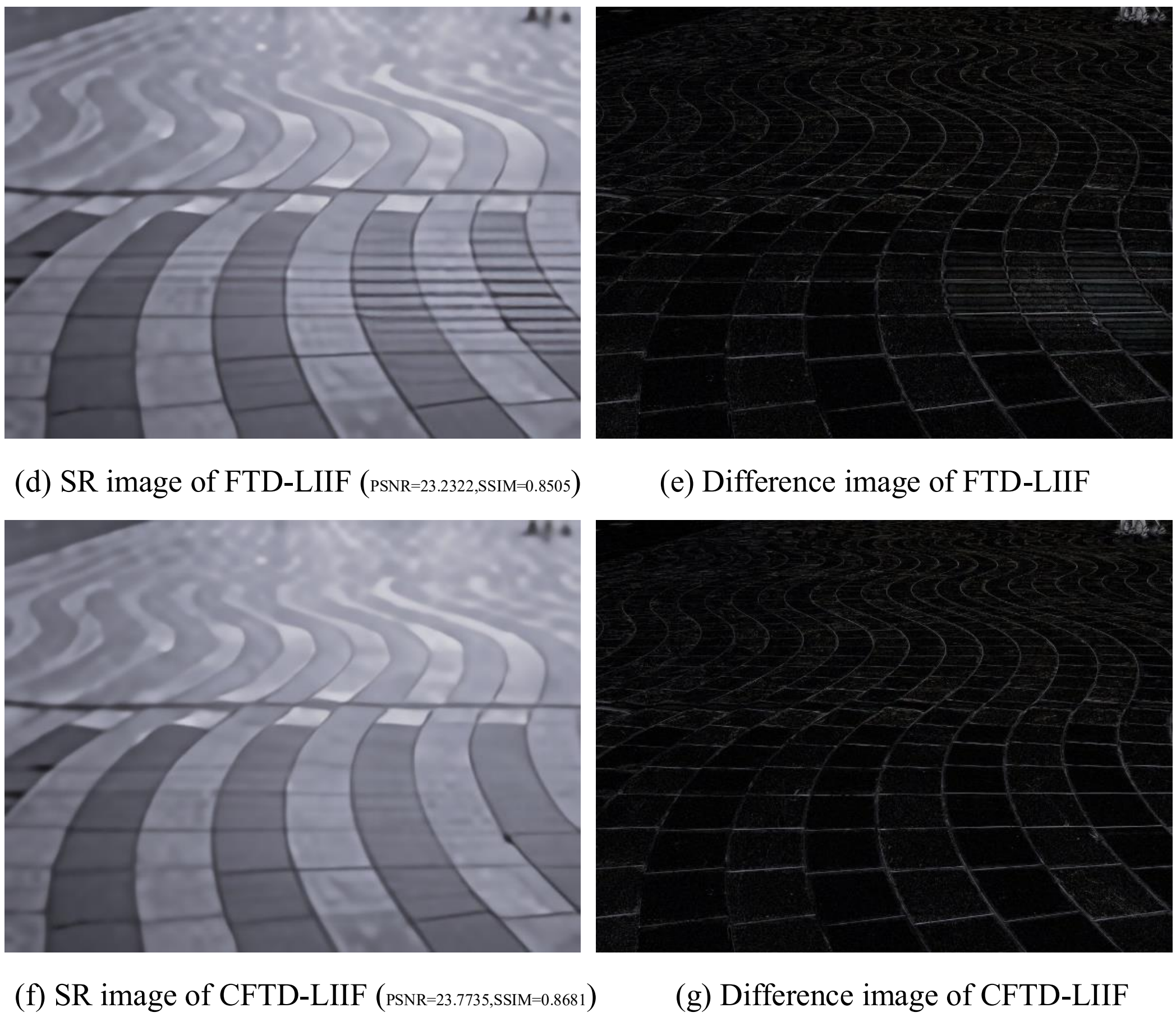

(d) SR image of FTD-LIIF (PSNR=23.2322,SSIM=0.8505) (e) Difference image of FTD-LIIF

(f) SR image of CFTD-LIIF (PSNR=23.7735,SSIM=0.8681) (g) Difference image of CFTD-LIIF

**Fig. 14. The results at scale factor 16 with maximum PSNR increment 0.5413 dB.**

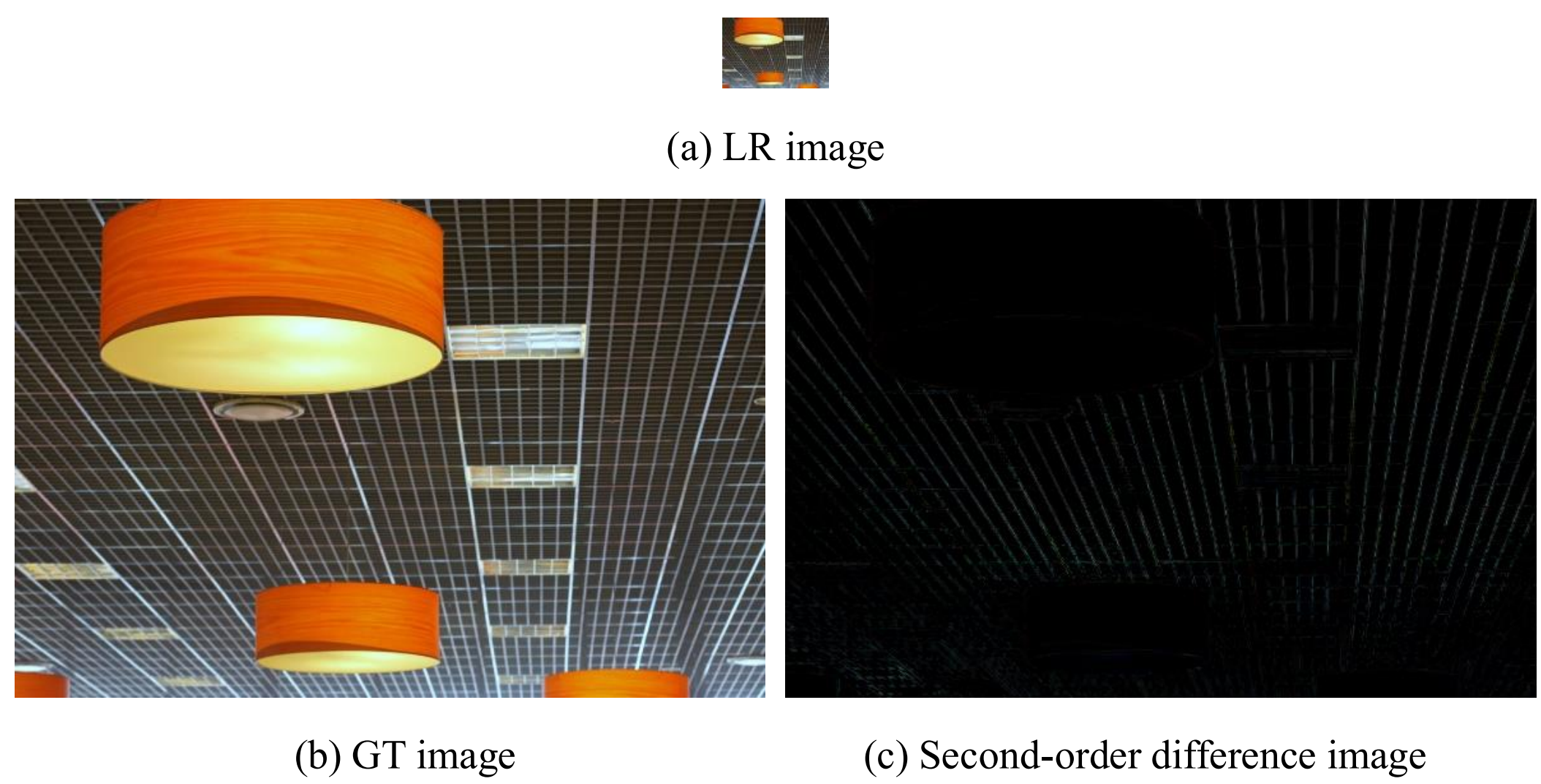

(a) LR image

(b) GT image (c) Second-order difference image

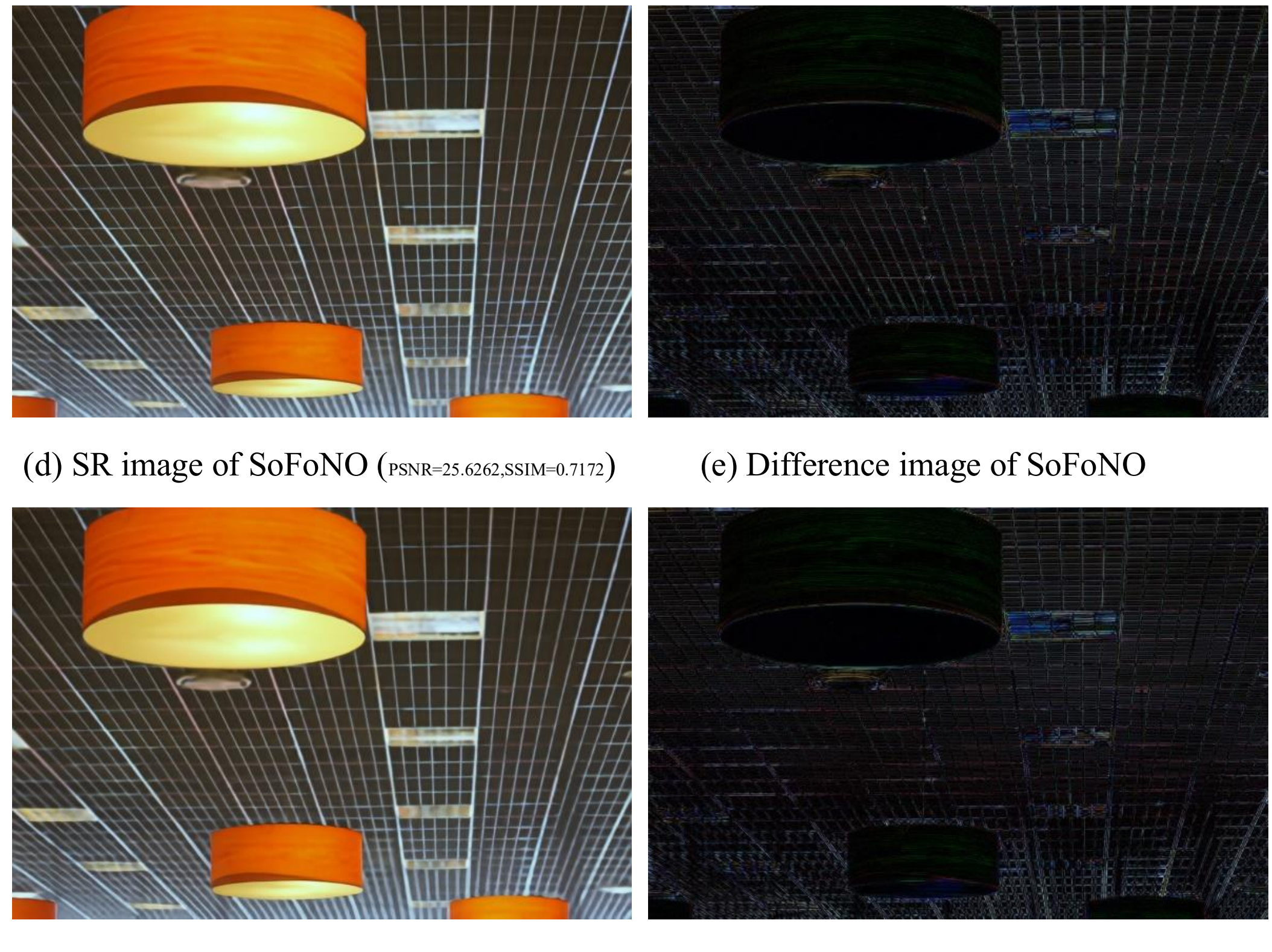

(d) SR image of SoFoNO (PSNR=25.6262,SSIM=0.7172) (e) Difference image of SoFoNO

(f) SR image of CSoFoNO (PSNR=26.2699,SSIM=0.7256) (g) Difference image of CSoFoNO

**Fig. 15. The results at scale factor 7 with maximum PSNR increment 0.6437 dB.**

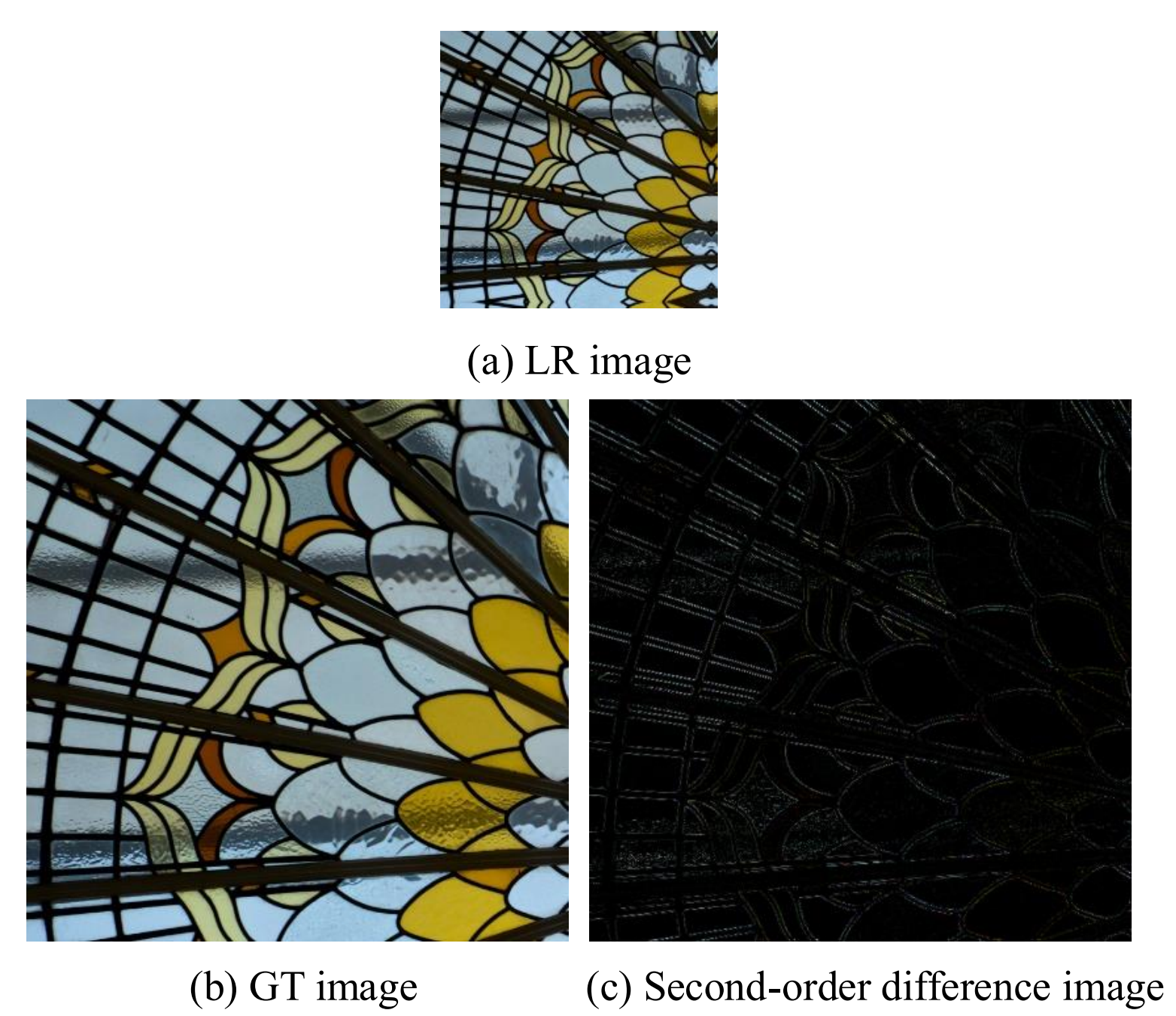

(a) LR image

(b) GT image (c) Second-order difference image

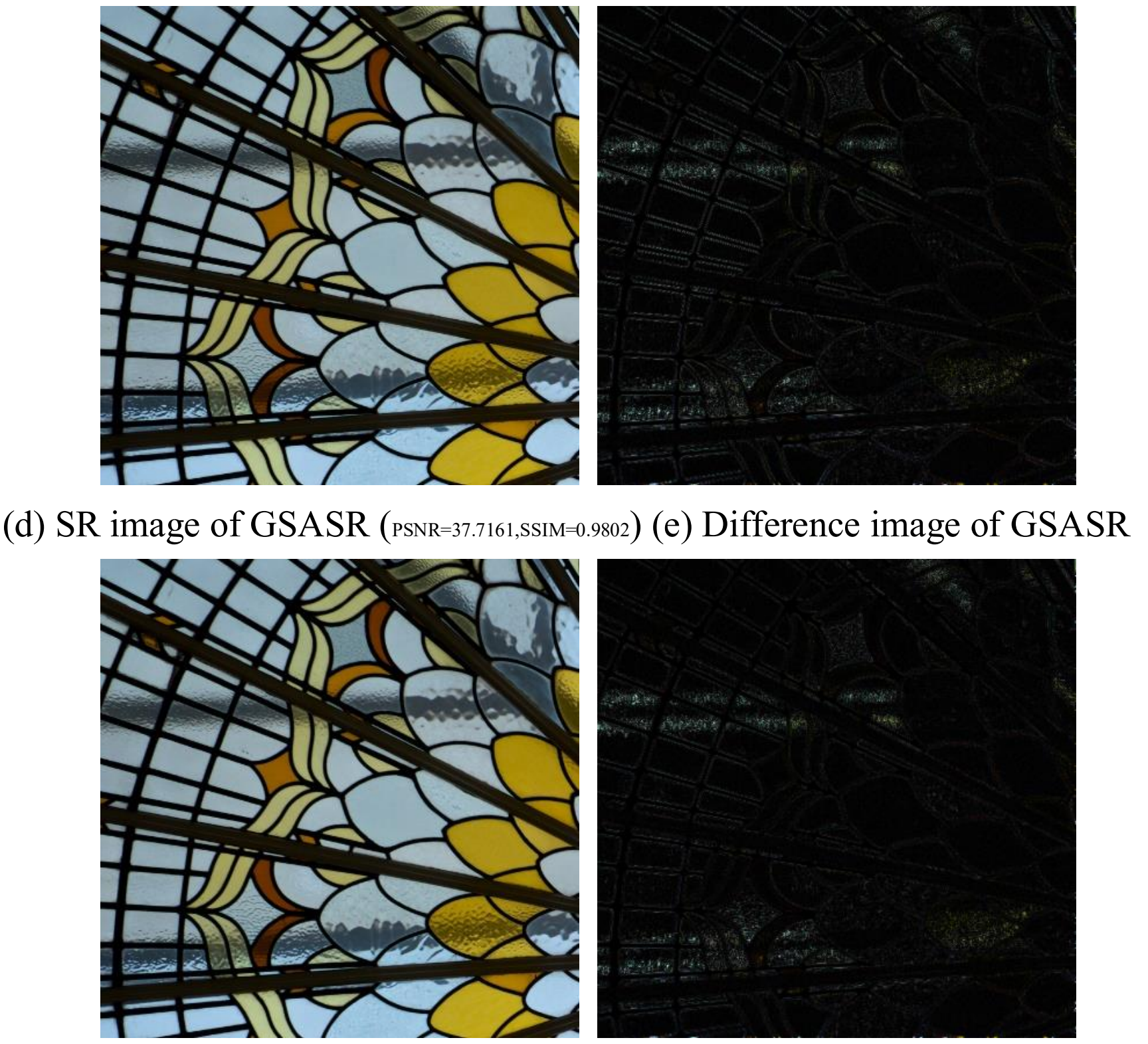

(d) SR image of GSASR (PSNR=37.7161,SSIM=0.9802) (e) Difference image of GSASR

(f) SR image of CGSASR (PSNR=38.7542,SSIM=0.9813) (g) Difference image of CGSASR

**Fig. 16. The results at scale factor 2 with maximum PSNR increment 1.0381 dB.**

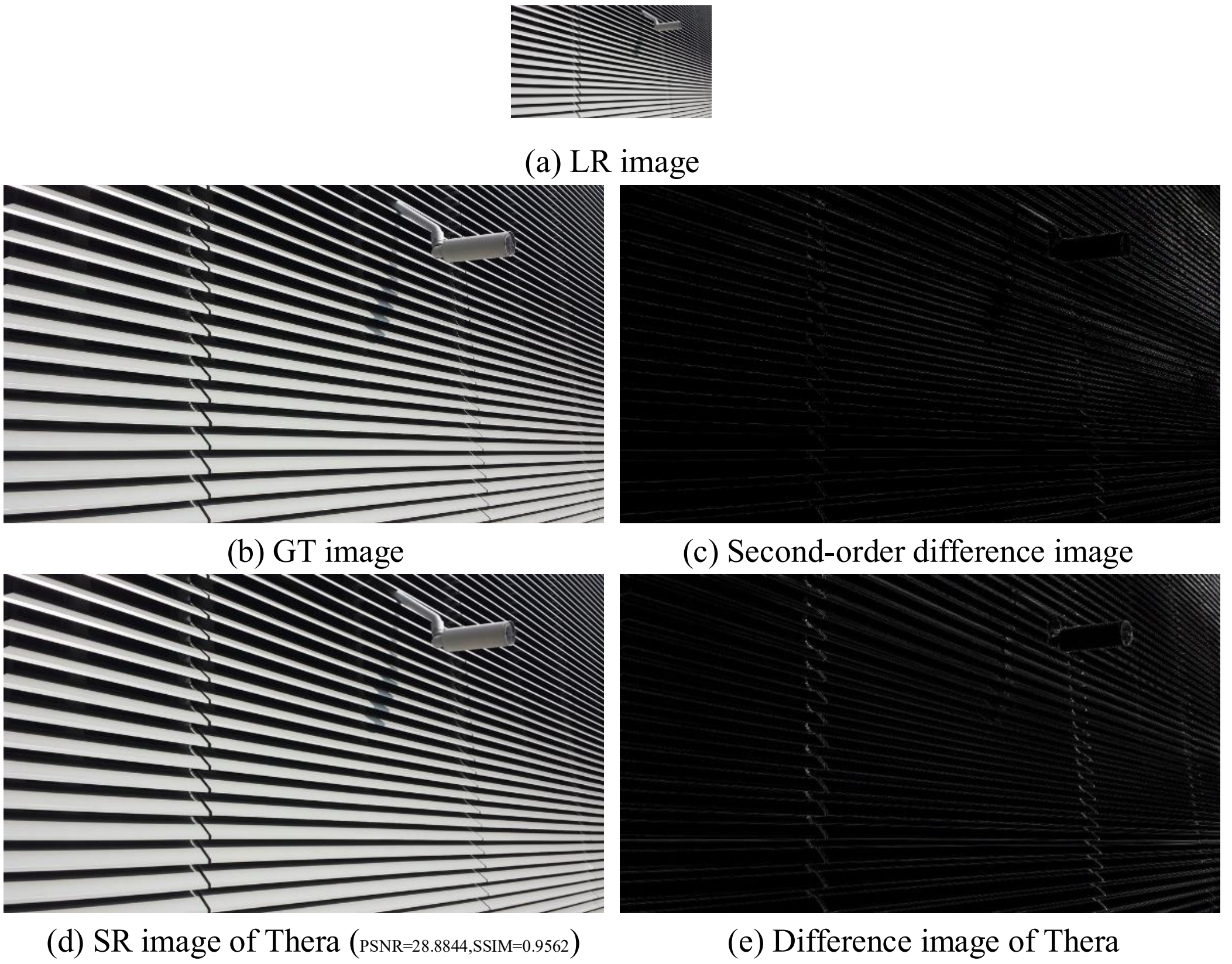

(a) LR image

(b) GT image (c) Second-order difference image

(d) SR image of Thera (PSNR=28.8844,SSIM=0.9562) (e) Difference image of Thera

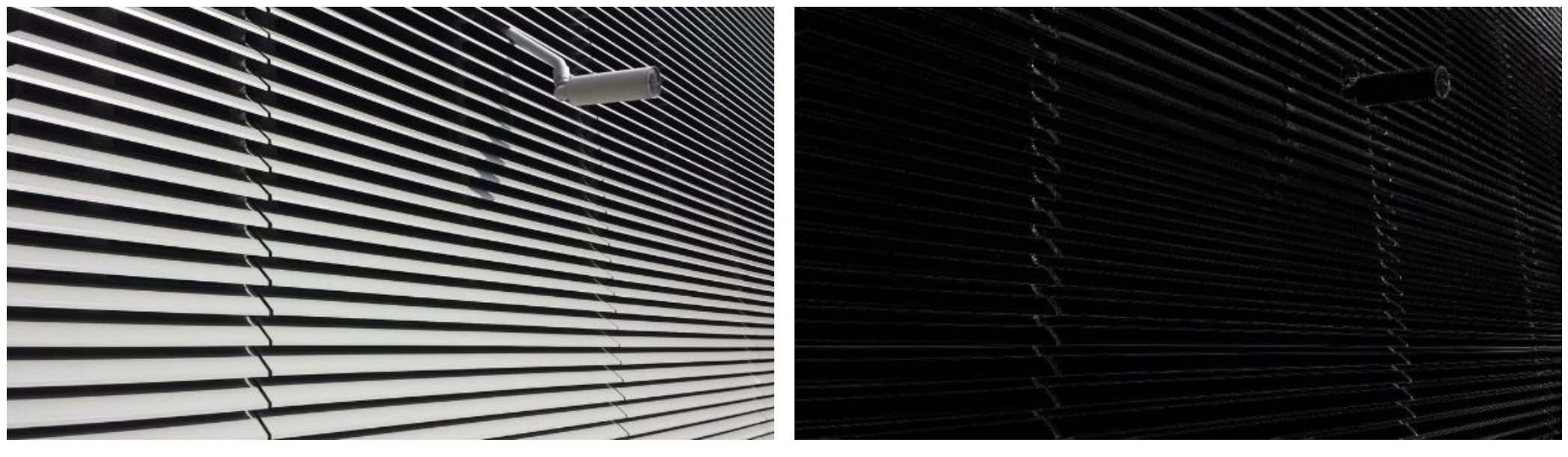

(f) SR image of CThera (PSNR=29.3615,SSIM=0.9608) (g) Difference image of CThera

**Fig. 17. The results at scale factor 3 with maximum PSNR increment 0.4771 dB.**

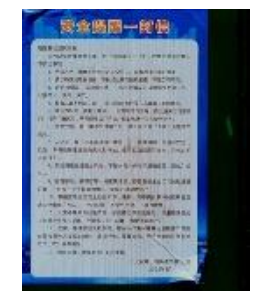

(a) LR image

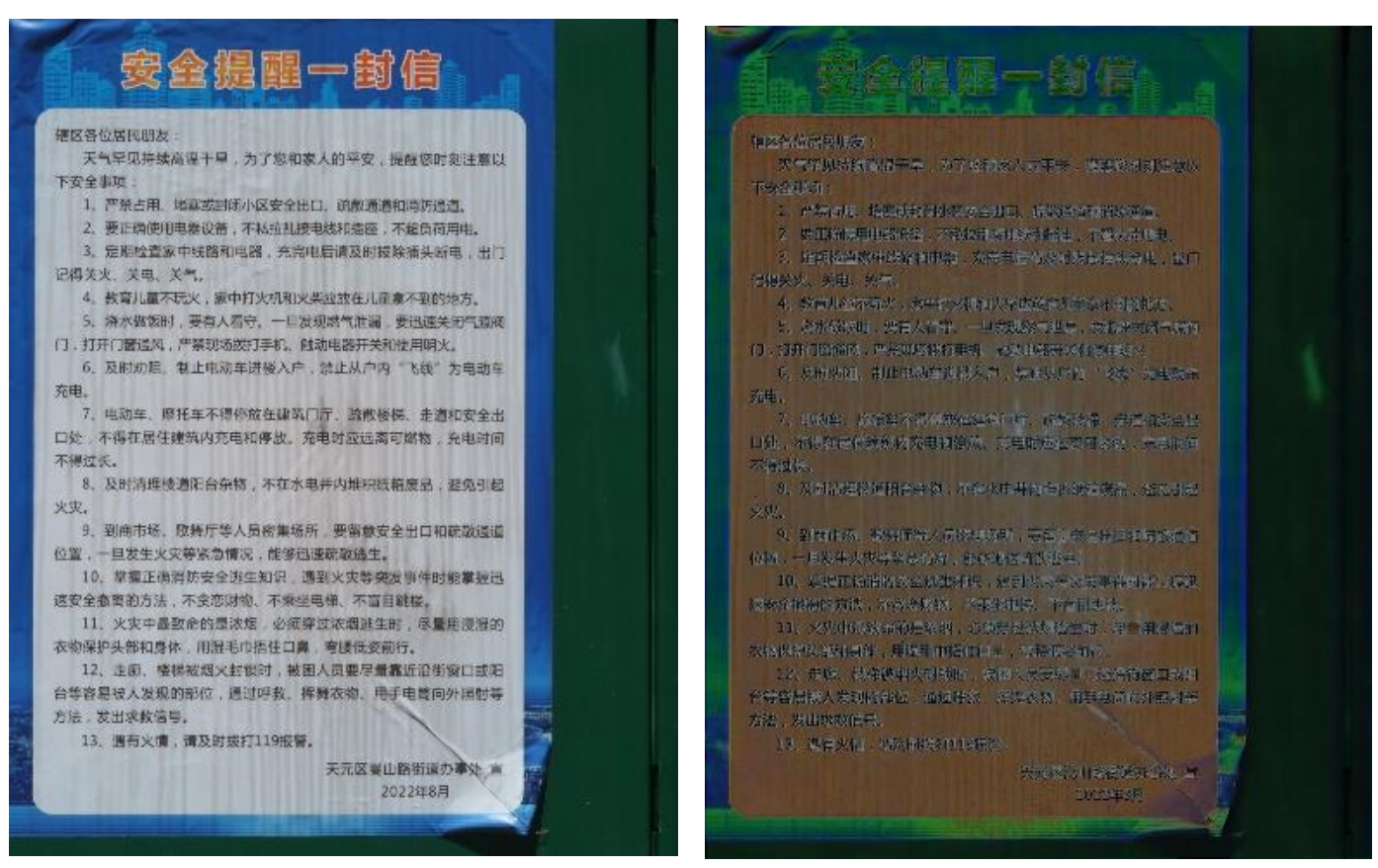

(b) GT image (c) Second-order difference image

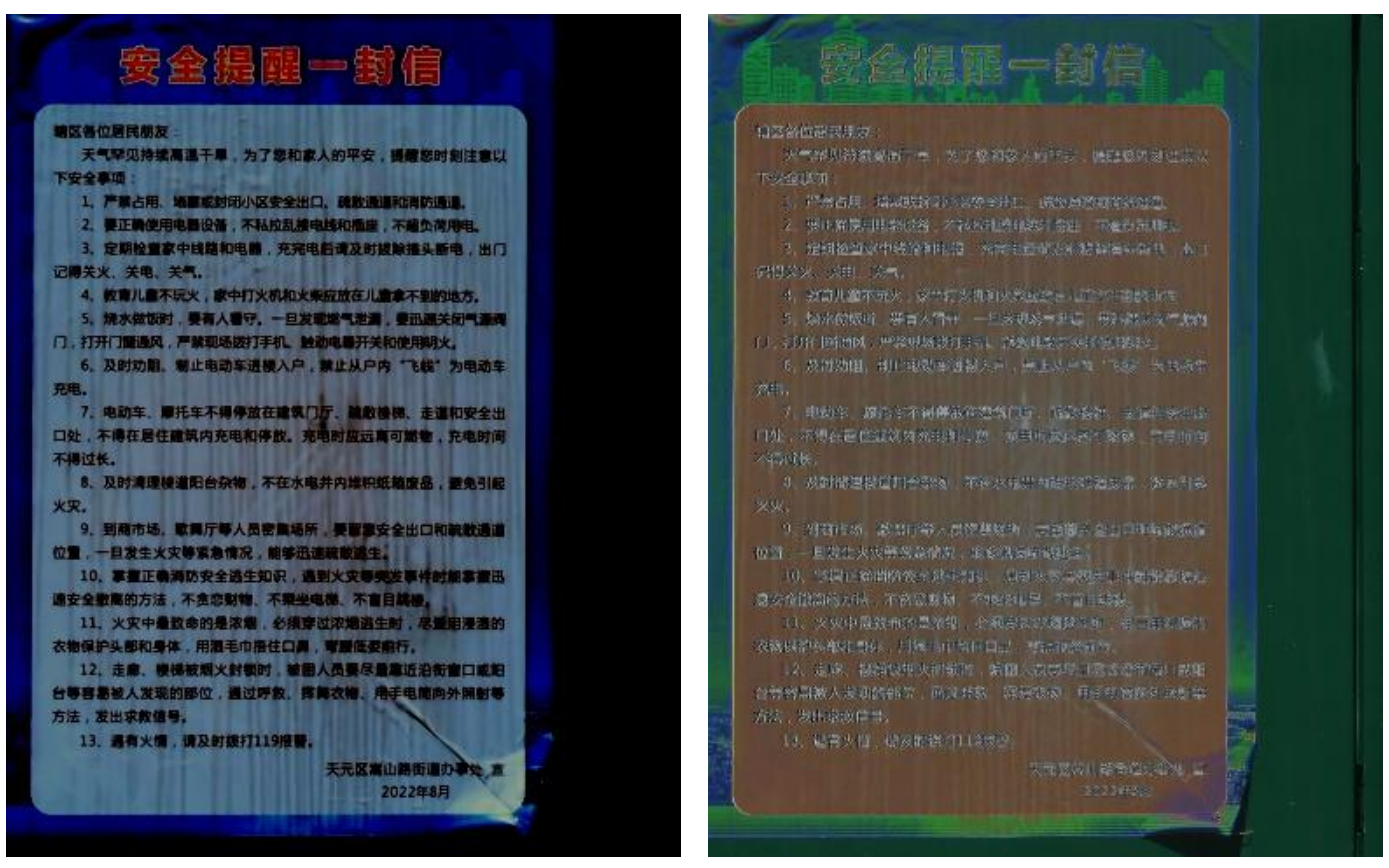

(d) SR image of COZ (PSNR=21.7322,SSIM=0.9084) (e) Difference image of COZ

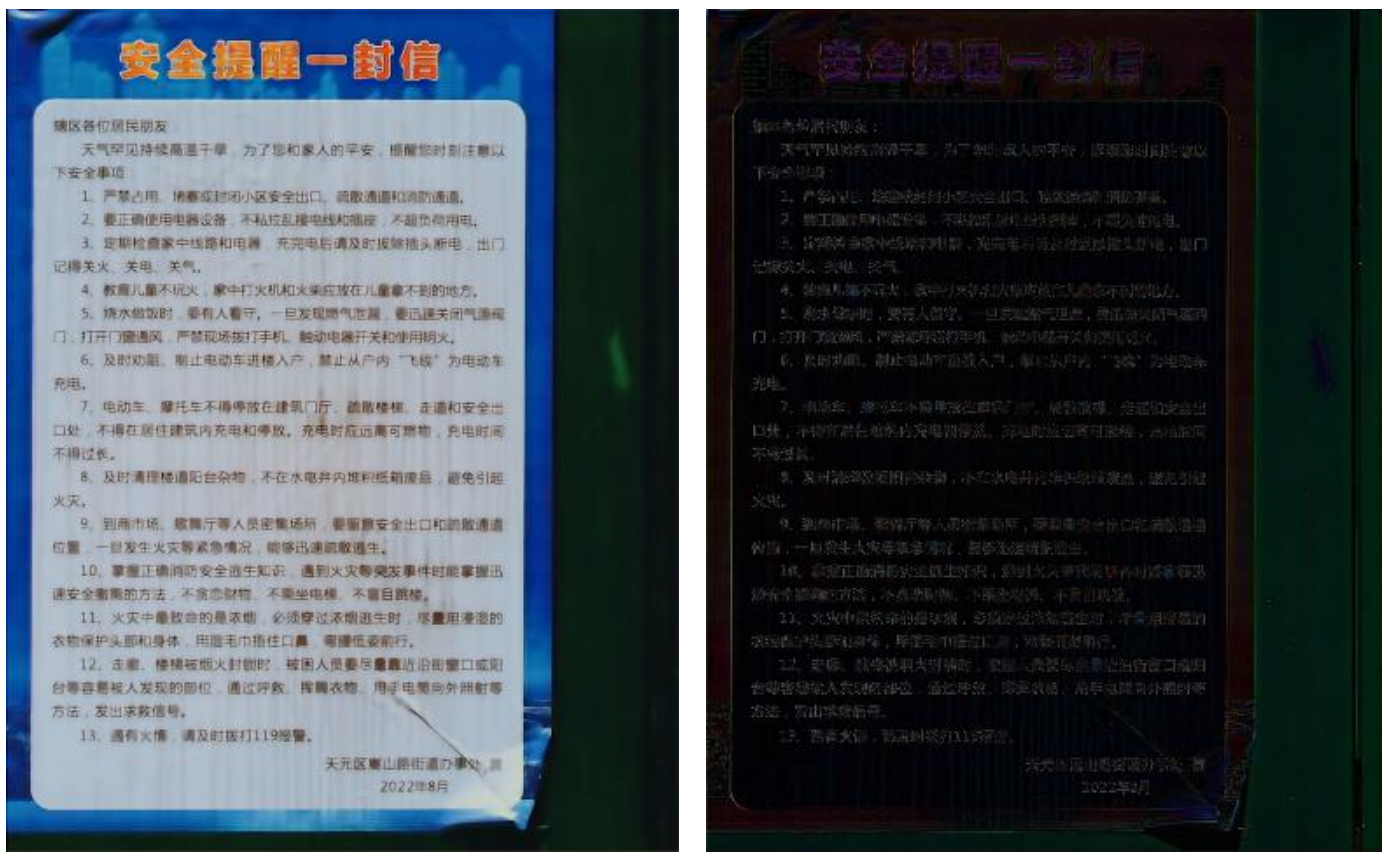

(f) SR image of CCOZ (PSNR=22.6451,SSIM=0.9109) (g) Difference image of CCOZ

**Fig. 18. The results at scale factor 3 with maximum PSNR increment 0.9129 dB.**

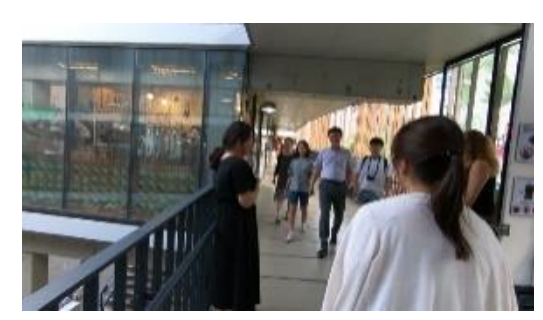

(a) LR image

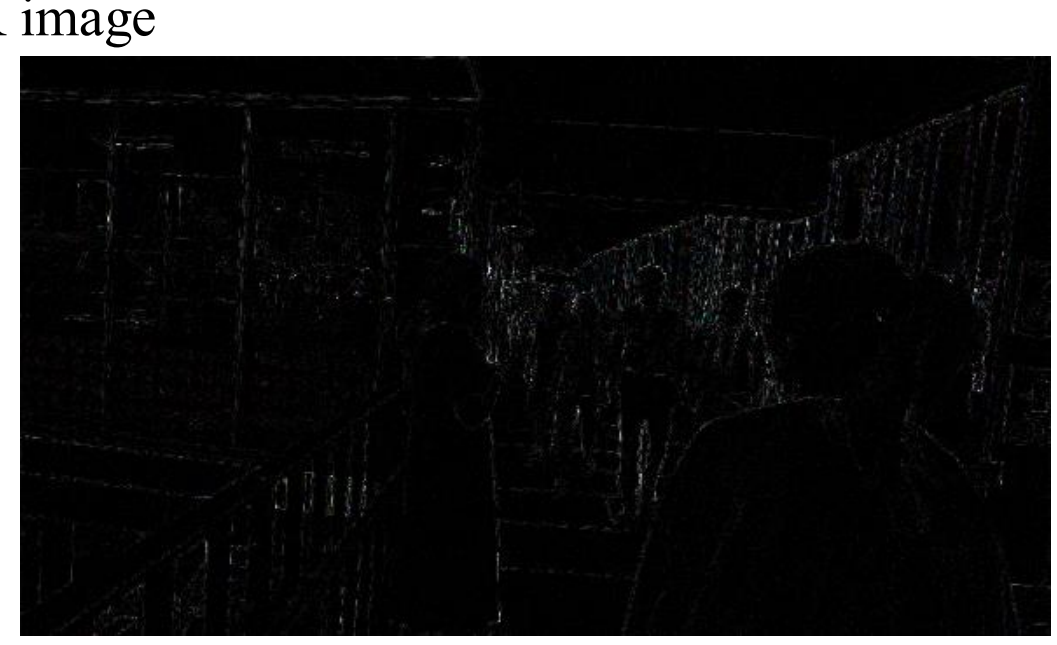

(b) GT image (c) Second-order difference image

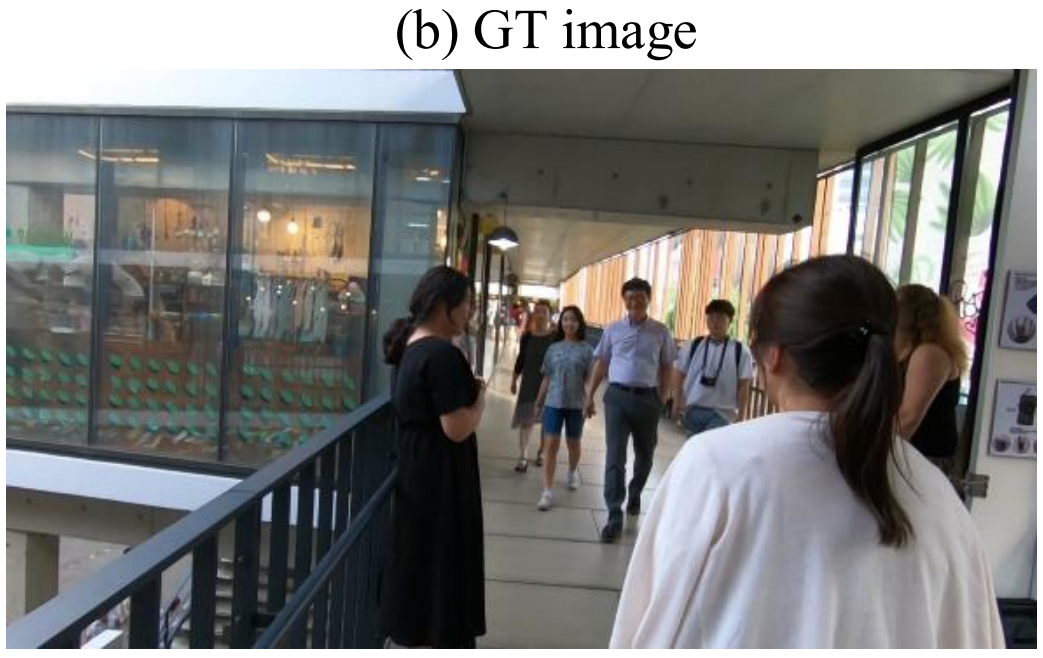

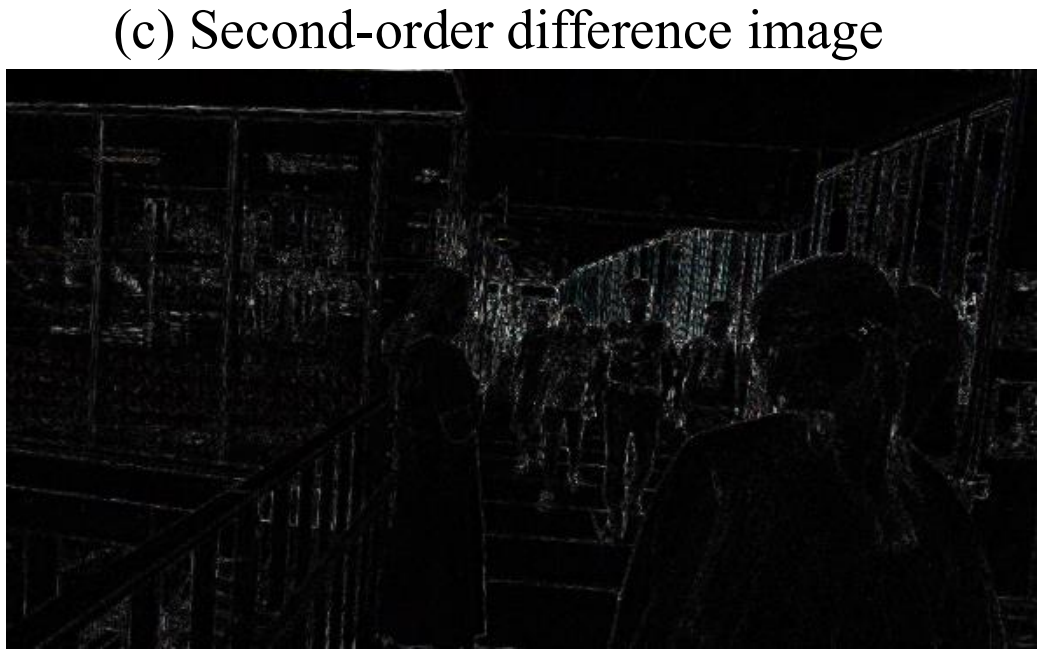

(d) SR image of CBasicAVSR (PSNR=36.5318,SSIM=0.9750) (e) Difference image of CBasicAVSR

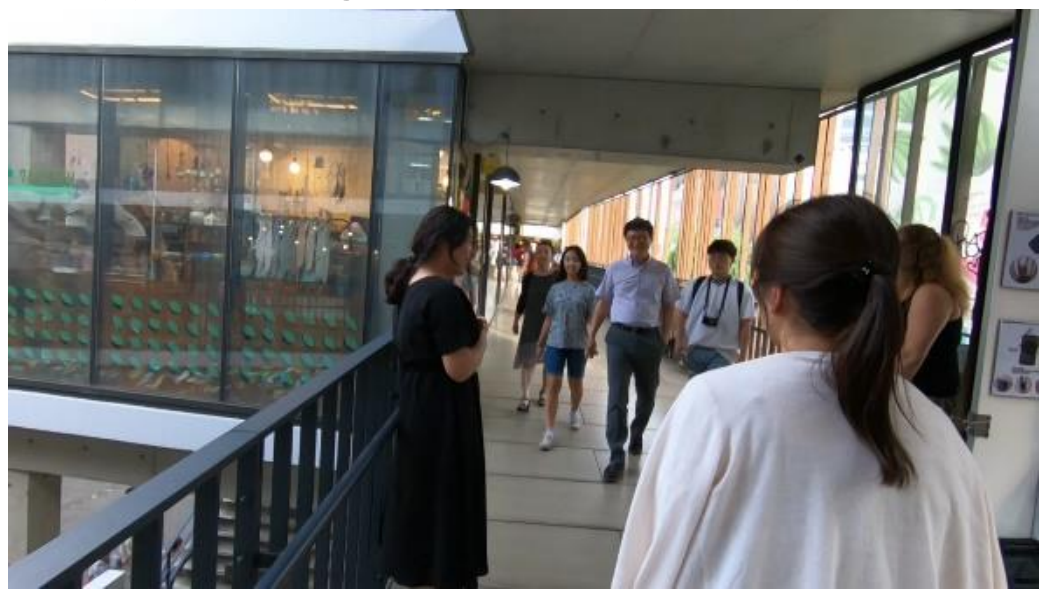

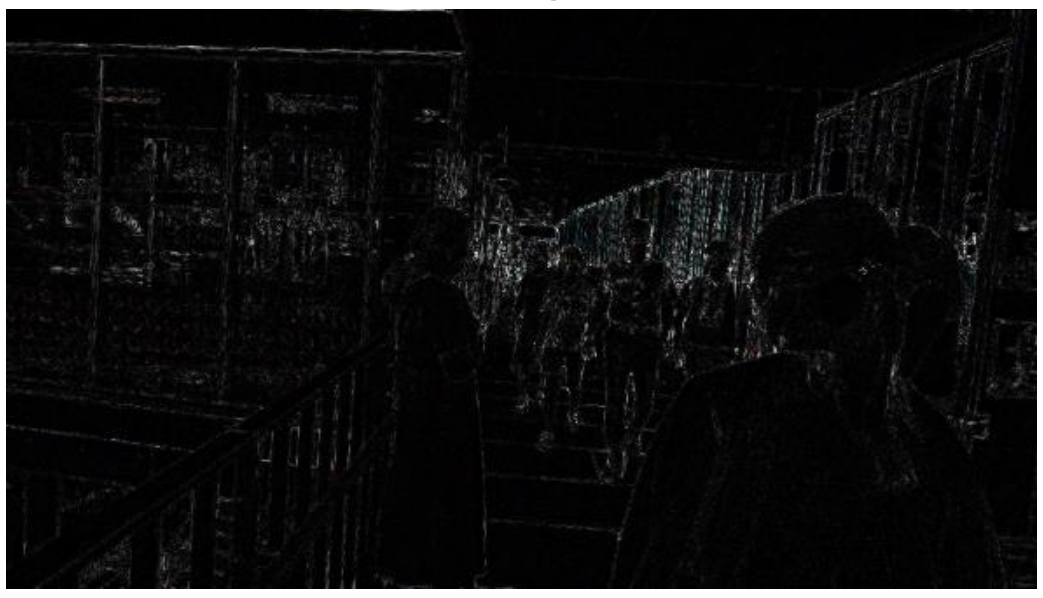

(f) SR image of CBasicAVSR (PSNR=36.7029,SSIM=0.9755) (g) Difference image of CBasicAVSR

**Fig. 19. The results at scale factor 3 with maximum PSNR increment 0.1771 dB.**

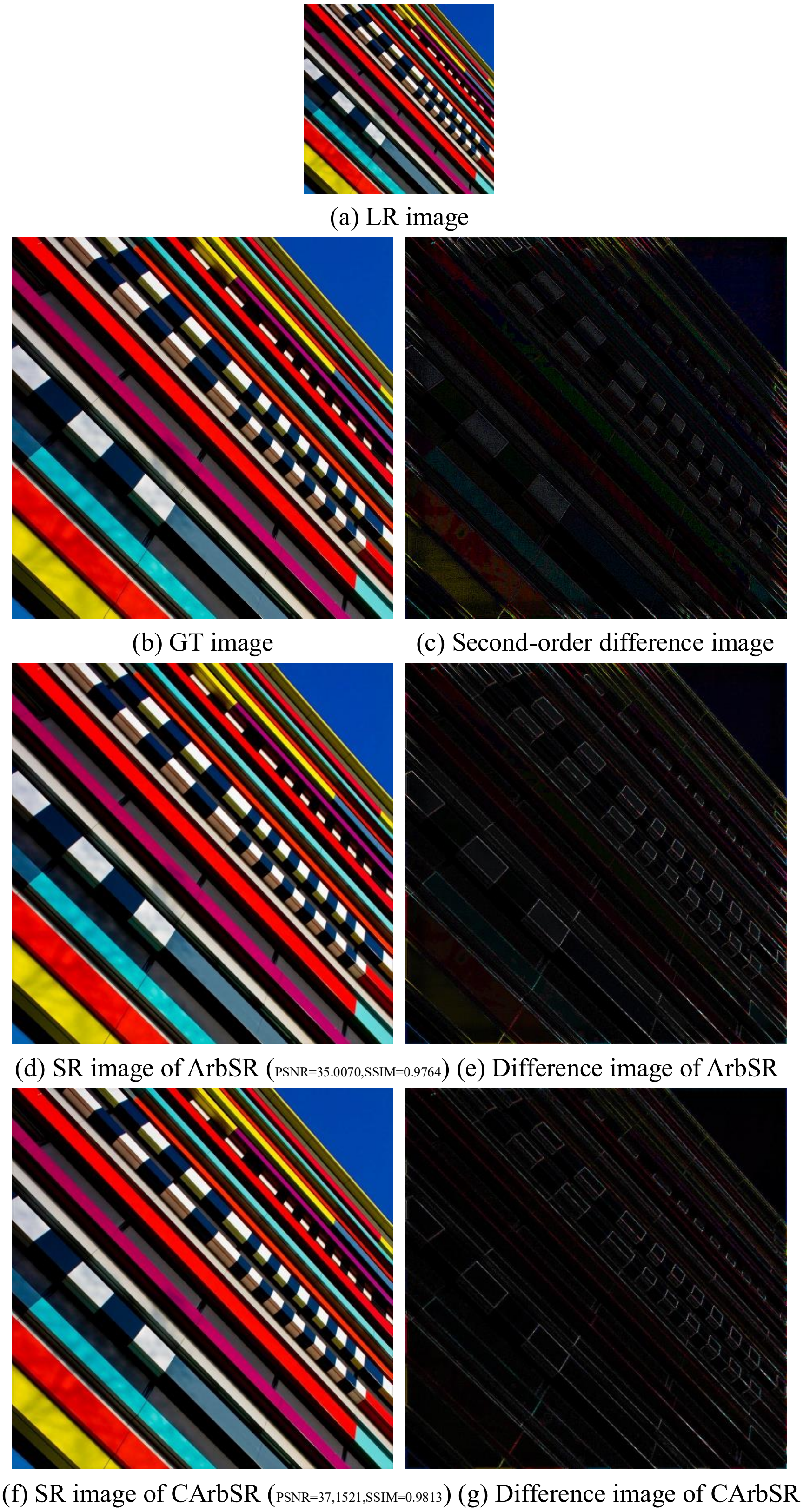

(a) LR image

(b) GT image (c) Second-order difference image

(d) SR image of ArbSR (PSNR=35.0070,SSIM=0.9764) (e) Difference image of ArbSR

(f) SR image of CArbSR (PSNR=37,1521,SSIM=0.9813) (g) Difference image of CArbSR

**Fig. 20. The results at scale factor 3 with maximum PSNR increment 2.1451 dB.**

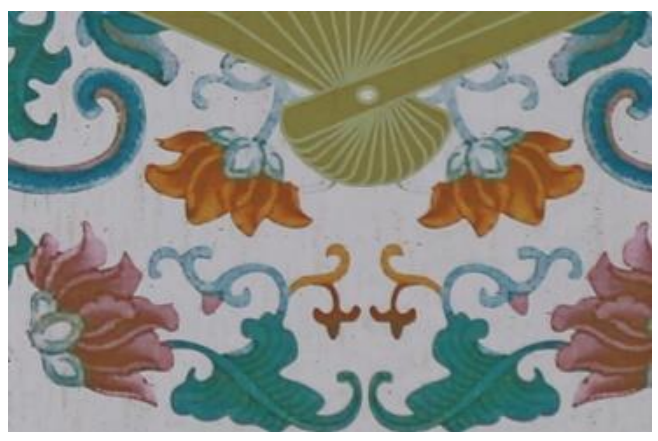

(a) LR image

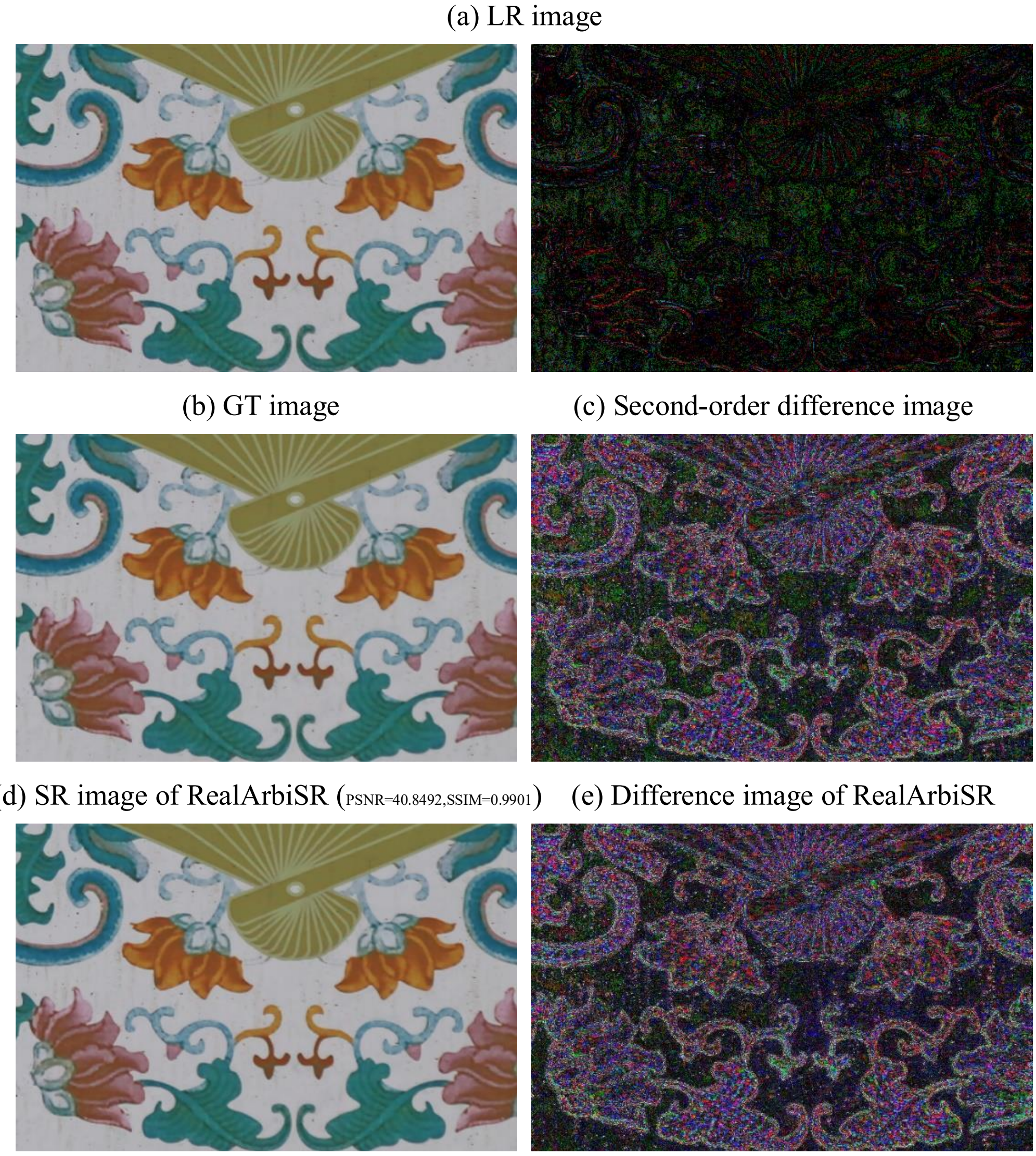

(b) GT image (c) Second-order difference image

(d) SR image of RealArbiSR (PSNR=40.8492,SSIM=0.9901) (e) Difference image of RealArbiSR

(f) SR image of CRealArbiSR (PSNR=40.9992,SSIM=0.9903) (g) Difference image of CRealArbiSR

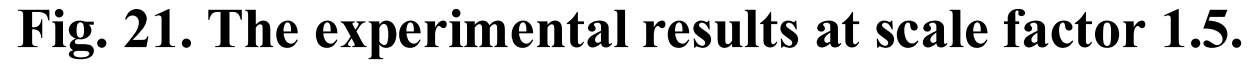

**Fig. 21. The experimental results at scale factor 1.5.**

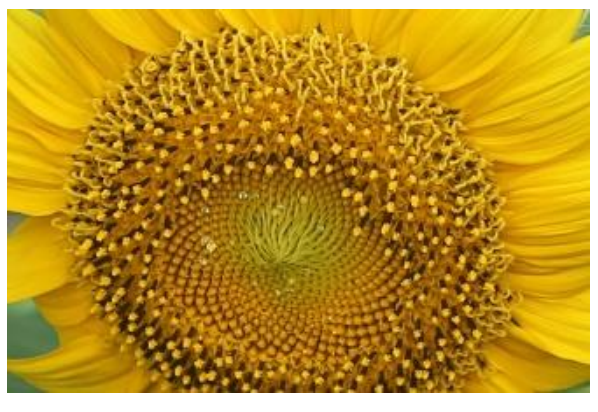

(a) LR image

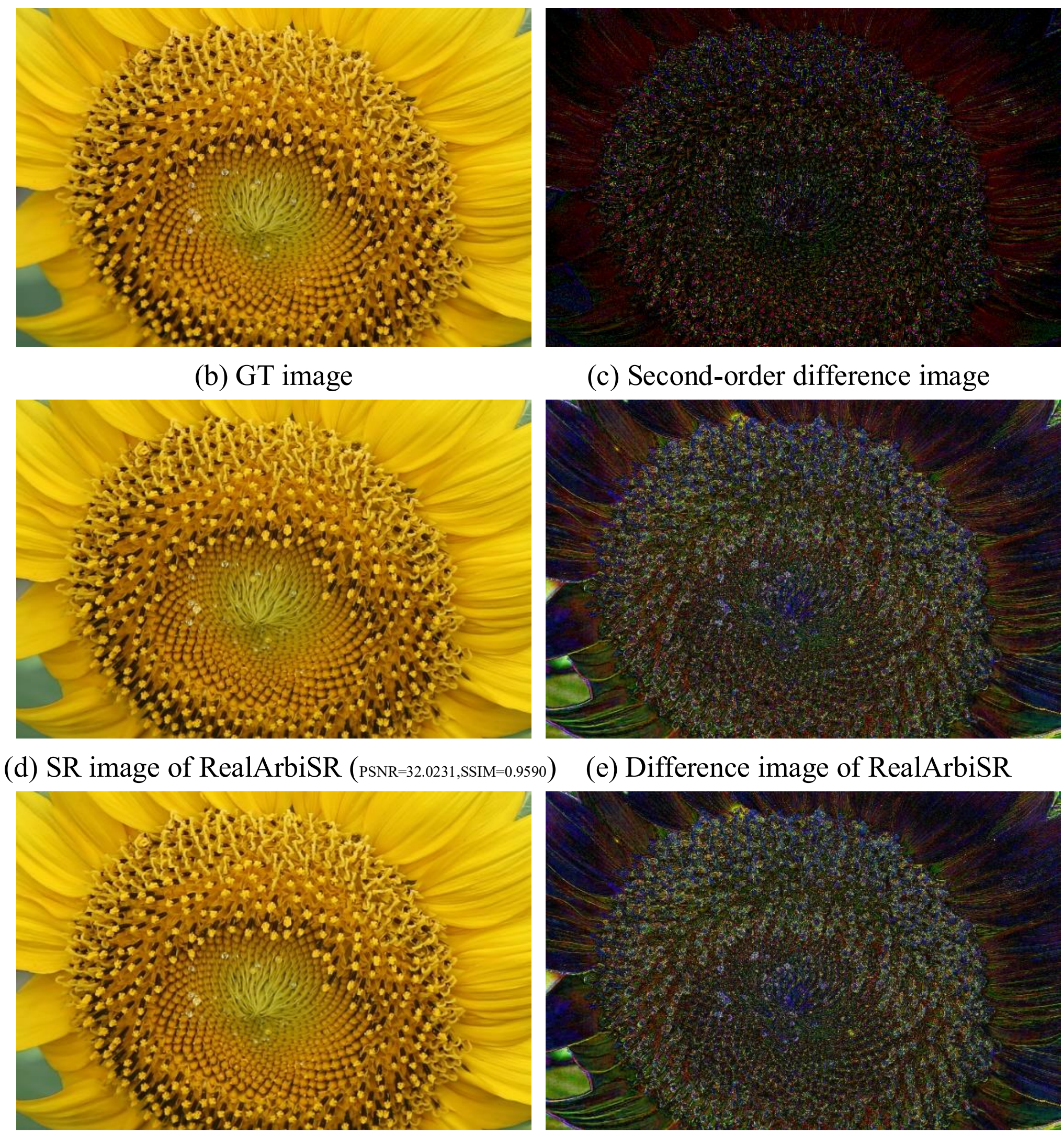
(b) GT image (c) Second-order difference image

(d) SR image of RealArbiSR (PSNR=32.0231,SSIM=0.9590) (e) Difference image of RealArbiSR

(f) SR image of CRealArbiSR (PSNR=32.1448,SSIM=0.9590) (g) Difference image of CRealArbiSR

**Fig. 22. The experimental results at scale factor 1.7.**

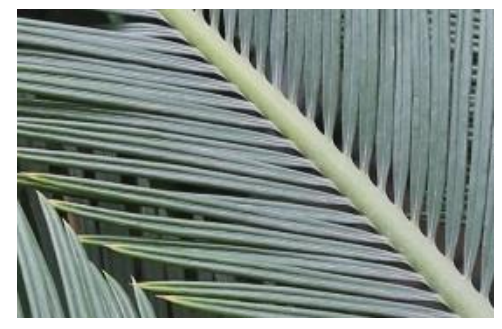
(a) LR image

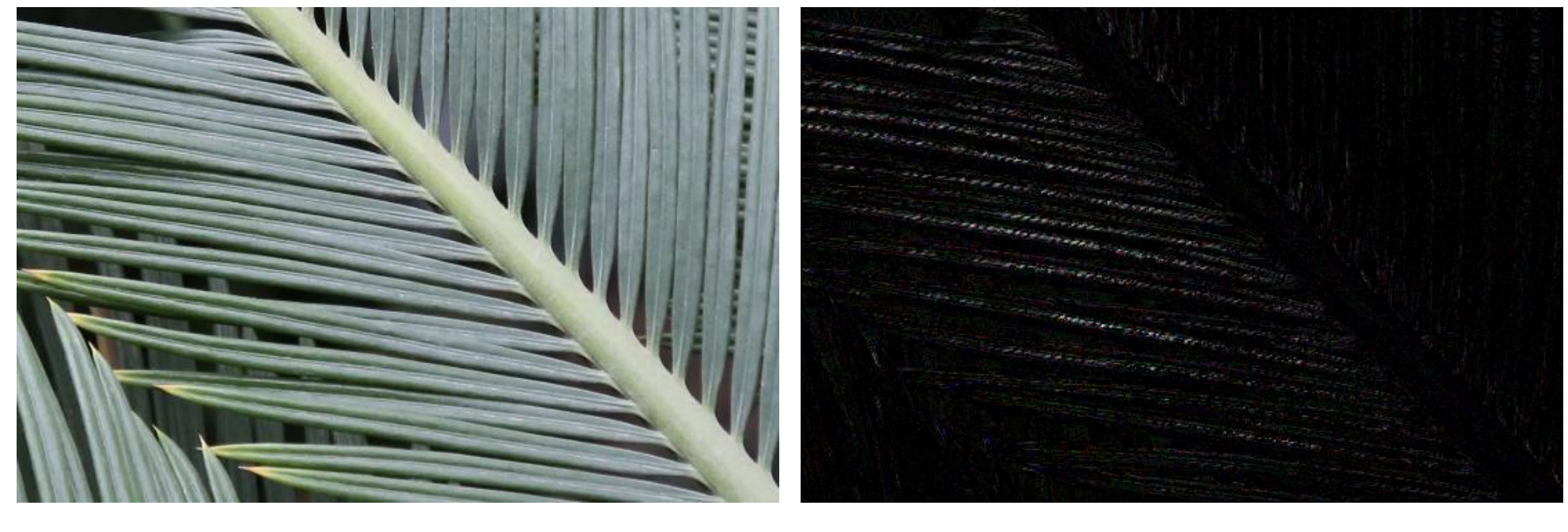
(b) GT image (c) Second-order difference image

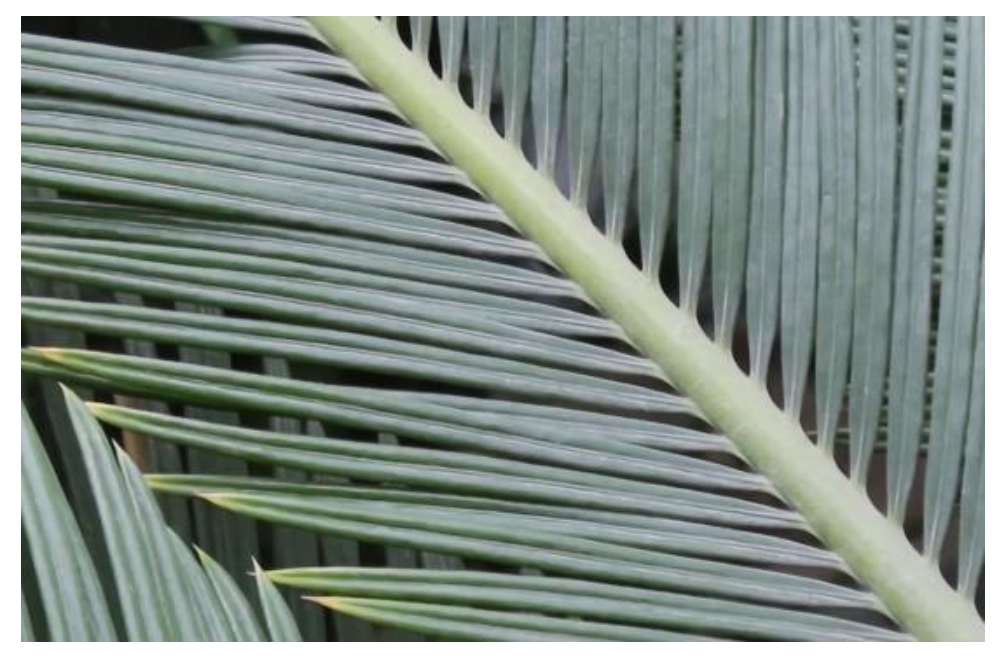

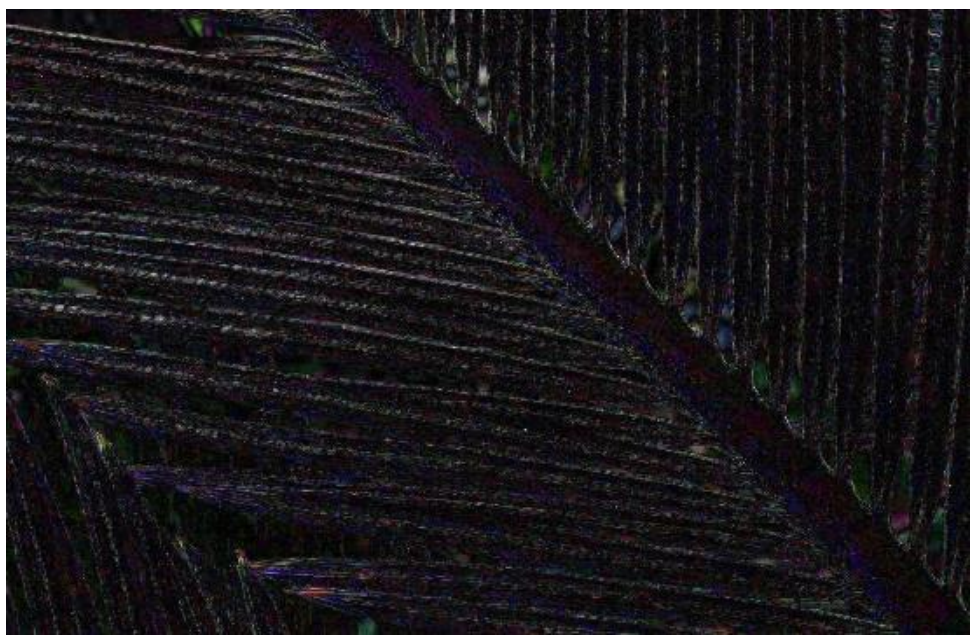

(d) SR image of RealArbiSR (PSNR=31.6568,SSIM=0.9798) (e) Difference image of RealArbiSR

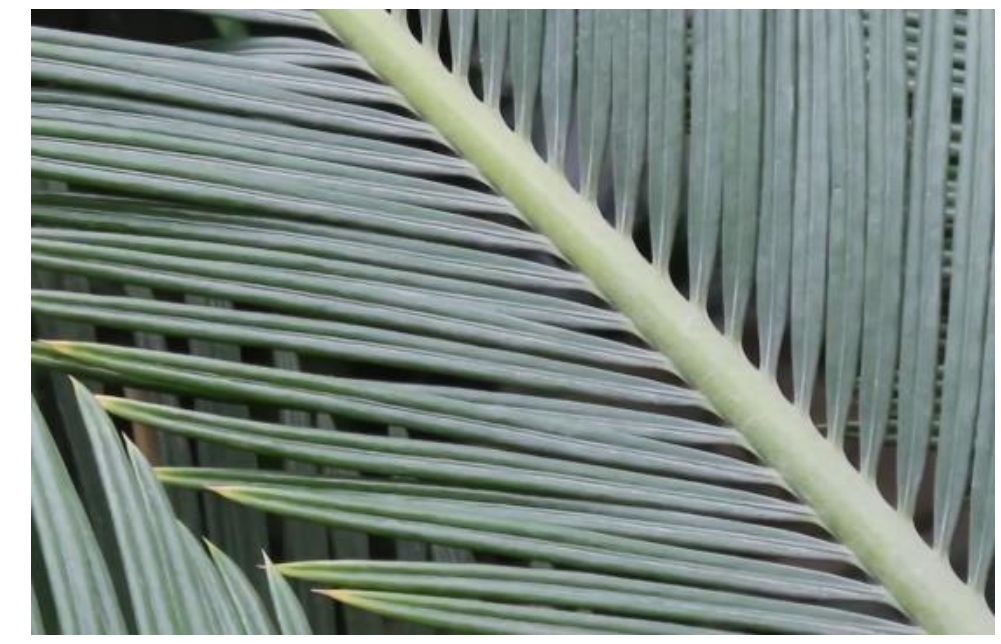

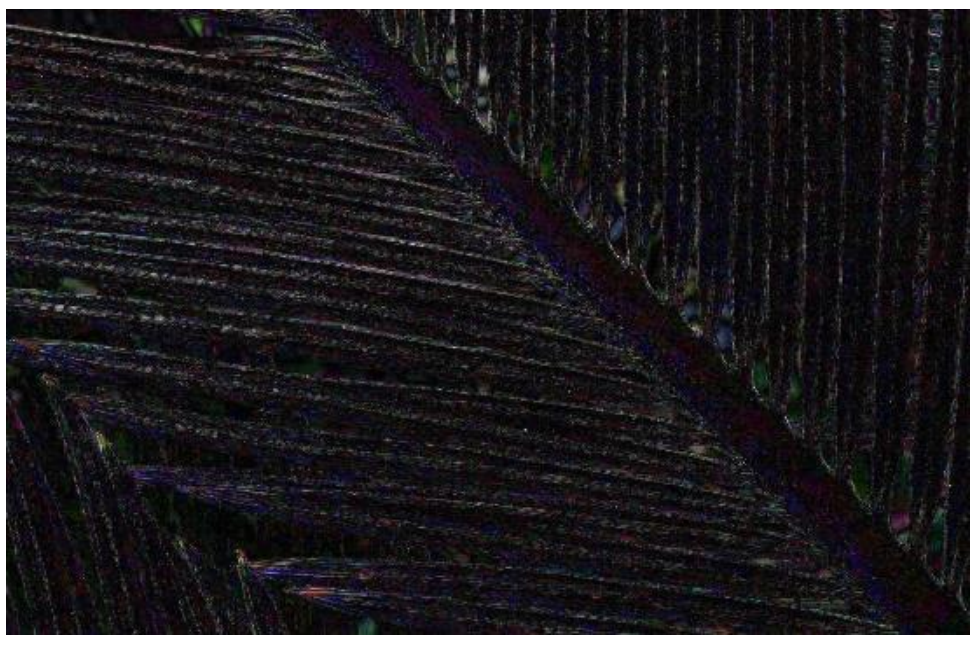

(f) SR image of CRealArbiSR (PSNR=31.7332,SSIM=0.9800) (g) Difference image of CRealArbiSR

**Fig. 23. The experimental results at scale factor 2.0.**

(a) LR image

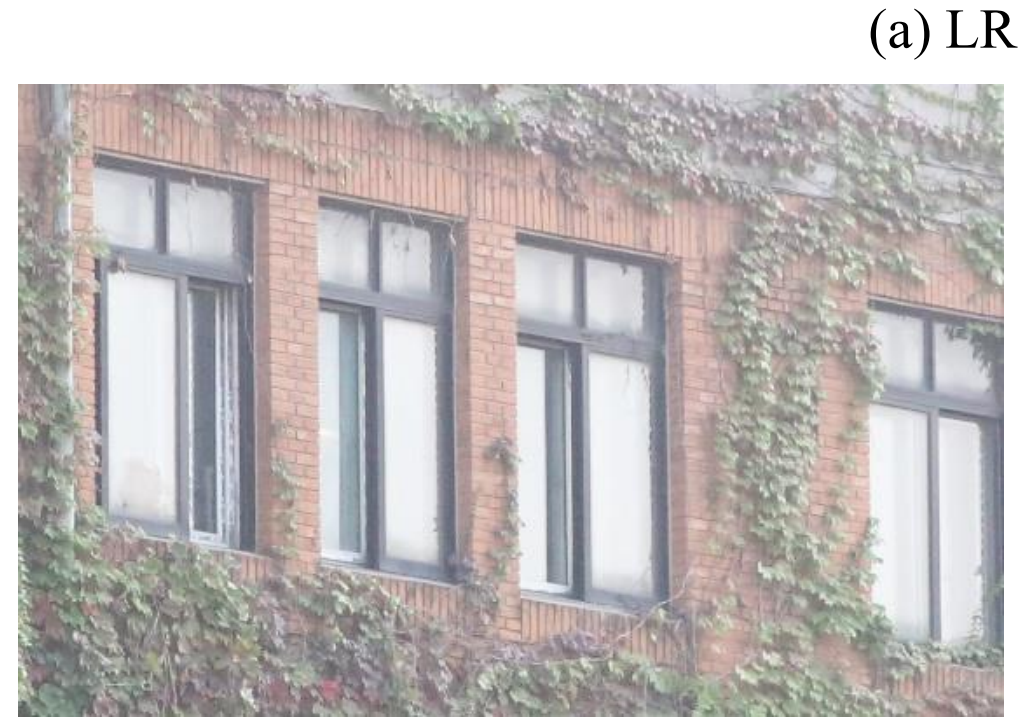

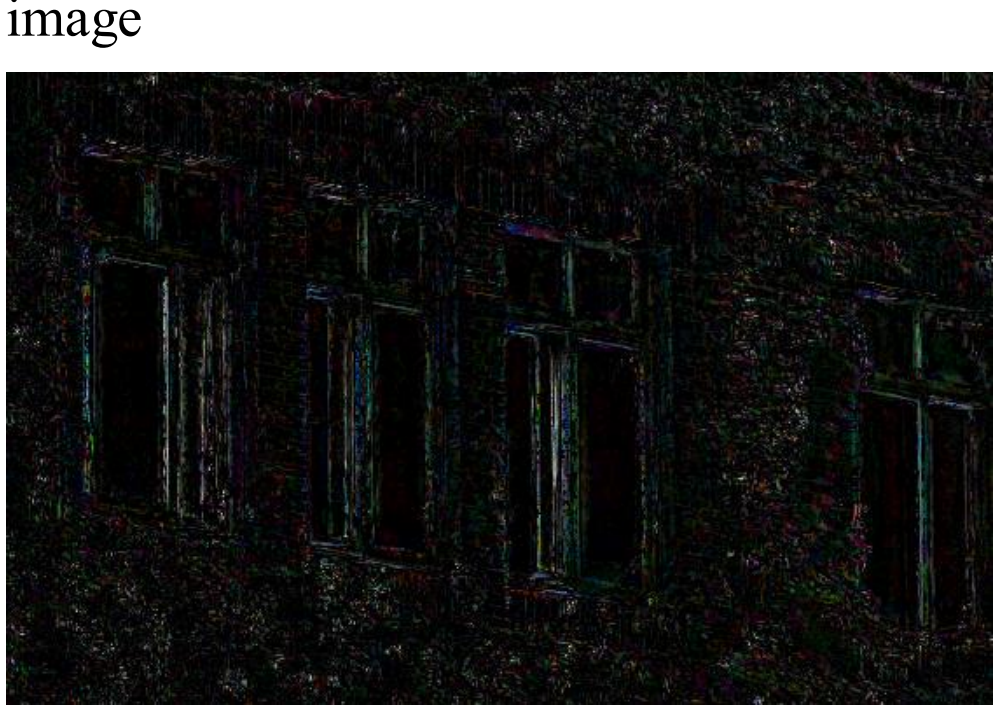

(b) GT image (c) Second-order difference image

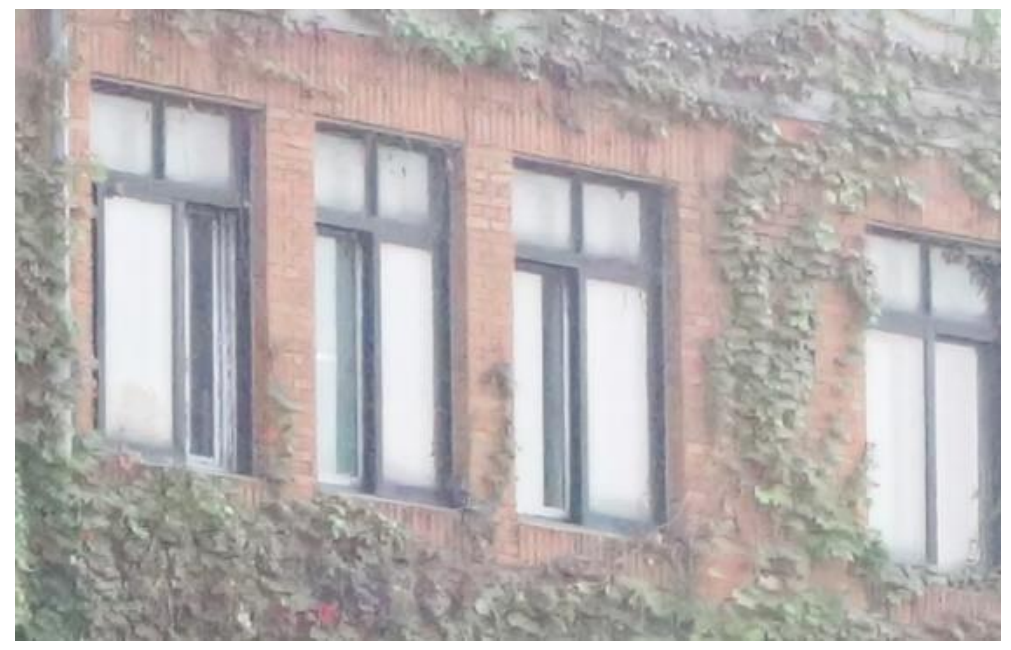

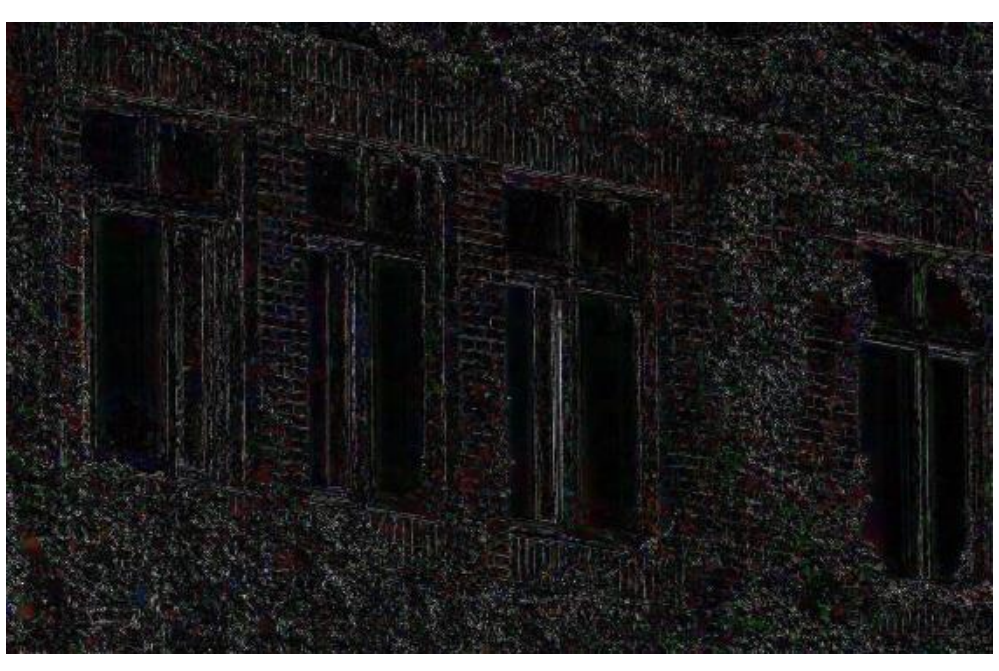

(d) SR image of RealArbiSR (PSNR=27.6414,SSIM=0.8667) (e) Difference image of RealArbiSR

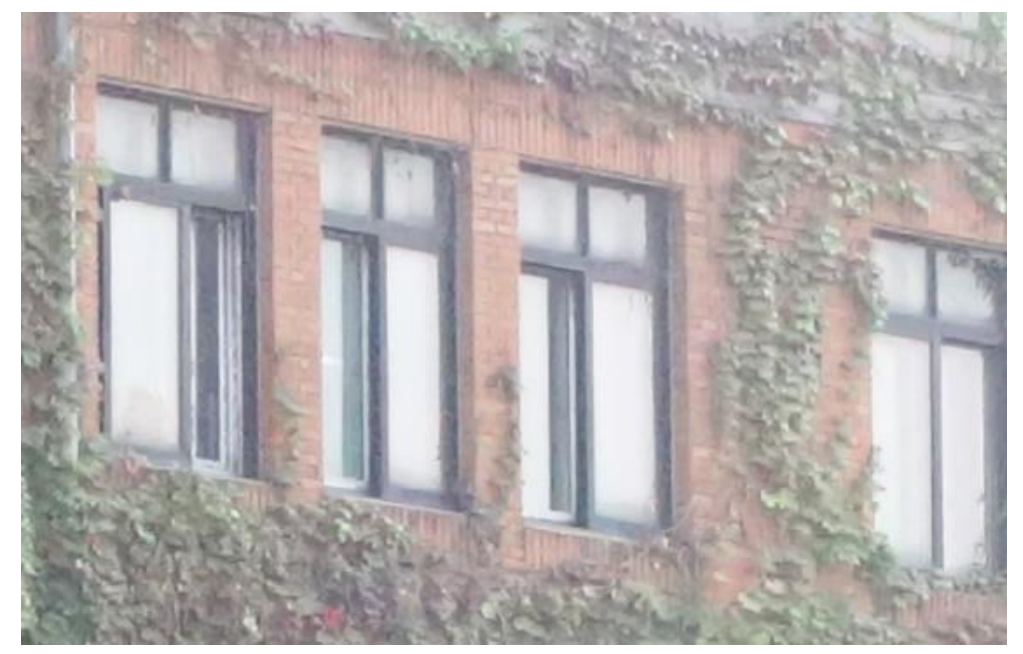

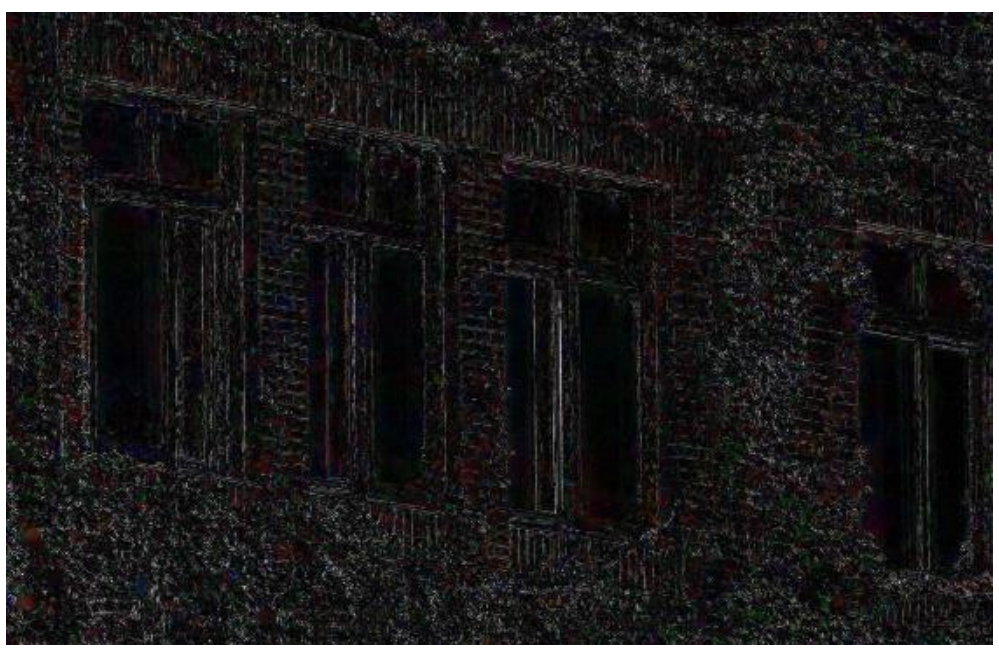

(f) SR image of CRealArbiSR (PSNR=27.7192,SSIM=0.8706) (g) Difference image of CRealArbiSR

**Fig. 24. The experimental results at scale factor 2.3.**

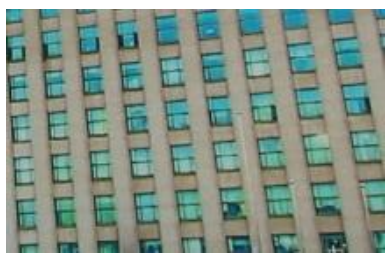

(a) LR image

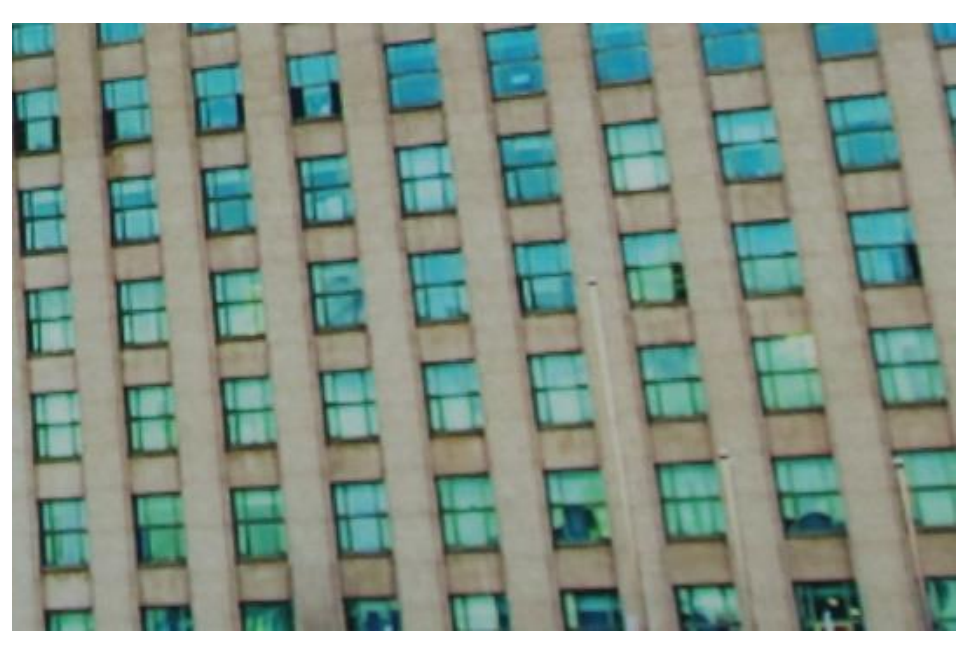

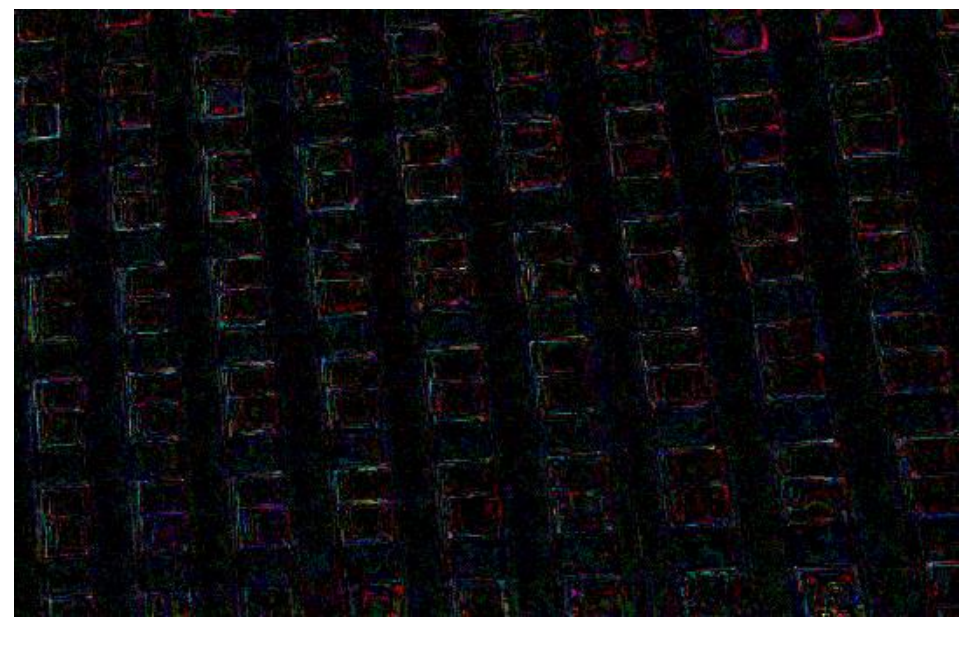

(b) GT image (c) Second-order difference image

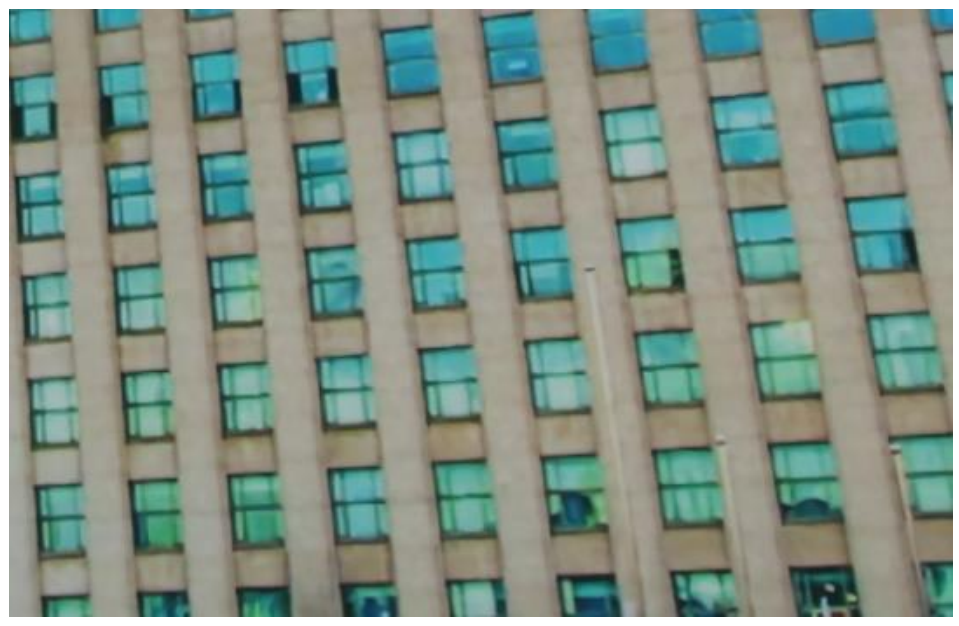

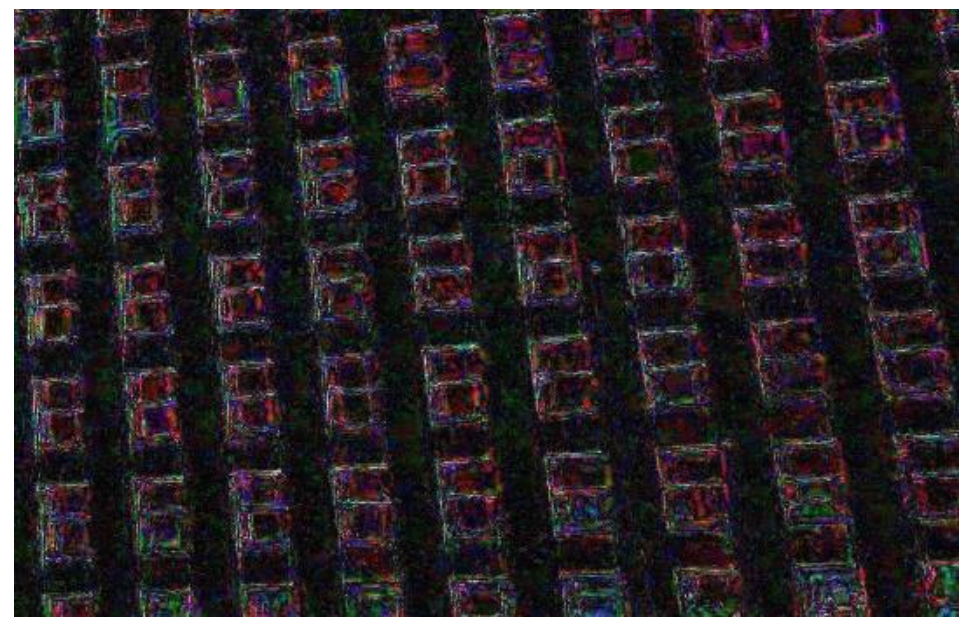

(d) SR image of RealArbiSR (PSNR=34.4944,SSIM=0.9739) (e) Difference image of RealArbiSR

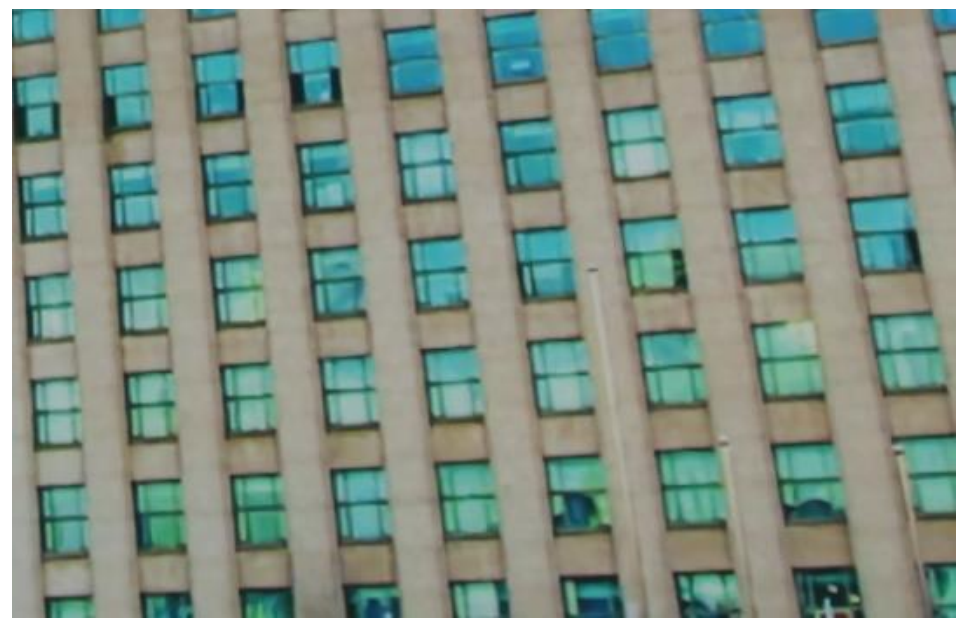

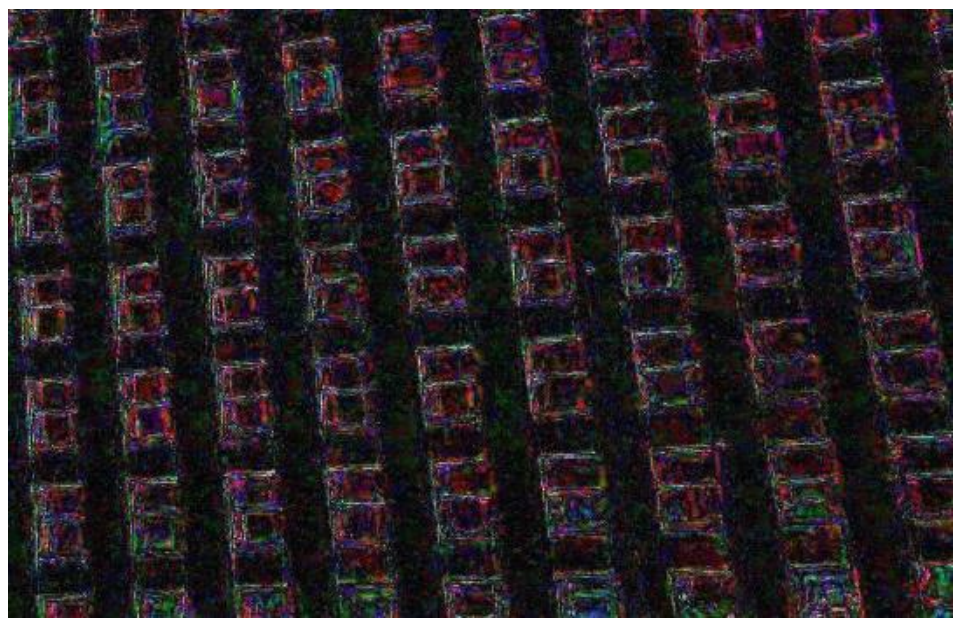

(f) SR image of CRealArbiSR (PSNR=34.5537,SSIM=0.9742) (g) Difference image of CRealArbiSR

**Fig. 25. The experimental results at scale factor 2.5.**

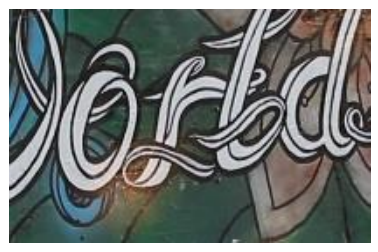

(a) LR image

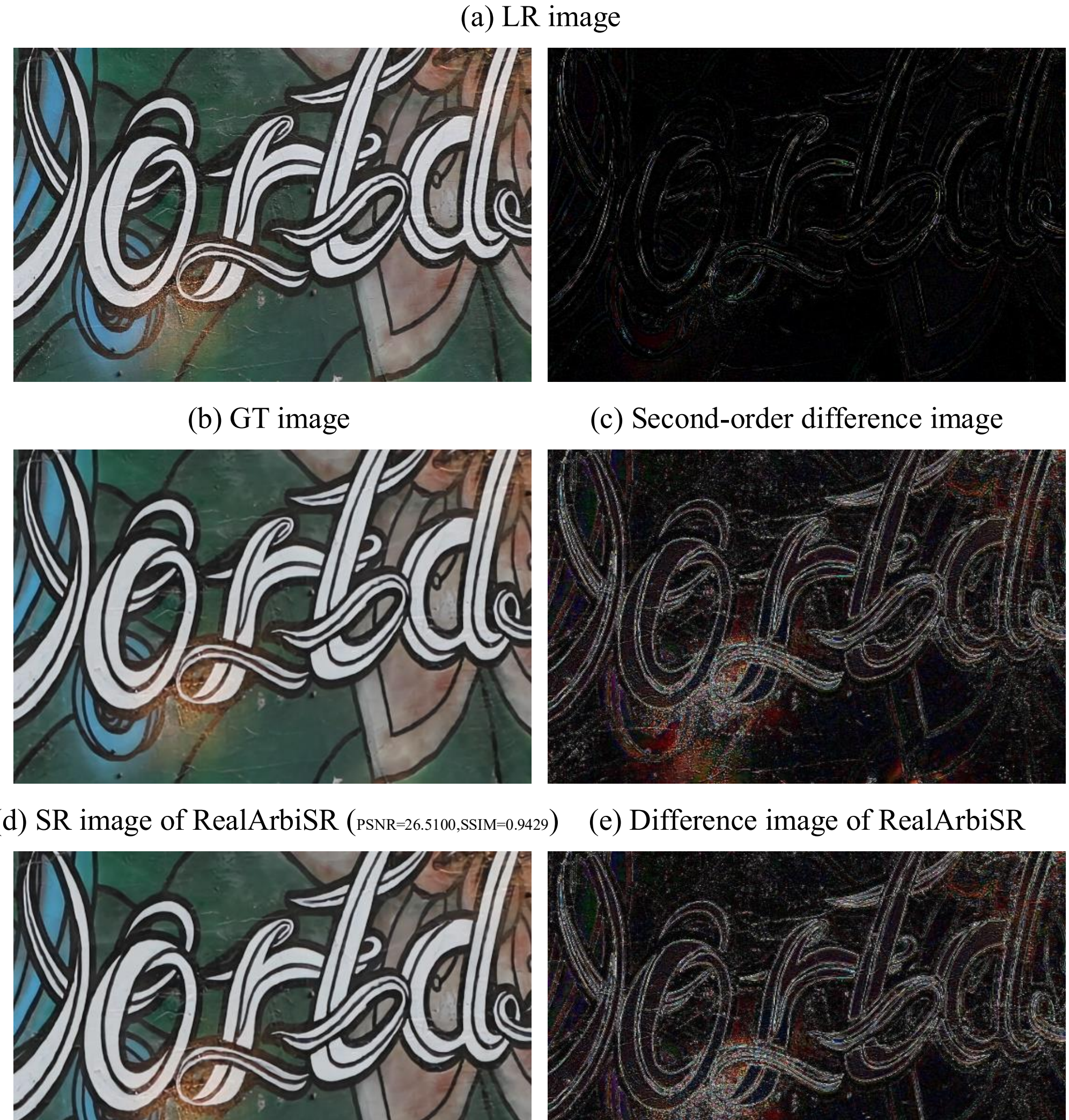

(b) GT image (c) Second-order difference image

(d) SR image of RealArbiSR (PSNR=26.5100,SSIM=0.9429) (e) Difference image of RealArbiSR

(f) SR image of CRealArbiSR (PSNR=26.6203,SSIM=0.9440) (g) Difference image of CRealArbiSR

**Fig. 26. The experimental results at scale factor 2.7.**

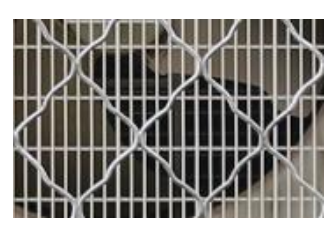

(a) LR image

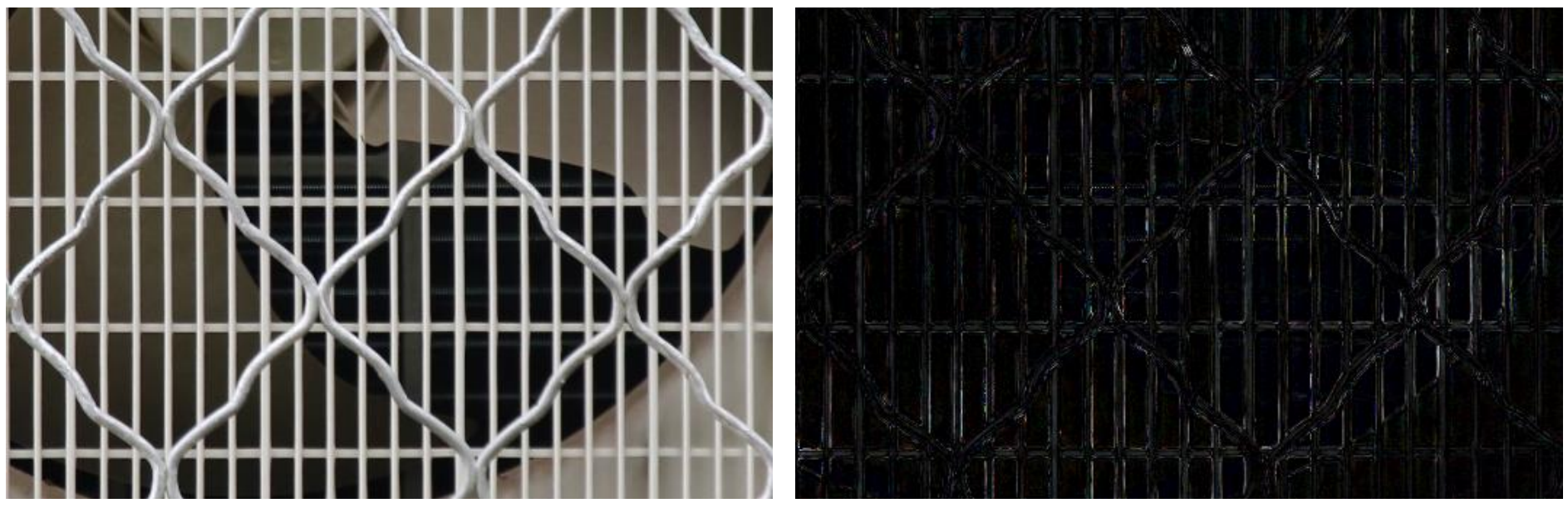

(b) GT image (c) Second-order difference image

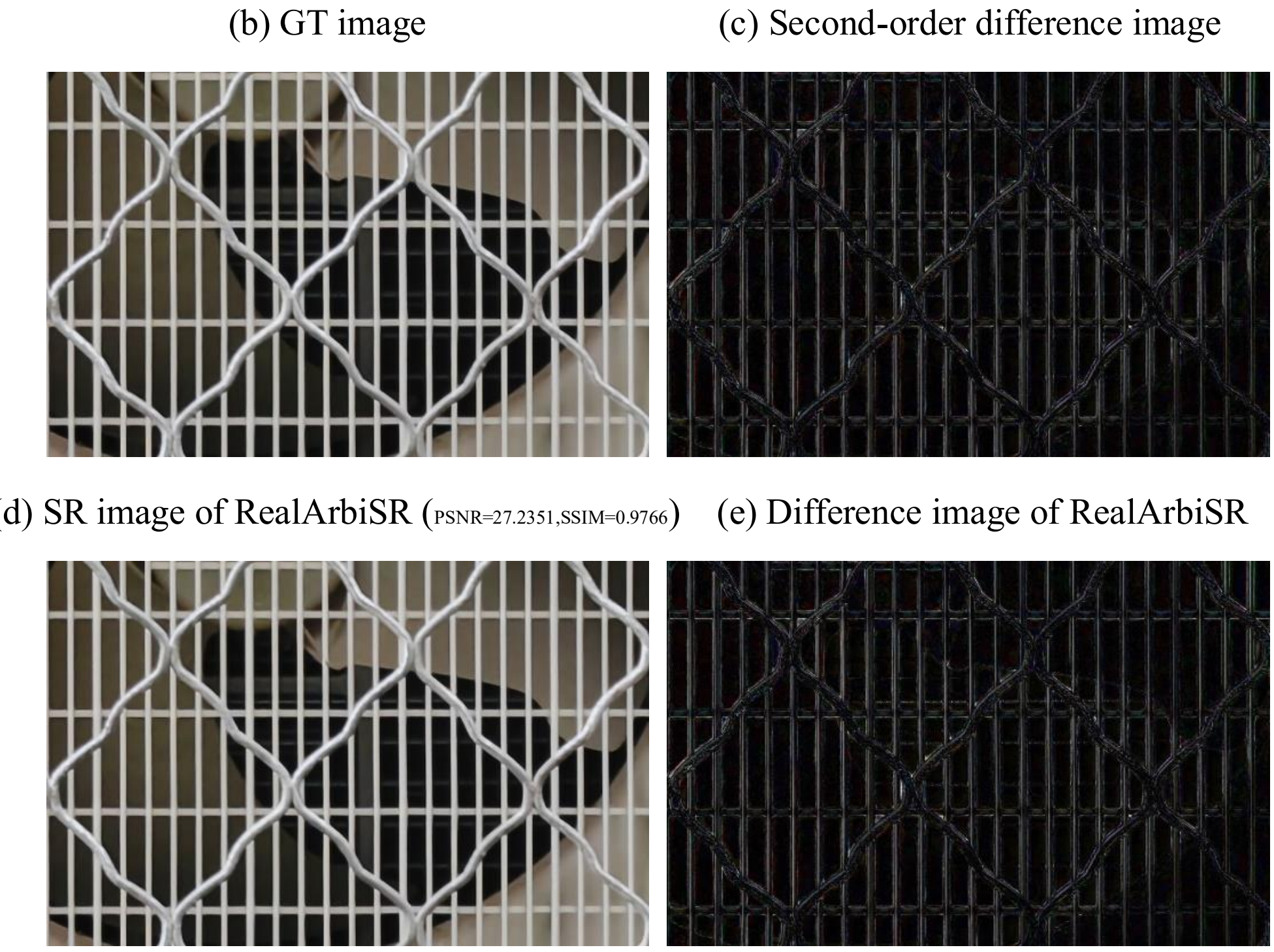

(d) SR image of RealArbiSR (PSNR=27.2351,SSIM=0.9766) (e) Difference image of RealArbiSR

(f) SR image of CRealArbiSR (PSNR=27.3526,SSIM=0.9766) (g) Difference image of CRealArbiSR

**Fig. 27. The experimental results at scale factor 3.0.**

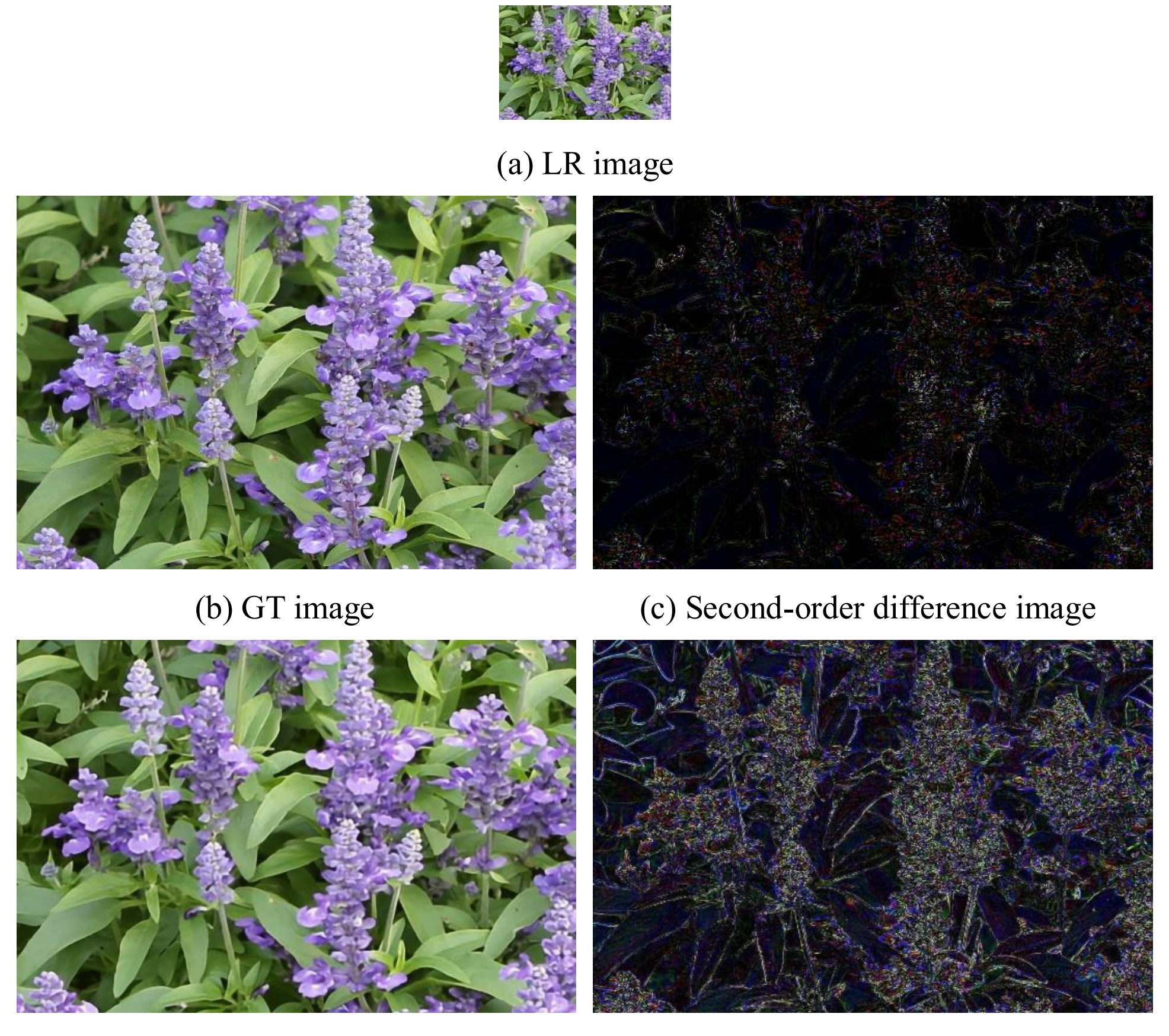

(a) LR image

(b) GT image (c) Second-order difference image

(d) SR image of RealArbiSR (PSNR=25.8766,SSIM=0.9161) (e) Difference image of RealArbiSR

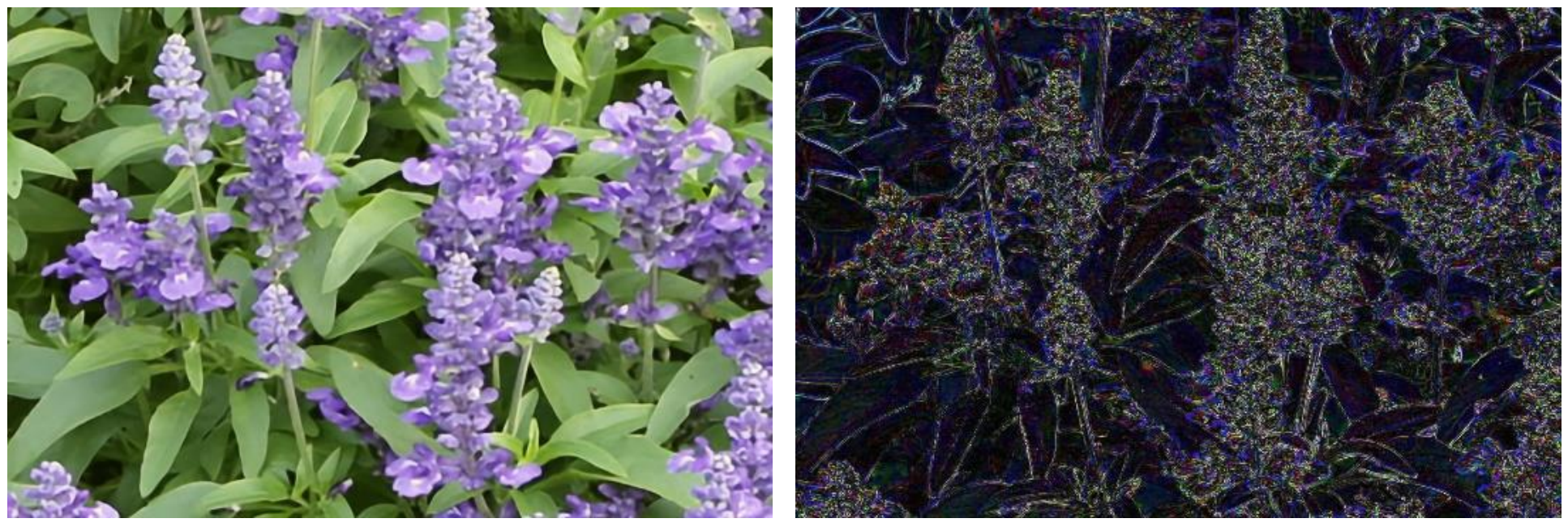

(f) SR image of CRealArbiSR (PSNR=25.9389,SSIM=0.9163) (g) Difference image of CRealArbiSR

**Fig. 28. The experimental results at scale factor 3.3.**

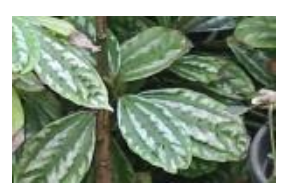

(a) LR image

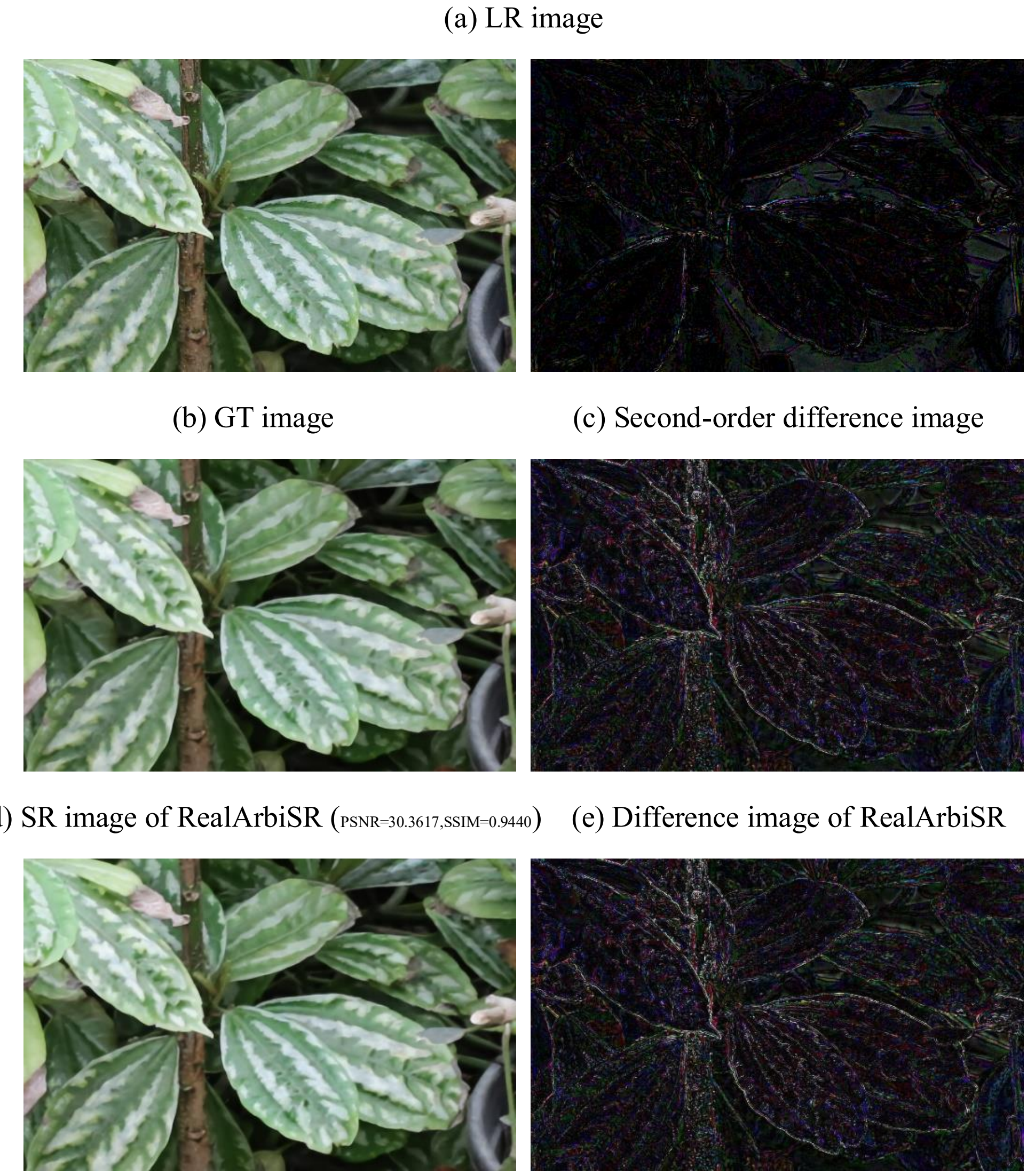

(b) GT image (c) Second-order difference image

(d) SR image of RealArbiSR (PSNR=30.3617,SSIM=0.9440) (e) Difference image of RealArbiSR

(f) SR image of CRealArbiSR (PSNR=30.4208,SSIM=0.9454) (g) Difference image of CRealArbiSR

**Fig. 29. The experimental results at scale factor 3.5.**

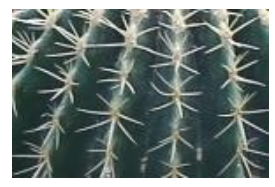

(a) LR image

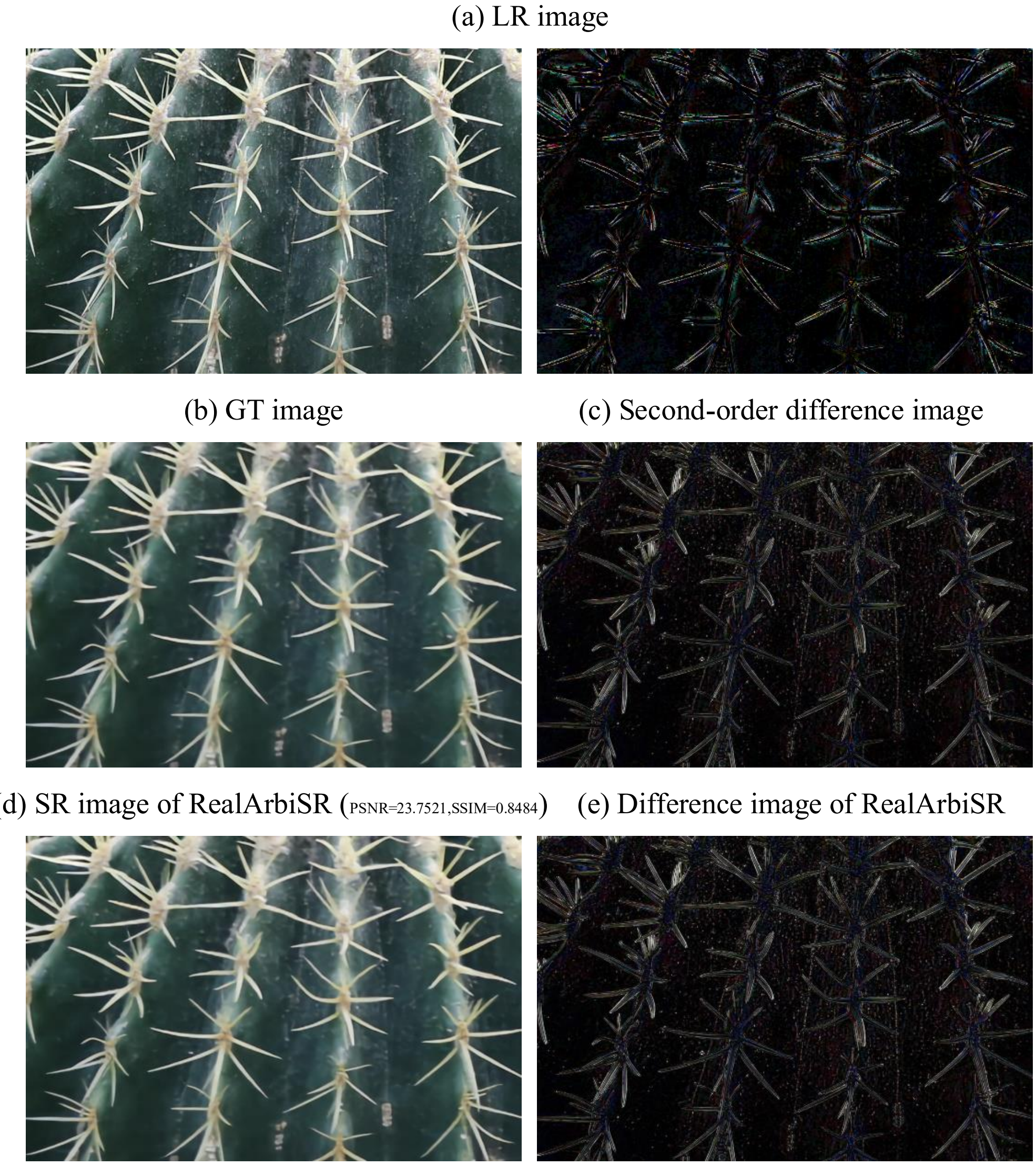

(b) GT image (c) Second-order difference image

(d) SR image of RealArbiSR (PSNR=23.7521,SSIM=0.8484) (e) Difference image of RealArbiSR

(f) SR image of CRealArbiSR (PSNR=23.9336,SSIM=0.8515) (g) Difference image of CRealArbiSR

**Fig. 30. The experimental results at scale factor 3.7.**

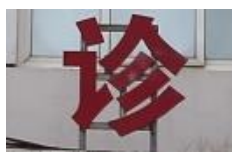

(a) LR image

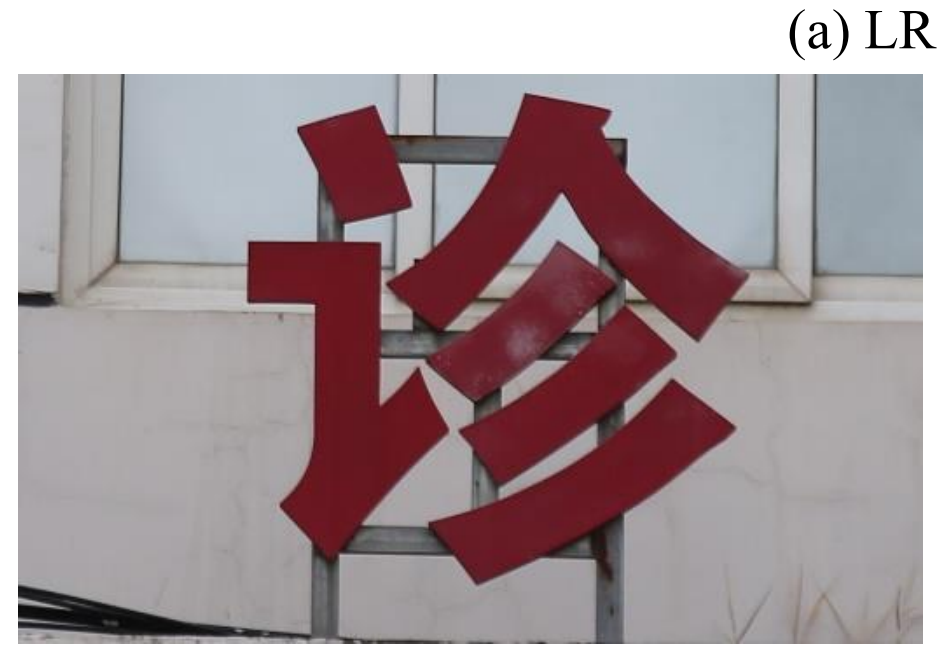

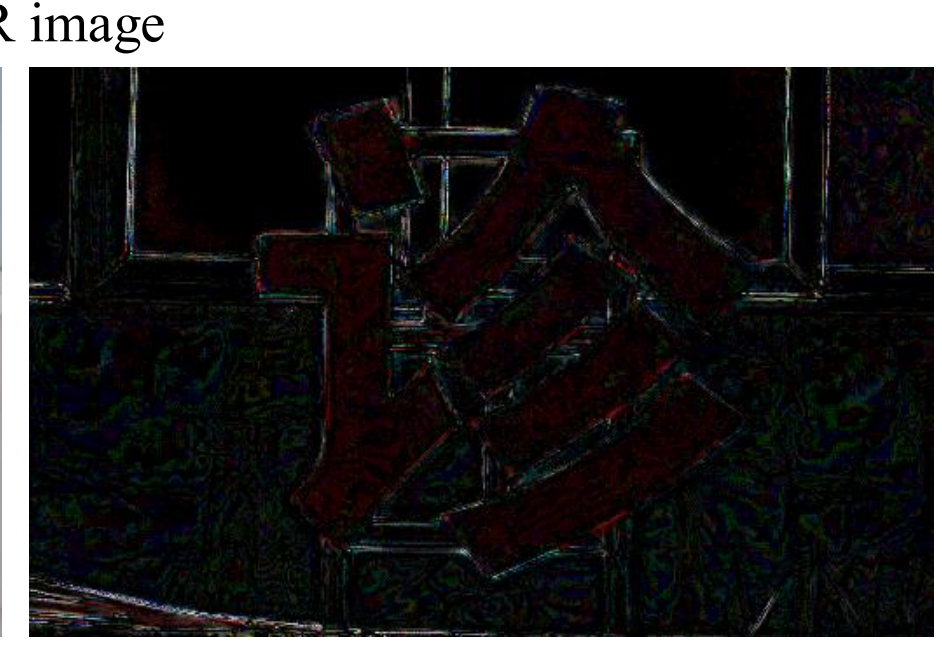

(b) GT image (c) Second-order difference image

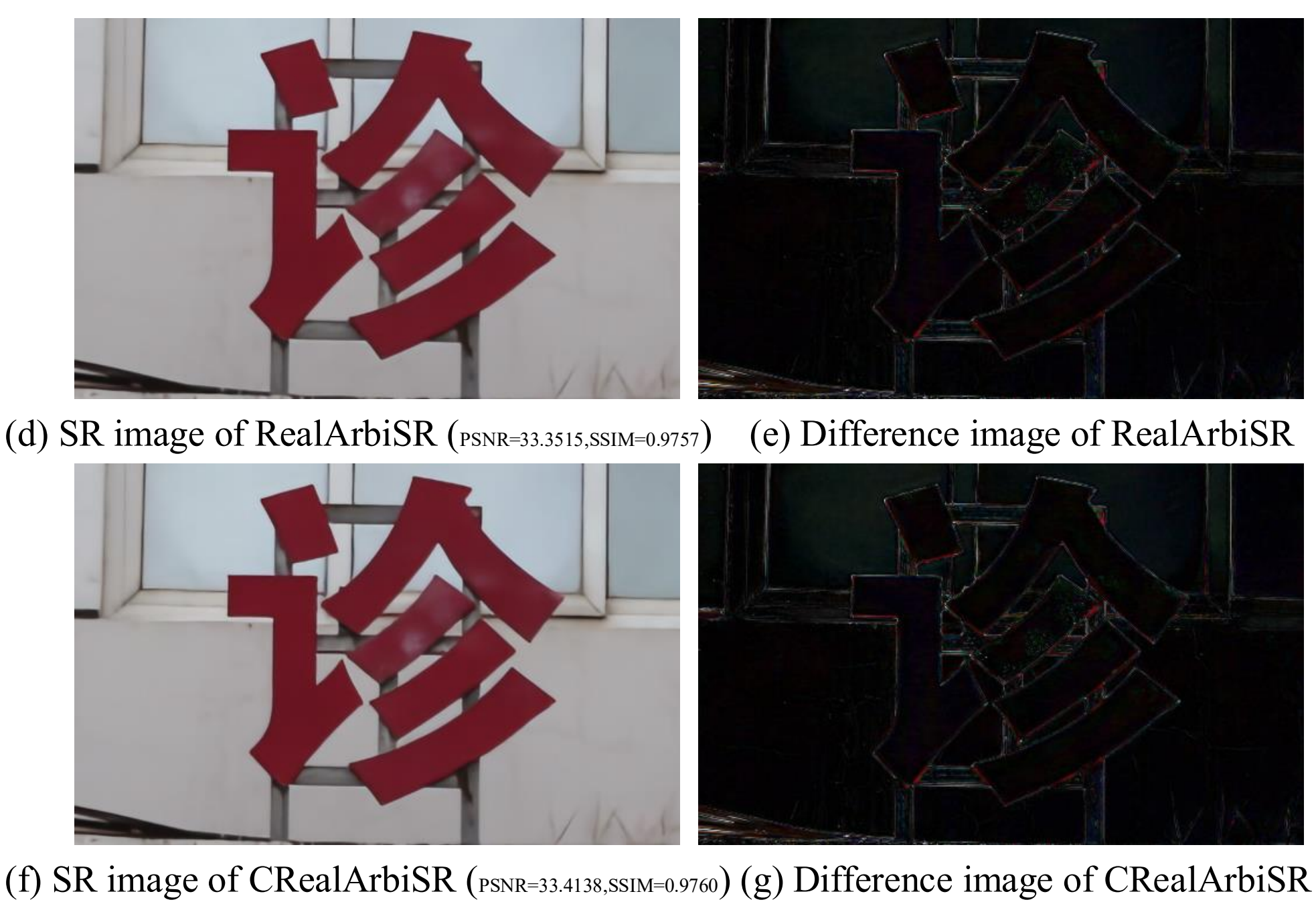

(d) SR image of RealArbiSR (PSNR=33.3515,SSIM=0.9757) (e) Difference image of RealArbiSR

(f) SR image of CRealArbiSR (PSNR=33.4138,SSIM=0.9760) (g) Difference image of CRealArbiSR

**Fig. 31. The experimental results at scale factor 4.**

## 5 Conclusions

This paper proposes a closed-loop CASISR framework based on automatic control theory to augment the GP of the pretrained advanced open-loop ASISR framework. The proposed CASISR architecture involves four elements: summation, PP, PT, and DG. The summation element introduces negative feedback, the PP element employs traditional PI control unit, the PT element adopts the advanced pretrained ASISR models, and the DG element utilizes common bicubic down-sampling. The reasonability of the proposed CASISR is demonstrated by conditional probability theory and the stability of the proposed CASISR is demonstrated by Taylor series expansion. The first-order and second-order absolute difference images are innovated to compare the image reconstruction performance of the ASISR and the CASISR methods. Simulation experiments with the eight pretrained developed ASISR models declare the proposed CSASR algorithm outperforms the competing ASISR algorithms in the performance metrics including PSNR and SSIM. The PSNR increment between the CASISR and the ASISR is huge while the related SSIM increment is non-negative at the same time. The average PSNR increment drops with oscillation while the SR scale factor rises. The average PSNR of the ASISR and the CASISR falls while the scale factor climbs. The proposed CASISR gains larger PSNR increment at the fractional SR scale factors than that at the integral SR scale factors. The proposed CASISR is extraordinarily effective for the text and stripe images with sharp edges.

In our future work, the proposed CASISR algorithm will be verified by more the ASISR algorithms, testing datasets, and SR scale factors. The PP module will also be implemented by modern control units, such as fuzzy control and adaptive control. Learnable PP, PT, and DG modules will further be explored.

# Statements and Declarations

**1 Funding**

The authors declare that no funds, grants, or other support were received during the preparation of this manuscript.

**2 Competing Interests**

The authors have no relevant financial or non-financial interests to disclose.

**3 Author Contributions**

Conceptualization, Honggui LI and Maria TROCAN; Methodology, Honggui LI and Dimitri GALAYKO; Writing, Honggui LI and Zhengyang ZHANG; Experiment, Honggui LI, Dingtai LI, Sinan CHEN, Nahid MD LOKMAN HOSSAIN, Xinfeng XU, Yinlu QIN, Ruobing WANG, Hantao LU, and Yuting FENG; Supervision, Amara AMARA and Mohamad SAWAN. All authors have read and agreed to the published version of the manuscript.

**4 Data Availability**

The datasets and pretrained models of competing algorithms analyzed during the current study are available in Tab. 2.

**5 Ethics Approval**

This research doesn't involve human or animal subjects, hence no ethical approval is required.

**6 Consent to Participate**

This research doesn't involve human subjects, hence no consent to participation is required.

**7 Consent to Publish**

This research doesn't involve human subjects, hence no consent to publication is required.

# Acknowledgment

The authors would very much like to thank all the authors of the competing algorithms for selflessly releasing their source codes of ASISR on the Microsoft GitHub website. The open-source codes allow us to easily implement the proposed CASISR depending on the competing algorithms. The authors also would very much like to express our deep thanks to Google COLAB for its free GPU computing service.

# Biographies

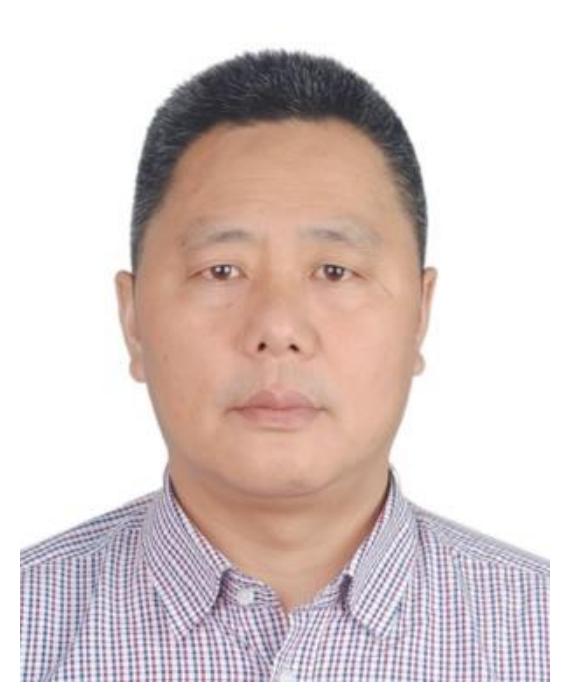

**Honggui LI** received a B.S. degree in electronic science and technology from Yangzhou University and received a Ph.D. degree in mechatronic engineering from Nanjing University of Science and Technology. He is a senior member of the Chinese Institute of Electronics. He is a visiting scholar and a post-doctoral fellow at Institut Supérieur d'Électronique de Paris for one year. He is an associate professor of electronic science and technology and a postgraduate supervisor of electronic science and technology at Yangzhou University. He is a reviewer for some international journals and conferences. He is the author of over 30 refereed journal and conference articles. His current research interests include image processing, machine learning, and integrated circuits engineering.

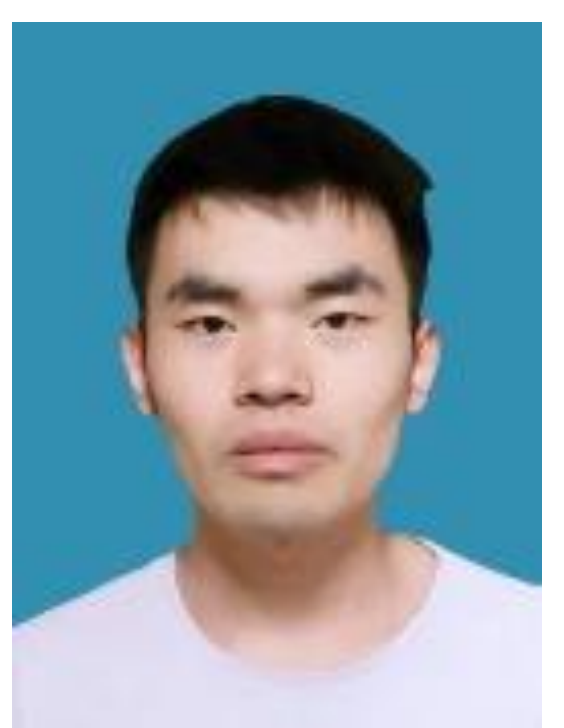

**Zhengyang ZHANG** received a B.E. degree in electronics information engineering from Yangzhou University in China. He is now studying for a master's degree in communication engineering at Yangzhou University in China. His research interests include image super-resolution, machine learning, and communication engineering.

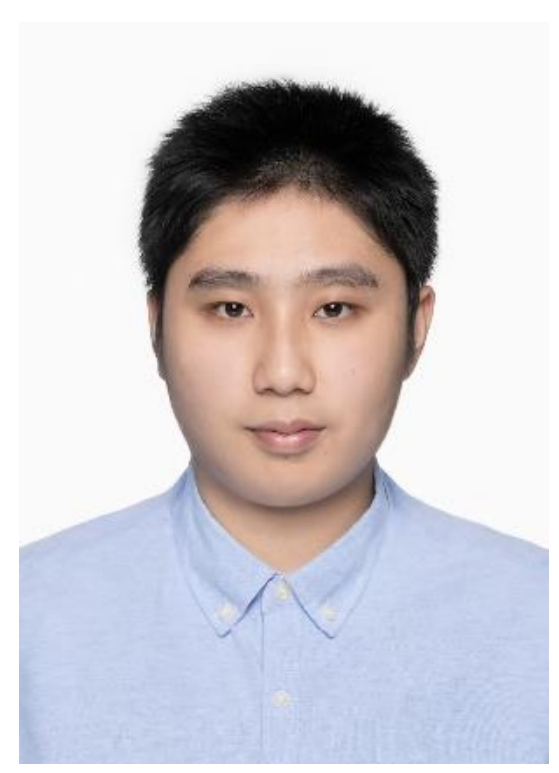

**Dingtai LI** received a B.M. degree in rehabilitation of Chinese medicine from Nanjing University of Chinese Medicine in China. He is now studying for a master's degree in acupuncture and massage at Shanghai University of Traditional Chinese Medicine in China. His research interests include Qigong, acupuncture, massage, image processing, and machine learning.

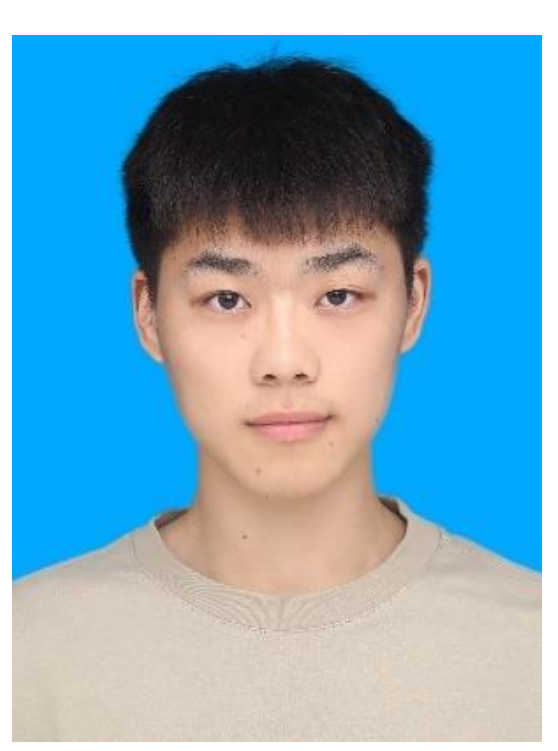

**Sinan CHEN** received a B.E. degree in electronic information engineering from Yangzhou University in China. He is now studying for a master's degree in integrated circuits engineering at Yangzhou University in China. His research interests include image compressive sensing, machine learning, and integrated circuits engineering.

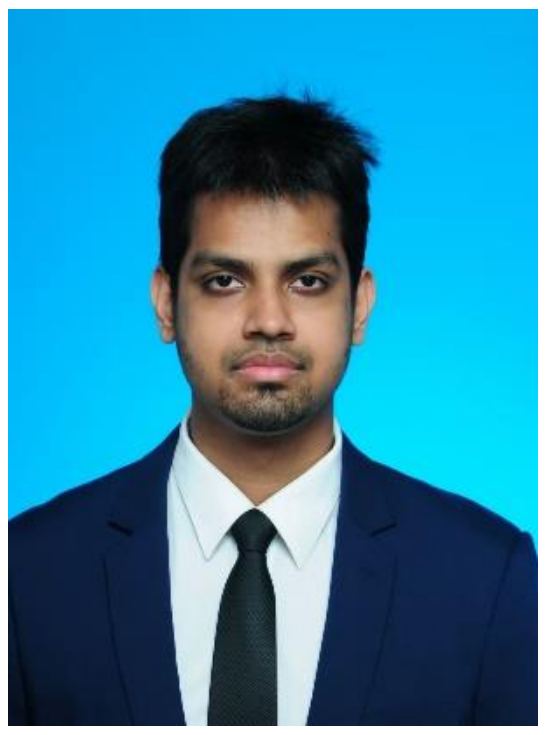

**Nahid MD LOKMAN HOSSAIN** received a B.S. degree in computer science and technology from Chongqing University of Posts and Telecommunications in China. He is now studying for a master's degree in software engineering at Yangzhou University in China. His research interests include image super-resolution, machine learning, innovation software, cloud computing, and big data.

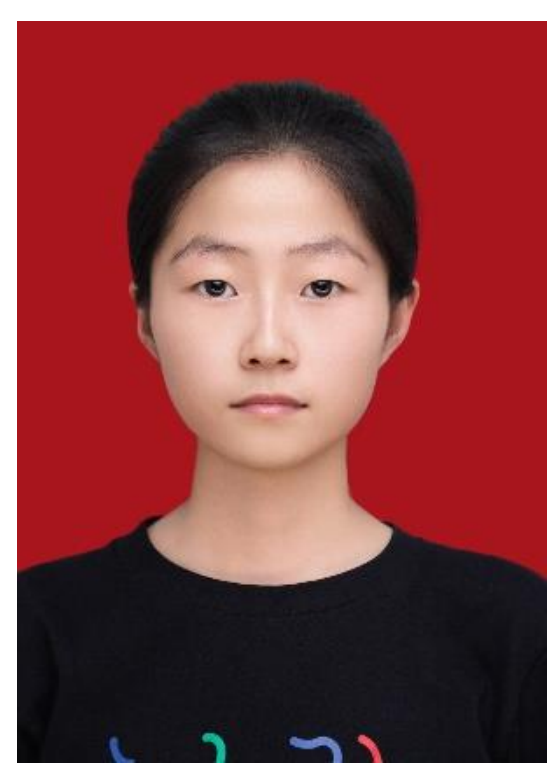

**Xinfeng XU** is now studying for a B.E. degree in electronic information engineering at Yangzhou University in China. Her research interests include image processing and machine learning.

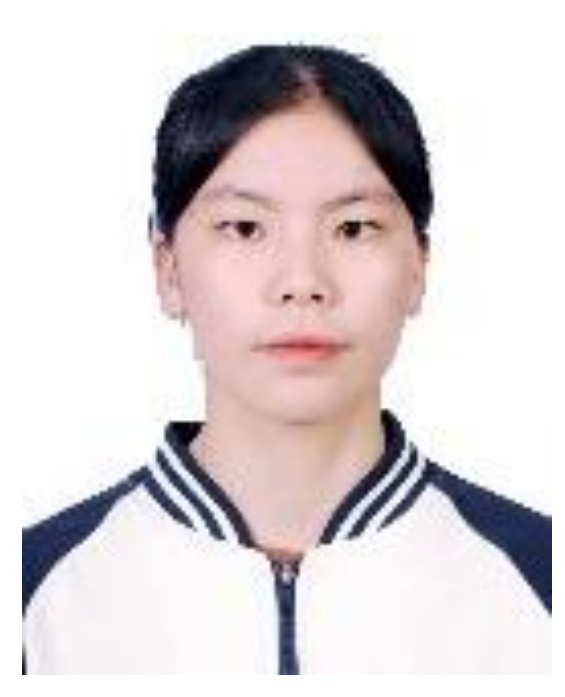

**Yinlu QIN** is now studying for a B.E. degree in electronic information engineering at Yangzhou University in China. Her research interests include image processing and machine learning.

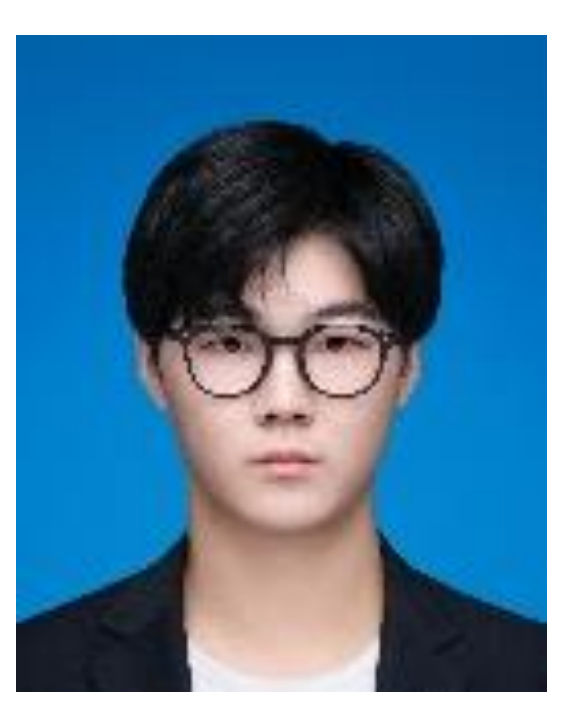

**Ruobing WANG** is now studying for a B.E. degree in electronic information engineering at Yangzhou University in China. His research interests include image processing and machine learning.

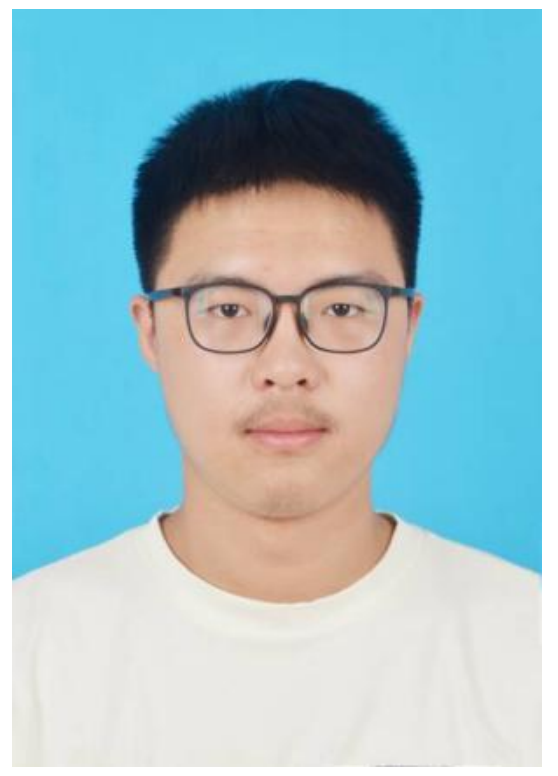

**Hantao LU** is now studying for a B.E. degree in electronic information engineering at Yangzhou University in China. His research interests include image processing and machine learning.

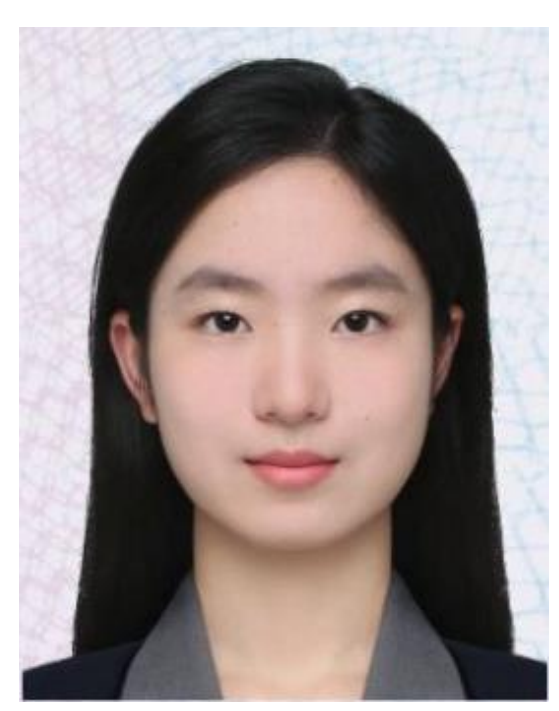

**Yuting FENG** is now studying for a B.E. degree in electronic information engineering at Yangzhou University in China. Her research interests include image processing and machine learning.

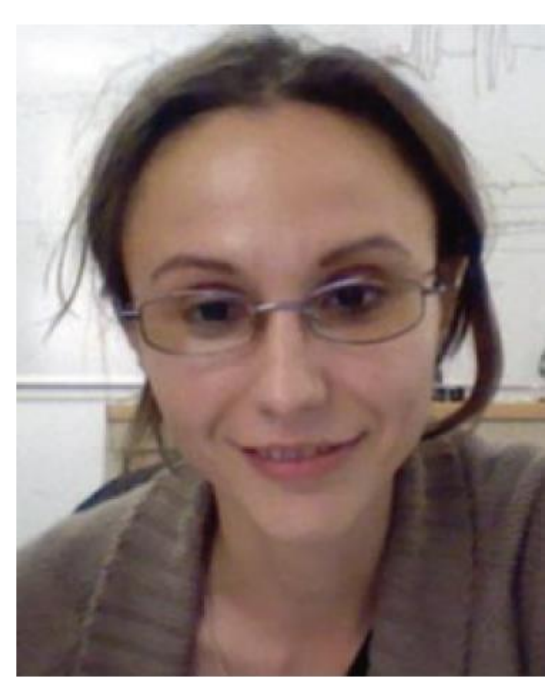

**Maria TROCAN** received a M.E. in Electrical Engineering and Computer Science from the Politehnica University of Bucharest, a Ph.D. in Signal and Image Processing from Telecom ParisTech, and the Habilitation to Lead Researches (HDR) from Pierre & Marie Curie University (Paris 6). She has joined Joost - Netherlands, where she worked as a research engineer involved in the design and development of video transcoding systems. She is firstly Associate Professor, then Professor at Institut Superieur d'Electronique de Paris (ISEP). She is an Associate Editor for the Springer Journal on Signal, Image and Video Processing and a Guest Editor for several journals (Analog Integrated Circuits and Signal Processing, IEEE Communications Magazine, etc.). She is an active member of IEEE France and served as a counselor for the ISEP IEEE Student Branch, IEEE France Vice-President responsible for Student Activities, and IEEE Circuits and Systems Board of Governors member, as Young Professionals representative. Her current research interests focus on image and video analysis & compression, sparse signal representations, machine learning, and fuzzy inference.

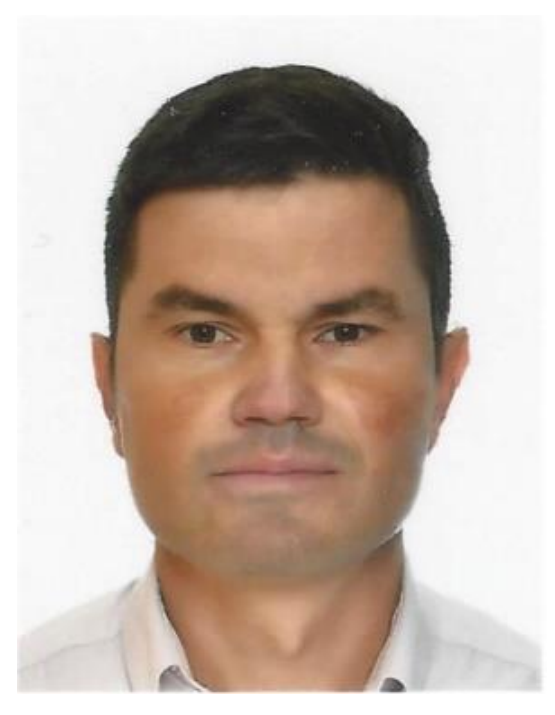

**Dimitri GALAYKO** received a bachelor's degree from Odessa State Polytechnic University in Ukraine, a master's degree from the Institute of Applied Sciences of Lyon in France, and a Ph.D. degree from University of Lille in France. He made his Ph.D. thesis at the Institute of Microelectronics and Nanotechnologies. His Ph.D. dissertation was on the design of micro-electromechanical silicon filters and resonators for radio communications. He is a Professor at the LIP6 research laboratory of Sorbonne University in France. His research interests include the study, modeling, and design of nonlinear integrated circuits for sensor interfaces and mixed-signal applications. His research interests also include machine learning and fuzzy computing.

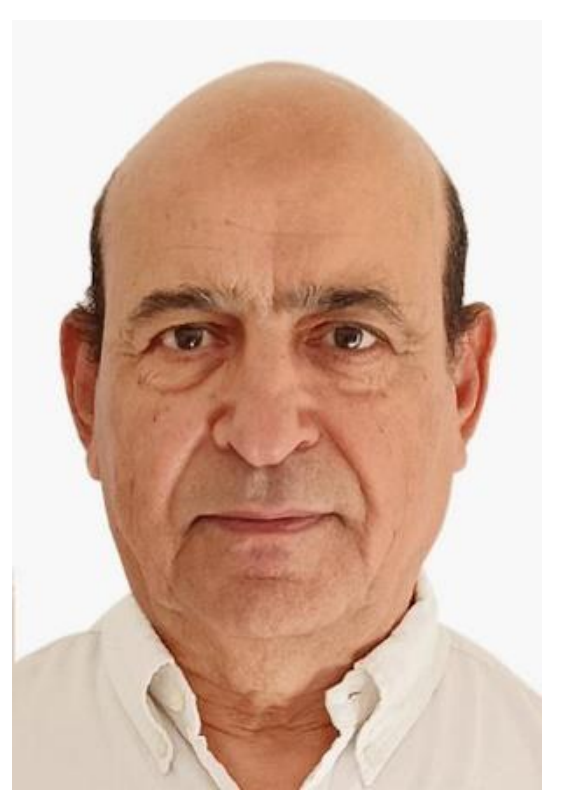

Amara AMARA (Senior Member, IEEE) received the HDR (Confirmation of Leading Research Capabilities) degree from Evry University and the Ph.D. degree from Paris VI University in 1989. In 1988, he joined the IBM Research and Development Laboratory, Corbeil-Essonnes, as a Visiting Researcher, where he was involved in SRAM memory design with advanced CMOS technologies. In 1992, he joined the Paris Institute for Electronics (ISEP) in charge of the Microelectronics Laboratory, where he headed a joint team (Paris VI and ISEP) involved in highspeed GaAs VLSI circuit design. He established the LISTE Laboratory composed of more than 40 researchers in the fields of micro and nano electronics, image and signal processing, and big data processing and analysis. He is the coauthor of three books on Molecular Electronics, Double Gate Devices and Circuits, and Emerging Technologies. He is the author or coauthor of more than 100 conference papers and journal articles. He was also the advisor of more than 20 Ph.D. students. He was the Deputy Managing Director of ISEP in charge of Research and International Cooperation since March 2017. He then joined in June 2017 "Terre des hommes" (Tdh) an international NGO specialized in child protection where he was in charge of ICT for Development (ICT4D) and Artificial Intelligence for Health. He was the President of the CAS Society from 2020 to 2021. He is the former IEEE France Section Chair and the Co-Founder and the Chair of the IEEE France CASS Chapter.

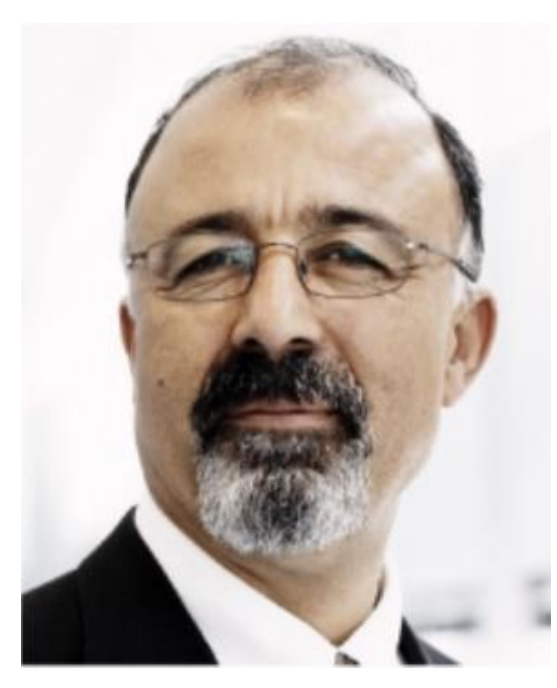

**Mohamad SAWAN** (Fellow, IEEE) received a Ph.D. degree in electrical engineering from the University of Sherbrooke, Sherbrooke, QC, Canada, in 1990. He was a Chair Professor awarded with the Canada Research Chair in Smart Medical Devices (2001–2015) and was leading the Microsystems Strategic Alliance of Quebec - ReSMiQ (1999–2018). He is a Professor of Microelectronics and Biomedical Engineering, in leave of absence from Polytechnique Montréal, Canada. He joined Westlake University, Hangzhou, China, in January 2019, where he is a Chair Professor, Founder, and Director of the Center for Biomedical Research And Innovation (CenBRAIN). He has published more than 800 peer-reviewed articles, two books, ten book chapters, and 12 patents. He founded and chaired the IEEE-Solid State Circuits Society Montreal Chapter (1999–2018) and founded the Polystim Neurotech Laboratory, Polytechnique Montréal (1994–present), including two major research infrastructures intended to build advanced Medical devices. He is the Founder of the International IEEE-NEWCAS Conference, and the Co-Founder of the International IEEE-BioCAS, ICECS, and LSC conferences. He is a Fellow of the Royal Society of Canada, a Fellow of the Canadian Academy of Engineering, and a Fellow of the Engineering Institutes of Canada. He is also the "Officer" of the National Order of Quebec. He has served as a member of the Board of Governors (2014–2018). He is the Vice-President of Publications (2019–present) of the IEEE CAS Society. He received several awards, among them the Queen Elizabeth II Golden Jubilee Medal, the Barbara Turnbull 2003 Award for spinal cord research, the Bombardier and Jacques-Rousseau Awards for academic achievements, the Shanghai International Collaboration Award, and the Medal of Merit from the President of Lebanon for his outstanding contributions. He was the Deputy Editor-in-Chief of the IEEE TRANSACTIONS ON CIRCUITS AND SYSTEMS-II: EXPRESS BRIEFS (2010–2013); the Co-Founder, an Associate

Editor, and the Editor-in-Chief of the IEEE TRANSACTIONS ON BIOMEDICAL CIRCUITS AND SYSTEMS; an Associate Editor of the IEEE TRANSACTIONS ON BIOMEDICALS ENGINEERING; and the International Journal of Circuit Theory and Applications.